\documentclass{article}
\usepackage[utf8]{inputenc}
\usepackage[backref=page]{hyperref}
\usepackage{natbib}
\usepackage{geometry}
\usepackage{graphicx}
\usepackage{booktabs}
\usepackage{amssymb}
\usepackage{microtype}
\usepackage{adjustbox}
\usepackage{subcaption}
\usepackage{float}

\usepackage{amsmath}
\usepackage{bm}
\usepackage[utf8]{inputenc} 
\usepackage[T1]{fontenc}    
\usepackage{hyperref}       
\usepackage{url}            
\usepackage{nicefrac}       
\usepackage{microtype}      
\usepackage{xcolor}         

\def\1{\bm{1}}

\def\vc{{\bm{c}}}

\def\vt{{\bm{t}}}

\def\vx{{\bm{x}}}

\def\vz{{\bm{z}}}

\DeclareMathAlphabet{\mathsfit}{\encodingdefault}{\sfdefault}{m}{sl}
\SetMathAlphabet{\mathsfit}{bold}{\encodingdefault}{\sfdefault}{bx}{n}

\def\sP{{\mathbb{P}}}

\DeclareMathOperator*{\argmin}{arg\,min}

\DeclareMathOperator{\CoSim}{CoSim}

\usepackage{tikz}

\usepackage{hyperref}
\usepackage{url}
\usepackage{multirow}
\usepackage{enumitem}

\usepackage{amsthm}
\usepackage{amssymb}
\usepackage{mathtools}
\theoremstyle{plain}

\usepackage{varwidth}
\usepackage{enumitem}
\usepackage{tikz}
\usetikzlibrary{positioning}
\usetikzlibrary{shapes}
\usetikzlibrary{fit}
\usetikzlibrary{calc}

\usepackage{listings}
\usepackage{xcolor}
\usepackage{courier}

\definecolor{codegreen}{rgb}{0,0.6,0}
\definecolor{codegray}{rgb}{0.5,0.5,0.5}
\definecolor{codeblack}{rgb}{0.,0.,0.}
\definecolor{codepurple}{rgb}{0.58,0,0.82}
\definecolor{backcolour}{rgb}{0.95,0.95,0.92}

\lstdefinestyle{mystyle}{
    backgroundcolor=\color{backcolour},   
    commentstyle=\color{codegreen},
    keywordstyle=\color{codeblack},
    numberstyle=\tiny\color{codegray},
    stringstyle=\color{codepurple},
    basicstyle=\ttfamily\footnotesize,
    breakatwhitespace=false,         
    breaklines=true,                 
    captionpos=b,                    
    keepspaces=true,                 
    numbers=left,                    
    numbersep=5pt,                  
    showspaces=false,                
    showstringspaces=false,
    showtabs=false,                  
    tabsize=2,
    aboveskip=0pt,
    belowskip=-3pt
}

\newcommand{\Quote}[2]{
}

\usepackage{pifont}
\usepackage{tgbonum}
\usepackage{cleveref}
\usepackage[framemethod=TikZ]{mdframed}
\usepackage{tikz}
\usepackage[colorinlistoftodos]{todonotes}
\usetikzlibrary{positioning,calc}
\usetikzlibrary{shapes,arrows,chains}
\usepackage[linesnumbered,ruled,vlined]{algorithm2e}
\usepackage[acronym,toc]{glossaries}

\usepackage{enumitem}
\setlist[itemize]{align=parleft,left=0pt..1em}

\usepackage{hyperref}
\hypersetup{%
    colorlinks,
    linkcolor=purple,
    citecolor=gray,
    urlcolor=purple
}

\title{An Introduction to Vision-Language Modeling}

\usepackage{authblk}

\author[*]{Florian Bordes}
\author[*$\wedge$]{\textcolor{purple}{Richard Yuanzhe Pang}}
\author[**$\clubsuit$]{\textcolor{purple}{Anurag Ajay}}
\author[**$\spadesuit$]{\textcolor{purple}{Alexander C. Li}}
\author[*]{\textcolor{purple}{Adrien Bardes}}
\author[$\bigtriangleup$]{\textcolor{purple}{Suzanne Petryk}}
\author[*$\ddagger$]{\textcolor{purple}{Oscar Mañas}}
\author[$\spadesuit$]{\textcolor{purple}{Zhiqiu Lin}}
\author[$\dagger$]{\textcolor{purple}{Anas Mahmoud}}
\author[*]{\textcolor{purple}{Bargav Jayaraman}}
\author[*]{\textcolor{purple}{Mark Ibrahim}}
\author[*]{\textcolor{purple}{Melissa Hall}}
\author[*]{\textcolor{purple}{Yunyang Xiong}}
\author[*$\heartsuit$]{\textcolor{purple}{Jonathan Lebensold}}
\author[*]{\textcolor{purple}{Candace Ross}}
\author[*]{\textcolor{purple}{Srihari Jayakumar}}
\author[*]{\textcolor{teal}{Chuan Guo}}
\author[*]{\textcolor{teal}{Diane Bouchacourt}}
\author[*]{\textcolor{teal}{Haider Al-Tahan}}
\author[*]{\textcolor{teal}{Karthik Padthe}}
\author[*]{\textcolor{teal}{Vasu Sharma}}
\author[*]{\textcolor{teal}{Hu Xu}}
\author[*]{\textcolor{teal}{Xiaoqing Ellen Tan}}
\author[*]{\textcolor{teal}{Megan Richards}}
\author[*$\ddagger$]{\textcolor{teal}{Samuel Lavoie}}
\author[*]{\textcolor{teal}{Pietro Astolfi}}
\author[*]{\textcolor{teal}{Reyhane Askari Hemmat}}
\author[**$\diamondsuit$]{\textcolor{teal}{Jun Chen}}
\author[*]{\textcolor{teal}{Kushal Tirumala}}
\author[*$\ddagger$]{\textcolor{teal}{Rim Assouel}}
\author[$\bigtriangledown$]{\textcolor{teal}{Mazda Moayeri}}
\author[*]{\textcolor{teal}{Arjang Talattof}}
\author[*]{\textcolor{teal}{Kamalika Chaudhuri}}
\author[*]{\textcolor{teal}{Zechun Liu}}
\author[*]{\textcolor{teal}{Xilun Chen}}
\author[*]{\textcolor{teal}{Quentin Garrido}}
\author[*]{\textcolor{teal}{Karen Ullrich}}
\author[$\ddagger$$\bullet$]{\textcolor{violet}{Aishwarya Agrawal}}
\author[*]{\textcolor{violet}{Kate Saenko}}
\author[*]{\textcolor{violet}{Asli Celikyilmaz}}
\author[*]{\textcolor{violet}{Vikas Chandra}}

\affil[*]{Meta}
\affil[**]{Work done while at Meta}
\affil[$\ddagger$]{Université de Montréal, Mila}
\affil[$\heartsuit$]{McGill University, Mila}
\affil[$\dagger$]{University of Toronto}
\affil[$\spadesuit$]{Carnegie Mellon University}
\affil[$\clubsuit$]{Massachusetts Institute of Technology}
\affil[$\wedge$]{New York University}
\affil[$\bigtriangleup$]{University of California, Berkeley}
\affil[$\bigtriangledown$]{University of Maryland}
\affil[$\diamondsuit$]{King Abdullah University of Science and Technology}
\affil[$\bullet$]{Canada CIFAR AI Chair}
\affil[ ]{\textcolor{purple}{Core contributors, random ordering}}
\affil[ ]{\textcolor{teal}{Additional contributors, random ordering}}
\affil[ ]{\textcolor{violet}{Senior contributors, random ordering}}
\date{}

\makeglossaries

\newacronym{VLMs}{VLMs}{Vision Language Models}

\newacronym{LLMs}{LLMs}{Large Language Models}

\newacronym{LM}{LM}{Language Model}

\newacronym{MLLMs}{MLLMs}{Multimodal Large Language Models}

\newacronym{CLIP}{CLIP}{Contrastive Language–Image Pre-training}

\newacronym{BERT}{BERT}{Bidirectional Encoder Representations from Transformers}

\newacronym{EBM}{EBM}{Energy-Based Models}

\newacronym{MCMC}{MCMC}{Markov Chain Monte Carlo}

\newacronym{NCE}{NCE}{Noise Contrastive Estimation}

\newacronym{SSL}{SSL}{Self-Supervised Learning}

\newacronym{MLM}{MLM}{Masked Language Modeling}

\newacronym{MIM}{MIM}{Masked Image Modeling}

\newacronym{FLAVA}{FLAVA}{Foundational Language And Vision Alignment}

\newacronym{ViT}{ViT}{Vision Transformer}

\newacronym{CNN}{CNN}{Convolutional Neural Network}

\newacronym{VQ-VAE}{VQ-VAE}{Vector Quantised-Variational AutoEncoder}

\newacronym{CoCa}{CoCa}{Contrastive Captioner}

\newacronym{BLIP}{BLIP}{Bootstrapping Language-Image Pre-training}

\newacronym{LLaVA}{LLaVA}{Large Language-and-Vision Assistant}

\newacronym{OBELICS}{OBELICS}{Open Bimodal Examples from Large fIltered Commoncrawl Snapshots}

\newacronym{FFCV}{FFCV}{Fast Forward Computer Vision}

\newacronym{RLHF}{RLHF}{Reinforcement Learning from Human Feedback}

\newacronym{PEFT}{PEFT}{Parameter-Efficient Fine-Tuning}

\newacronym{LoRa}{LoRa}{Low Rank Adapters}

\newacronym{VQA}{VQA}{Visual Question Answering}

\newacronym{OOD}{OOD}{Out-Of Distribution}

\newacronym{OCR}{OCR}{Optical Character Recognition}

\newacronym{ARO}{ARO}{Attribution, Relation, and Order}

\newacronym{DCI}{DCI}{Densely Captioned Images}

\newacronym{PUG}{PUG}{Photorealistic Unreal Graphics}

\newacronym{MERLOT}{MERLOT}{Multimodal Event Representation Learning Over Time}

\newacronym{DP}{DP}{Differential
Privacy}

\newacronym{STR}{STR}{Scene text recognition}
\newacronym{ROI}{ROI}{Region of Interest}

\newacronym{IoU}{IoU}{Intersection over Union}

\begin{document}

\pgfmathsetmacro{\cardroundingradius}{3mm}
\pgfmathsetmacro{\striproundingradius}{2mm}
\pgfmathsetmacro{\cardwidth}{5}
\pgfmathsetmacro{\cardheight}{8}
\pgfmathsetmacro{\stripheight}{0.4}
\pgfmathsetmacro{\strippadding}{0.1}
\pgfmathsetmacro{\textpadding}{0.2}
\pgfmathsetmacro{\ruleheight}{0.05}
\newcommand{\topsize}{\footnotesize}
\newcommand{\bottomsize}{\tiny}

\maketitle

\newpage

\tableofcontents

\newpage

\begin{abstract}
Following the recent popularity of Large Language Models (LLMs), several attempts have been made to extend them to the visual domain. From having a visual assistant that could guide us through unfamiliar environments to generative models that produce images using only a high-level text description, the vision-language model (VLM) applications will significantly impact our relationship with technology. However, there are many challenges that need to be addressed to improve the reliability of those models. While language is discrete, vision evolves in a much higher dimensional space in which concepts cannot always be easily discretized. To better understand the mechanics behind mapping vision to language, we present this introduction to VLMs which we hope will help anyone who would like to enter the field. First, we introduce what VLMs are, how they work, and how to train them. Then, we present and discuss approaches to evaluate VLMs. Although this work primarily focuses on mapping images to language, we also discuss extending VLMs to videos. 
\end{abstract}

\section{Introduction}

In recent years, we have seen impressive developments in language modeling. Many \acrfull{LLMs} such as Llama or ChatGPT are now able to solve such a large variety of tasks that their usage is becoming more and more popular. Such models that were mostly limited to text inputs are now extended to having visual inputs. Connecting vision to language will unlock several applications that will be key to the current AI-based technological revolution. Even though several works have already extended large language models to vision, connecting language to vision is not completely solved. For example, most models struggle to understand spatial relationships or count without complicated engineering overhead that relies on additional data annotation. Many \acrfull{VLMs} also lack an understanding of attributes and ordering. They often ignore some part of the input prompt, leading to significant prompt engineering efforts to produce the desired result. Some of them can also hallucinate and produce content that is neither required nor relevant. As a consequence, developing reliable models is still a very active area of research.\\

In this work, we present an introduction to \acrfull{VLMs}. We explain what VLMs are, how they are trained, and how to effectively evaluate VLMs depending on different research goals. This work \textit{should not be considered as a survey or a complete guide on VLMs}\footnote{For complete and more technical surveys on VLMs, please refer to \citet{zhang2024visionlanguage, ghosh2024exploring, zhou2023vision, Chen_2023, du2022survey, uppal2020multimodal}, and \citet{liang2023foundations}.}. Hence, we do not aim to cite every work from the VLM research field\footnote{Nevertheless, if you find errors or have any comments about missing important references, please adress them to fbordes@meta.com. We will try to include them in the next revision of this work.}; nor does this work capture every best practice in this space. Instead, we aim to provide a \textit{clear and easy-to-understand introduction} to VLM research and highlight effective practices for research in this space. This introduction should be especially useful for students or researchers in other areas who want to enter the field.\\

We start by presenting the different VLM training paradigms. We discuss how contrastive methods have changed the field. Then, we present methods that leverage masking strategies or generative components. Lastly, we present VLMs which use pre-trained backbones (such as LLMs). Categorizing VLMs into different families is not an easy task, since most of them have overlapping components. However, we hope that our categorization will help new researchers navigate the field and shed light on the inner mechanisms behind VLMs.\\ 

Next, we present typical recipes for training VLMs. For example, we cover: Which datasets are appropriate given different research goals? Which data curation strategy? Do we need to train a text encoder, or can we leverage a pre-trained LLM? Is a contrastive loss enough for vision understanding or is a generative component key? We also present common techniques used to improve model performance as well as grounding and better alignment.\\ 

While providing the recipes for training models is a crucial step for better understanding VLMs' needs, providing robust and reliable evaluation of those models is equally important. Many benchmarks that are used to evaluate VLMs have been introduced recently. However, some of these benchmarks have essential limitations that researchers should be aware of. By discussing the strengths and weaknesses of VLM benchmarks, we hope to shed light on the challenges ahead to improve our understanding of VLMs. We start by discussing the benchmarks that evaluate the visio-linguistic abilities of VLMs, and then we present how to measure biases.\\ 

The next generation of VLMs will be able to understand videos by mapping video to language. However, there are different challenges with videos that are not present with images. The computational cost is of course much higher but there are also other considerations on how to map the temporal dimension through text. By shedding light on the current methods that learn from videos, we hope to highlight the current research challenges to tackle on.\\ 

By lowering the barrier to entry into VLM research, we hope to provide the foundations for more responsible development of VLMs while pushing the boundaries of vision understanding.

\section{The Families of VLMs}

\begin{figure}
    \centering
    \includegraphics[width=\linewidth]{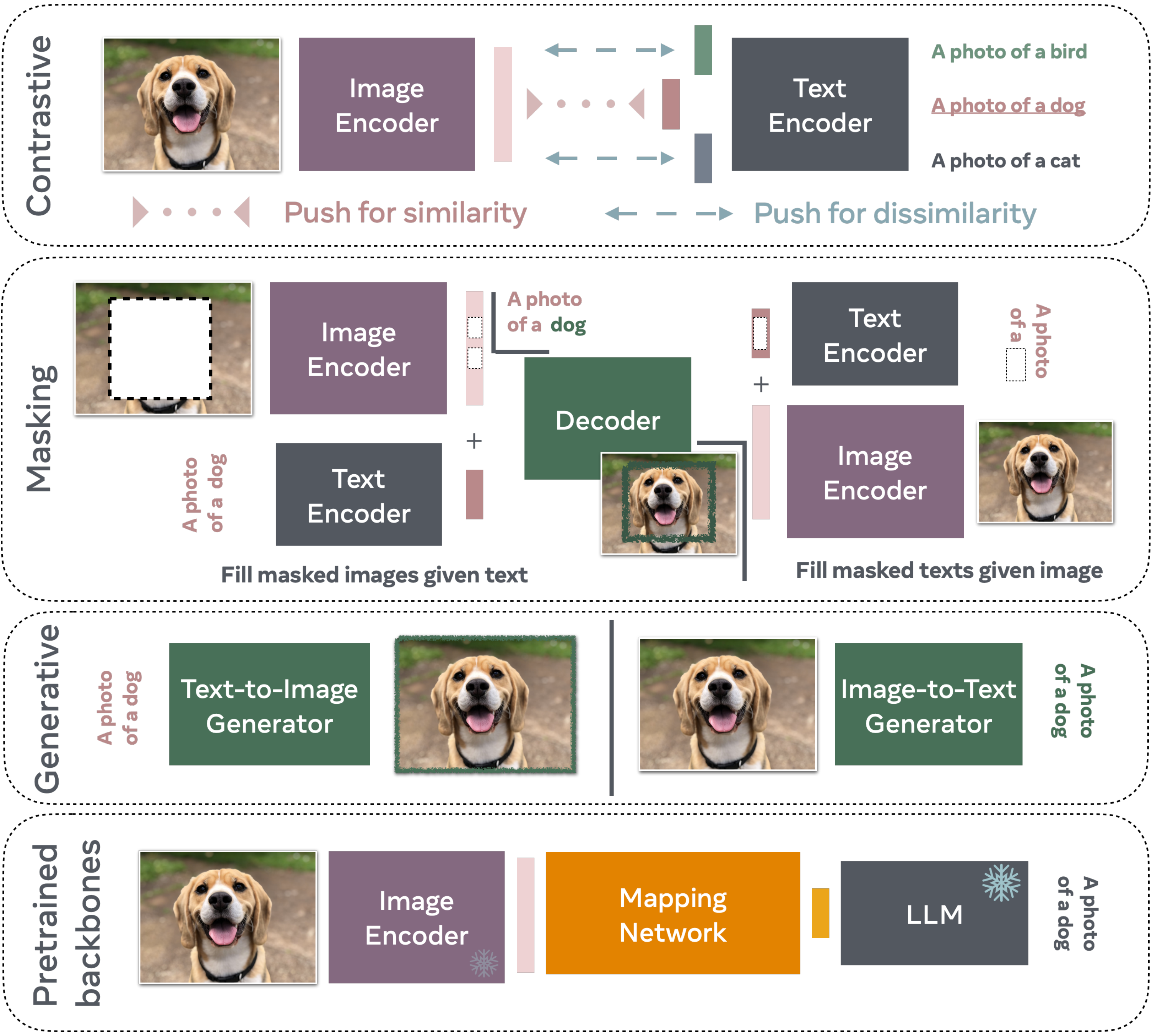}
    \caption{\small Families of VLMs. \textbf{Contrastive} training is a commonly use strategy that uses pairs of positive and negative examples. The VLM is trained to predict similar representations for the positive pairs while predicting different representations for the negative pairs. \textbf{Masking} is another strategy that can be leveraged to train VLMs by reconstructing the missing patches given an unmasked text caption. Similarly, by masking words in a caption, it is possible to train a VLM to reconstruct those words given an unmasked image. While most of those approaches leverage intermediate representations or partial reconstructions, \textbf{generative} VLMs are trained in such a way they can generate entire images or very long captions. Given the nature of those models, they are often the most expensive to train. \textbf{Pretrained backbones} based VLMs often leverage open-source LLMs like Llama to learn a mapping between an image encoder (which could also be pre-trained) and the LLM.
    It is important to highlight that these paradigms are not mutually exclusive; many approaches rely on a mix of contrastive, masking, and generative criteria.}
    \label{fig:families_vlms}
\end{figure}

Given the impressive progress powered by deep learning in the fields of computer vision and natural language processing, there have been several initiatives to bridge the two domains. In this paper we \textbf{focus on the most recent techniques based on transformers}~\citep{transformers}. We categorize these recent initiatives into four different training paradigms (Figure \ref{fig:families_vlms}). The first one around \textbf{contrastive} training is a commonly used strategy which leverages pairs of positive and negative examples. The VLM is then trained to predict similar representations for the positive pairs while predicting different representations for the negative pairs. The second initiative, \textbf{masking}, leverages reconstruction of masked image patches given some unmasked text. Similarly, by masking words in a caption, it is possible to train a VLM to reconstruct those words given an unmasked image. VLMs based on \textbf{pretrained backbones} often leverage open-source LLMs like Llama~\citep{touvron2023llama} to learn a mapping between an image encoder (which could also be pre-trained) and the LLM. Learning a mapping between pre-trained models is often less computationally expensive than training text and image encoders from scratch. While most of those approaches leverage intermediate representations or partial reconstructions, \textbf{generative} VLMs are trained in such a way that they can generate images or  captions. Given the nature of those models, they are often the most expensive to train. We highlight that \textbf{these paradigms are not mutually exclusive; many approaches rely on a mix of contrastive, masking, and generative criteria}. For each of these paradigms, we present only one or two models to give the reader some high-level insights on how those models are designed.

\subsection{Early work on VLMs based on transformers}

By using a transformer architecture~\citep{transformers}, \acrfull{BERT}~\citep{devlin-etal-2019-bert} significantly outperformed all language modelling approaches at that time. Unsurprisingly, researchers have extended BERT to process visual data. Two of them are visual-BERT~\citep{li2019visualbert} and ViLBERT~\citep{vilbert} that combine text with images tokens. The models are trained on two objectives: 1) a classical masked modelling task that aims to predict the missing part in a given input; and 2) a sentence-image prediction task that aims to predict if a caption is actually describing an image content. By leveraging these two objectives, the models achieve strong performance across several vision-language tasks, mostly explained by the ability of the transformer model to learn to associate words with visual clues through the attention mechanisms.

\subsection{Contrastive-based VLMs}

Contrastive-based training is often better explained through an \acrfull{EBM} point of view~\citep{lecun_energy} in which a model $E_{\theta}$, parameterized by $\theta$, is trained to assign low energy to observed variables and high energy to unobserved ones. Data from a target distribution should have low energy while \textit{any other data points} should have higher energy. To train these models, we consider input data $x$ with an energy function $E_{\theta}(x)$ of parameters $\theta$. The corresponding Boltzman distribution density function to learn can be written as:
\[
    p_{\theta}(x) = \frac{e^{- E_{\theta}(x)}}{Z_{\theta}}
\]
with normalization factor $Z_{\theta} = \sum_{x} e^{-E_{\theta}(x)}$. To estimate the target distribution $P_D$ from which input data are drawn, we can in principle use the traditional maximum likelihood objective:
\[
     \argmin_{\theta} \mathbb{E}_{x \sim P_D(x)} [ - \log P_{\theta}(x)]
\]
whose gradient is:
\[
    \frac{\partial \mathbb{E}_{x \sim P_D(x)} [ - \log P_{\theta}(x)]}{\partial \theta} = 
        \mathbb{E}_{x^{+} \sim P_D(x)}\frac{\partial E_{\theta}(x^{+})}{\partial \theta} - \mathbb{E}_{x^{-} \sim P_{\theta}(x)}\frac{\partial E_{\theta}(x^{-})}{\partial \theta}
\]
However, the above requires $x^{-} \sim P_{\theta}(x)$, which corresponds to a sample from the model distribution that can be intractable. There are several techniques to approximate such a distribution. One relies on \acrfull{MCMC} techniques to find examples that minimize the predicted energy through an iterative process. A second one relies on Score Matching~\citep{hyvarinen2005estimation} and Denoising Score Matching~\citep{vincent2011connection} criteria which remove the normalization factor by learning only the gradient of the probability density with respect to the input data. Another class of method, on which most of the recent works on Self-Supervised Learning and VLM are based, is \acrfull{NCE}~\citep{nce}.\\  

Instead of using the model distribution to sample negative examples, the intuition behind NCE is that sampling from a noise distribution $u'\sim p_n(u')$ might approximate samples from the model distribution well enough in certain instances. Even if it can be theoretically difficult to justify why such an approach might work, there is ample empirical evidence of the success of NCE-based methods in recent \acrfull{SSL} literature~\citep{chen2020simple}. The original NCE framework can be described as a binary classification problem in which a model should predict the label $C=1$ for samples from the real data distribution and $C=0$ for those coming from the noise distribution. By doing so, the model learns to discriminate between the real data points and the noisy ones. Thus the loss function can be defined as a binary classification with cross-entropy:
\begin{equation}
    \label{eqn:nce}
        \mathcal{L}_\text{NCE}(\theta) := -\sum_i \log P(C_i=1|x_i;\theta) -
          \sum_j \log P(C_j=0|x_j;\theta)
\end{equation}
with $x_i$ sampled from the data distribution and $x_j\sim p_n(x), j\neq i$ sampled from the noise distribution.\\

\citet{wu2018unsupervised} introduced NCE without positive pairs with a non-parametric softmax using explicit normalization and a temperature parameter $\tau$. \citet[{\bf CPC}]{oord2018representation} kept the non-parametric softmax while using positive pairs and coined this approach as {\bf InfoNCE} such that:
\begin{align}   \label{eq:infonce}       
    \mathcal{L}_{\rm infoNCE}=-\sum_{(i,j)\in\sP}\log\left(\frac{e^{\CoSim( \vz_i,\vz_j)/\tau}}{\sum_{k=1}^{N}e^{\CoSim(\vz_i,\vz_k)/\tau}}\right),
\end{align}
Instead of predicting a binary value, the InfoNCE loss leverages a distance metric, such as cosine similarity, computed in a model representation space. This requires computing this distance between the positive pairs of examples and between all of the negative pairs of examples. The model learns to predict, through the softmax, the most likely pair of examples that is closest in the representation space while associating lower probability to all other pairs of negative examples. For SSL methods such as SimCLR~\citep{chen2020simple}, a positive pair of examples is defined as one image and its corresponding handcrafted data-augmented version (such as after applying grayscaling on the original image) while the negative pairs of examples are built using one image and all other images that are present in a mini-batch. The major drawback of InfoNCE-based methods is the introduction of a dependence on mini-batch content. This often requires large mini-batches to make the contrastive training criterion between the positive and negative samples more effective.

\subsubsection{CLIP}

A common contrastive method using the InfoNCE loss is \acrfull{CLIP}~\citep{clip_openai_radford21a}.
The positive pairs of examples are defined as one image and its corresponding ground truth caption while the negative examples are defined as the same image but with all the other captions contained in the mini-batch that described the other images. One novelty of CLIP is training a model to incorporate vision and language in a shared representation space. CLIP trains randomly initialized vision and text encoders to map the representation of an image and its caption to similar embedding vectors using a contrastive loss. The original CLIP model trained on 400 million caption-image pairs collected from the web showed remarkable zero-shot classification transfer capabilities. Specifically, a ResNet-101 CLIP matched the performance of a supervised ResNet~\citep{resnet} model (attaining $76.2\%$ zero-shot classification accuracy) and surpassed it on several robustness benchmarks.\\

\textbf{SigLIP}~\citep{zhai2023sigmoid} is similar to CLIP with the exception that it uses the original NCE loss based on a binary cross-entropy instead of using the CLIP's multi-class objective based on InfoNCE. This change enables better 0-shot performances on smaller batch sizes than CLIP.\\

\textbf{Latent language image pretraining (Llip)}~\citep{lavoie2024modeling} accounts for the fact that an image can be captioned in several different ways. It proposes to condition the encoding of an image on the target caption via a cross-attention module. Accounting for the caption diversity increases the representation's expressivity and it generally improves the downstream zero-shot transfer classification and retrieval performance.

\subsection{VLMs with masking objectives}

Masking is a commonly used technique in deep learning research. It can be viewed as a specific form of denoising autoencoder~\citep{denoising_autoencoder} in which the noise has a spatial structure. It is also related to \textit{inpainting} strategies that are notably used by \citet{inpainting} to learn strong visual representations. More recently, BERT~\citep{devlin-etal-2019-bert} used \acrfull{MLM} during training to predict missing tokens in a sentence. Masking is particularly well-suited for the transformer architecture~\citep{transformers} since the tokenization of an input signal makes it easier to randomly drop specific input tokens. 
There have also been several works on the vision side to learn representations by using \acrfull{MIM} such as MAE~\citep{he2022masked} or I-JEPA~\citep{ijepa}. Naturally, there have been works that combined both techniques to train VLMs. A first one is \acrshort{FLAVA}~\citep{singh2022flava} that leverages several training strategies including masking to learn text and image representations. A second one is MaskVLM~\citep{kwon2023masked} which is a standalone model. Lastly, we make some connections between information theory and masking strategies. 

\subsubsection{FLAVA}
A first example of the masking-based approach is \acrfull{FLAVA}~\citep{singh2022flava}. Its architecture comprises three core components, each based on a transformer framework and tailored to process-specific modalities. The Image Encoder employs the \acrfull{ViT}~\citep{dosovitskiy2020image} to process images into patches for linear embedding and transformer-based representation, including a classification token $([\text{CLS}_\text{I}])$. The Text Encoder tokenizes textual input using a transformer~\citep{transformers} and embeds them into vectors for contextual processing and outputting hidden state vectors alongside a classification token $([\text{CLS}_\text{T}])$. Both of those encoders are trained using masking approaches. Building upon these, the Multimodal Encoder fuses hidden states from both the image and text encoders, leveraging learned linear projections and cross-attention mechanisms within the transformer framework to integrate visual and textual information, highlighted by an additional multimodal classification token $([\text{CLS}_\text{M}])$. The model employs a comprehensive training regimen that combines multimodal and unimodal masked modeling losses along with a contrastive objective. It is pretrained on a dataset of 70 million publicly-available image and text pairs. Through this approach, FLAVA demonstrates remarkable versatility and efficacy, achieving state-of-the-art performance across an array of 35 diverse tasks which span vision, language, and multimodal benchmarks, thereby illustrating the model's ability to understand and integrate information across different domains.

\subsubsection{MaskVLM}
One limitation of FLAVA is the use of pre-trained vision encoders such as dVAE~\citep{zhang2019d}. To make a VLM that is less dependent on third-party models, \citet{kwon2023masked} introduced MaskVLM which applies masking directly in the pixel space and in the text token space. One of the keys to make it work across both text and image is to use the flow of information coming from one modality to the other; the text reconstruction task receives the information coming from the image encoder and vice versa.

\subsubsection{Information theoretic view on VLM objectives}

\citet{federici2020learning} first show that VLMs can be understood to solve a rate-distortion problem, by reducing superfluous information and maximizing predictive information.  
\citet{dubois2021lossy} show more
specifically, that we can understand any transformation $f(X)$ on data $X$ to implicitly induce an equivalence relationship which partitions the space $f(\mathcal{X})$
into disjoint equivalence classes. We aim to constrain conditional densities to be constant within one region, i.e., $f(x)\sim f(x^\prime) \Rightarrow p(z|f(x))=p(z|f(x^\prime))$, where $Z$ is the learned representation of $X$.
This view  unifies masking and other forms of augmentation as well as a choice function between two data modalities; all can be represented as some transformation of the data.\\

We can formulate the related rate-distortion problem \citep{shwartz2024compress}:
\begin{align}\label{eq:rd}
    \argmin_{p(z| x)} \quad I(f(X);Z)  \; + \; \beta  \cdot H(X|Z).
\end{align}
To recover the masked VLM objective, we bound Equation \eqref{eq:rd};
\begin{equation}\label{eq:invariant_vae}
\mathcal{L}
=
- \sum_{x \in \mathcal{D}}
\mathbb{E}_{p(f)p(Z|f(x))}\left[
\log q(z) + \beta \cdot \log q(x | z)\right].
\end{equation}
where  $\log q(z)$ is an entropy bottleneck, bounding the rate $I(f(X);Z)$, removing superfluous information. Note that the entropy bottleneck in masking VLMs is typically bounded by a constant that depends on the amount information removed by masking. For multimodal VLMs, the amount of information in $Z$ is reduced to the minimum amount of information from either source.
The term $\log q(x | z)$ bounds the distortion $H(Z|X)$ and ensures the preservation of  information and hence maximizes predictive information. Practically, this term is realized by auto-encoding.
In contrast, contrastive losses can be seen as compression without data reconstruction. Here the distortion, see \eqref{eq:infonce}, scores the equivalence of two representations. InfoNCE retains the necessary information by classifying  which $Z$ is associated with an equivalent example $X$.\\

As a result of the information theoretic view, we understand the contrastive loss and auto-encoding loss as implementations of distortions, whereas the rate is mostly determined by the data transformation used.

\subsection{Generative-based VLMs}
In contrast to previous training paradigms which mostly operate on latent representations to build images or text abstractions that are then mapped between each other, the generative paradigm considers the generation of text and/or images. Some methods like CoCa~\citep{yu2022coca} learn a complete text encoder and decoder which enable image captioning. Some others, like Chameleon \cite{chameleonteam2024chameleon} and CM3leon~\citep{yu2023scaling}, are multi-modals generative models that are explicitly trained to generate both text and images. Lastly, some models are only trained to generate images based on text such as Stable Diffusion~\citep{rombach2022high}, Imagen~\citep{saharia2022photorealistic}, and Parti~\citep{parti}. However, even if they are trained to only generate images, they can also be leveraged to solve several vision-language understanding tasks.

\subsubsection{An example of learning a text generator: CoCa}
Besides the contrastive loss that works well in CLIP, \acrfull{CoCa}~\citep{yu2022coca} also employs a generative loss, which is the loss corresponding to captions generated by a multimodal text decoder that takes in (1) image encoder outputs and (2) representations produced by the unimodal text decoder as inputs. The new loss allows the ability to perform new multimodal understanding tasks (e.g., VQA) without the need for further adaptation using multimodal fusion modules. 
CoCa is pretrained from scratch by simply treating annotated image labels as text. Pretraining relies on two datasets: ALIGN which contains $\sim$1.8B images with alt-text, as well as JFT-3B which is an internal dataset that consists of $>$29.5k classes as labels but treating labels as alt-text. 

\subsubsection{An example of multi-modal generative model: Chameleon and CM3leon}
\citet{yu2023scaling} introduce CM3Leon, a foundation model for text-to-image and image-to-text generation. CM3Leon borrows the image tokenizer from~\citet{gafni2022make} which encodes a 256 $\times$ 256 image into 1024 tokens from a vocabulary of 8192. It borrows the text tokenizer from~\citet{zhang2022opt} with a vocabulary size of 56320. It introduces a special token \textit{<break>} to indicate transitions between modalities. This tokenization approach allows the model to process interleaved text and images. The tokenized images and texts are then passed to a decoder-only transformer model~\citep{brown2020language, zhang2022opt} which parameterizes the CM3Leon model.\\

The CM3Leon model undergoes a two-stage training process. The first stage is retrieval-augmented pretraining. This phase uses a CLIP-based encoder~\citep{clip_openai_radford21a} as a dense retriever to fetch relevant and diverse multimodal documents and prepends these documents to the input sequence. The model is then trained using next token prediction on the input sequence. The retrieval augmentation effectively increases the tokens available during pretraining thereby increasing data-efficiency. The second stage involves supervised fine-tuning (SFT), where the model undergoes multi-task instruction tuning. This stage allows the model to process and generate content across different modalities, significantly improving its performance on a variety of tasks including text-to-image generation and language-guided image editing. These stages collectively enable CM3Leon to achieve state-of-the-art performance in multi-modal tasks, demonstrating a significant advancement in the capabilities of autoregressive models for handling complex interactions between text and images. \\

An extension to this work is Chameleon, a new series of mixed-modal foundation models \citep{chameleonteam2024chameleon} that can generate and reason with mixed sequences of interleaved textual and image content. This capability allows for comprehensive multimodal document modeling, extending beyond typical multimodal tasks like image generation, image comprehension, and text-only language models. Chameleon is uniquely designed to be mixed-modal from the beginning, utilizing a uniform architecture trained from scratch in an end-to-end manner on a blend of all modalities—images, text, and code. This integrated approach employs fully token-based representations for both images and text. By converting images into discrete tokens, similar to words in text, the same transformer architecture can be applied to sequences of both image and text tokens without needing separate encoders for each modality. This early-fusion strategy, where all modalities are mapped into a shared representational space from the outset, enables seamless reasoning and generation across different modalities. However, this also introduces significant technical challenges, especially in terms of optimization stability and scaling. These challenges are addressed through a combination of architectural innovations and training techniques, including novel modifications to the transformer architecture such as query-key normalization and revised layer norm placements, which are crucial for stable training in a mixed-modal environment. Additionally, they demonstrate how to adapt supervised fine-tuning approaches used for text-only language models to the mixed-modal context, achieving strong alignment at scale.

\subsubsection{Using generative text-to-image models for downstream vision-language tasks}

Large advancements have recently been made on language-conditioned image generative models~\citep{bie2023renaissance, zhang2023texttoimage}, from diffusion models like Stable Diffusion~\citep{rombach2022high} and Imagen~\citep{saharia2022photorealistic} to autoregressive models like Parti~\citep{parti}. While the focus has been on their \textit{generative} abilities, they can actually be directly used for \textit{discriminative} tasks like classification or caption prediction without any retraining.\\

These generative models are trained to estimate $p_\theta({\vx} \mid {\vc})$, the conditional likelihood of the image ${\vx}$ given a text prompt ${\vc}$. Then, given an image ${\vx}$ and a set of $n$ text classes $\{{\vc}_i\}_{i=1}^n$, classification can be easily done via Bayes' theorem: 
\begin{align}
\label{eq:gen_clf}
    p_\theta({\vc}_i \mid {\vx}) = \frac{p({\vc}_i)\ p_\theta({\vx} \mid {\vc}_i)}{\sum_j p({\vc}_j)\ p_\theta({\vx} \mid {\vc}_j)}
\end{align}

Performing discriminative tasks with conditional generative models is not a new idea -- generative classification, or ``analysis by synthesis''~\citep{yuille2006vision}, has been a core idea behind foundational methods like Naive Bayes~\citep{rubinstein1997discriminative,ng2001discriminative} and linear discriminant analysis~\citep{fisher1936use}. These generative approaches to classification have traditionally been limited by weak generative modeling capabilities; however, today's generative models are so good that generative classifiers are becoming competitive again. 

\paragraph{Likelihood estimation with autoregressive models.}

Most state-of-the-art autoregressive models in other modalities (such as language or speech) act on discrete tokens as opposed to raw inputs. This is relatively simple for modalities such as language and speech, which are inherently discrete, but difficult for continuous modalities such as images. In order to effectively leverage techniques from auto-regressive modeling such as LLMs, practitioners generally train an image tokenizer, which maps an image to a sequence of discrete tokens $({\vt}_1, \cdots,
{\vt}_K)$. After turning an image into a sequence of discrete tokens (e.g., tokenizing the image), estimating the image likelihood is straightforward: 
\begin{align}
\label{eq:autoregressive}
    \log p_\theta({\vx} \mid {\vc}_i) = \sum_{j=1}^K \log p_\theta({\vt}_j \mid {\vt}_{< j}, {\vc}_i)
\end{align}
where $p_\theta$ is parameterized by the autoregressive VLM. Given that this tokenization is a crucial part of auto-regressive VLMs, one might ask: how do we train image tokenizers? Many current image tokenizers are based on the \acrfull{VQ-VAE}~\citep{van2017neural} framework, which stitches together an auto-encoder (responsible for creating good compressed \textit{continuous} representations) with a Vector Quantization layer (responsible for mapping continuous representations to discrete representations). The architecture is generally a \acrfull{CNN}~~\citep{CNN} encoder, followed by a Vector Quantization layer, followed by a CNN decoder. The actual discretization step occurs in the vector quantization layer, which maps encoder outputs to the closest embedding in a learned embedding table (``learned'' here means that the embedding table is updated throughout training). The loss function for the tokenizer is a combination of reconstruction loss in pixel space (e.g., L2 distance between input and reconstructed pixels) as well as codebook commitment losses to encourage encoder outputs and codebook embeddings to be close to each other. Most modern image tokenizers improve upon this VQ-VAE framework, by either adding different losses or changing the architecture of the encoder/decoder. Notably, VQ-GAN~\citep{esser2021taming} adds perceptual losses and adversarial losses (which involve including a discriminator between ground truth and reconstructed images) to capture more fine-grained details. VIT-VQGAN~\citep{yu2021vector} uses a Vision Transformer instead of CNN for the encoder and decoder architecture.

\paragraph{Likelihood estimation with diffusion models.}
Obtaining density estimates with diffusion models is more challenging, as they do not directly output $p_\theta(\mathbf{x} \mid \mathbf{c})$. Instead, these networks $\epsilon_\theta$ are typically trained to estimate the noise $\epsilon$ in a noisy image $\mathbf{x_t}$. Thus, diffusion-based classification techniques~\citep{li2023diffusion,clark2023text} estimate a (typically reweighted) variational lower bound for the conditional image likelihood: 
\begin{align}
\label{eq:diffusion_elbo}
    \log p_\theta({\vx} \mid {\vc}_i) \propto -\mathbb{E}_{t, \epsilon}\left[\|\epsilon - \epsilon_\theta({\vx_t}, {\vc}_i)\|^2 \right]
\end{align}
The lower the noise prediction error, the higher the conditional likelihood $p_\theta({\vx} \mid {\vc})$ is.
Measuring the bound in Equation~\eqref{eq:diffusion_elbo} relies on repeated sampling to obtain a Monte Carlo estimate. 
\citet{li2023diffusion} and \citet{clark2023text} develop techniques for reducing the number of samples required, dynamically allocating samples to the most likely classes and ensuring that the added noise $\epsilon$ is matched across all potential classes. However, even with these techniques, classification with conditional diffusion models is still computationally expensive, scaling with the number of classes and requiring hundreds or thousands of network evaluations per test image. Thus, while classification performance with diffusion models is quite good, inference is impractical until further optimizations are developed. 

\paragraph{Advantages of generative classifiers.}
Though inference with these generative classifiers is more expensive, they do have significant advantages. Generative classifiers have more ``effective robustness,'' which means that they have better out-of-distribution performance for a given in-distribution accuracy~\citep{li2023diffusion}. On compositional reasoning tasks like Winoground~\citep{thrush2022winoground}, generative classifiers far outperform discriminative methods like CLIP~\citep{li2023diffusion,clark2023text}. Generative classifiers, whether autoregressive (Parti) or diffusion-based (Imagen), have been shown to have more shape bias and align better with human judgement~\citep{jaini2023intriguing}. Finally, generative classifiers can be jointly adapted with discriminative models at test-time using only unlabeled test examples~\citep{prabhudesai2023test}. This has been shown to improve performance on classification, segmentation, and depth prediction tasks, especially in online distribution shift scenarios. 

\subsection{VLMs from Pretrained Backbones}

A downside of VLMs is that they are costly to train from scratch. They often require hundreds to thousands of GPUs while having to use hundreds of millions of images and text pairs. Thus, there has been much research work that instead of training models from scratch tried to leverage existing large-language models and/or existing visual extractors. Most of those works are motivated by the fact that many large language models are open-source and thus can be easily used. By leveraging such models, it is possible to then learn a mapping only between the text modality and the image modality. Learning such a mapping enables the LLMs to answer visual questions while requiring a low amount of compute resources. In this section, we present only two of those models, the first one being Frozen~\citep{tsimpoukelli2021multimodal} which is a first model that leverages pretrained LLMs. Then, we introduce the family of model Mini-GPT~\citep{zhu2023minigpt}. 

\subsubsection{Frozen}
Frozen~\citep{tsimpoukelli2021multimodal} is a first example of a model leveraging a pretrained LLM. This work proposes to connect vision encoders to \emph{frozen} language models through a lightweight mapping network which projects visual features to text token embeddings. The vision encoder (NF-ResNet-50~\citep{brock2021high}) and the linear mapping are trained from scratch, while the language model (a 7 billion-parameter transformer trained on C4~\citep{raffel2020exploring}) is kept frozen (this is crucial  to maintain the features that the pre-trained model had already learned). The model is supervised with a simple text generation objective on Conceptual Captions~\citep{sharma2018conceptual}. At inference time, the language model can be conditioned on interleaved text and image embeddings. The authors show the model is capable of rapid adaptation to new tasks, fast access to general knowledge, and fast binding of visual and linguistic elements. While achieving only modest performance, Frozen has been an important first step toward the current Multimodal LLMs capable of open-ended multimodal zero/few-shot learning.

\subsubsection{The example of MiniGPT}
Starting from models like Flamingo~\citep{alayrac2022flamingo}, a recent trend is to train multimodal language models where the input contains text and images, and the output contains text (and optionally images). MiniGPT-4 \citep{zhu2023minigpt} accepts text input and image input, and it produces text output. In MiniGPT-4, a simple linear projection layer is used in order to align image representation (using the same visual encoder in BLIP-2~\citep{blip2}, which is based on Q-Former and a ViT backbone) with the input space of the \textit{Vicuna language model}~\citep{vicuna2023}. Given that the visual encoder and Vicuna language model are already pretrained and used as off-the-shelf from prior work, MiniGPT-4 requires only training the linear project layer which is done in two rounds. The first involves 20k training steps (with a batch size of 256), corresponding to around 5M image-text pairs from Conceptual Caption~\citep{sharma2018conceptual}, SBU~\citep{sbu}, and LAION~\citep{schuhmann2021laion}.  The authors only used four A100 GPUs for around ten hours given that only the linear projection layer parameters needed to be trained. The second round of training leveraged highly-curated data in an instruction-tuning format, needing only 400 training steps (with a batch size of 12).\\

MiniGPT-5~\citep{zheng2023minigpt5} extends MiniGPT-4 so that the output can contain text interleaved with images. To generate images as well, MiniGPT-5 used generative tokens which are special visual tokens that can be mapped (through transformer layers) to feature vectors, which in turn can be fed into a frozen Stable Diffusion 2.1 model~\citep{rombach2021highresolution}. The authors used supervised training on downstream tasks (e.g., multi-modal dialogue generation and story generation).\\

LLMs have served as a universal interface for many language-related applications, e.g., a general chatbot. Inspired by this, MiniGPT-v2~\citep{chen2023minigpt} proposed to perform various vision-language tasks such as image captioning, visual question answering, and object grounding, through a unified interface. To achieve the goal of performing these effectively, MiniGPT-v2 introduced unique identifiers for different tasks when training, enabling the model to distinguish each task instruction effortlessly and also learn efficiently. The experimental results on visual question answering and visual grounding benchmarks show that MiniGPT-v2 demonstrates strong vision-language understanding abilities.

\subsubsection{Other popular models using pretrained backbones}

\paragraph{Qwen.} Similar to MiniGPT-4, Qwen-VL and Qwen-VL-Chat~\citep{Qwen-VL} models rely on an LLM, a visual encoder, and a mechanism that aligns the visual representation to the input space of the LLM. In Qwen, the LLM is \textit{initialized from Qwen-7B}~\citep{qwen}, the visual encoder is based on ViT-bigG, and a one-layer cross-attention module is used to compress visual representation to a sequence of fixed length (256) which is fed into the LLM. 

\paragraph{BLIP2.} 
\citet{blip2} introduce BLIP-2, a vision-language model that takes images as input and generates text output. It leverages pretrained, frozen models to greatly shorten training time: a vision encoder (such as CLIP) produces image embeddings that are mapped into the input space of an \textit{LLM such as OPT}. A relatively small ($\sim$100-200M parameters) component called a Q-Former is trained for this mapping -- it is a Transformer that takes in a fixed number of randomly-initialized ``query'' vectors; in the forward pass, the queries interact with image embeddings via cross-attention in the Q-Former, followed by a linear layer that projects the queries to the LLM's input space.\\

There are many more models based on pretrained LLMs in the literature. Each LLM ends up being extended to a VLM version which means that the scope of a specific survey on such topic would be very large. In this introduction, we aim to present a select few as they all rely on the same principles of learning mappings between representations.

\section{A Guide to VLM Training}

Several works~\citep{scaling_law, scaling_law_gen} have shed light on the importance of scaling to push further the performances of deep neural networks. Motivated by these scaling laws, most recent works have focused on increasing compute and scale to learn better models. This led to a model like CLIP~\citep{clip_openai_radford21a} which was trained on 400M images using a remarkably high compute budget. Even its corresponding open-source implementation, OpenCLIP~\citep{openclip} was trained using between 256 and 600 GPUs across multiple days or weeks depending on the model size. However, recent work~\citep{sorscher2022beyond} has shown that it is possible to beat the scaling law using a data curation pipeline. In this section, we first discuss the importance of data when training models and present some of the recipes that are used to create datasets for training VLMs. Then, we discuss the common software, tools and tricks that practitioners might use to train VLMs more efficiently. Since there are different methods to train VLMs, we also discuss what type of models to choose in specific situations. Lastly, we present some tricks on how to improve grounding (the ability to correctly map text with visual clues). We also introduce techniques to improve alignment using human preferences. VLMs are often used to read and translate text, so we also present some of the techniques that can be used to push further the OCR capabilities of VLMs. Lastly, we discuss the common fine-tuning methods. 

\begin{figure}[ht]
    \centering
    \includegraphics[width=\linewidth]{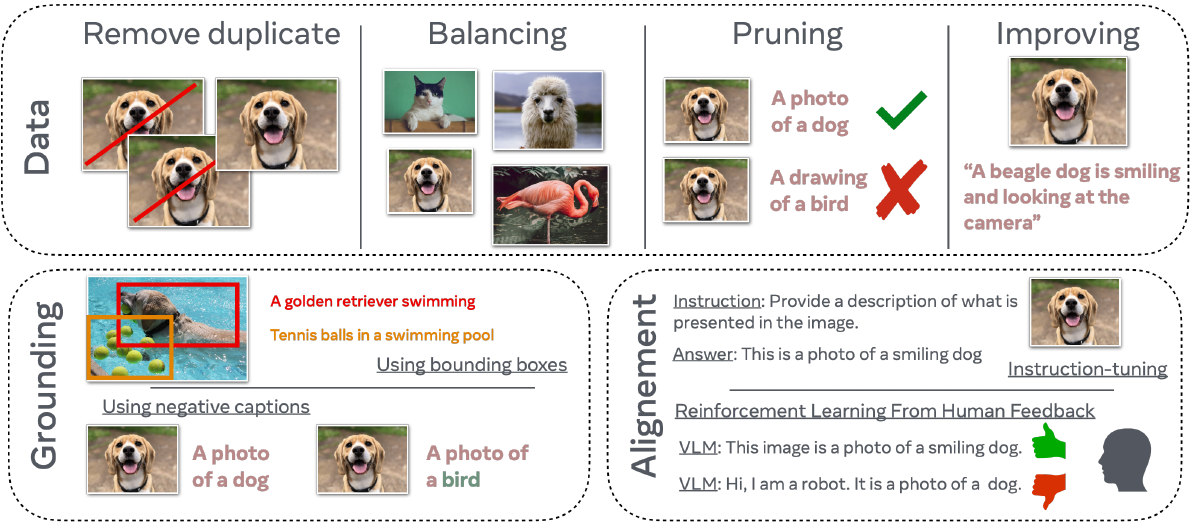}
    \caption{Important considerations to keep in mind when training VLMs. \textbf{Data} is one of the most important aspects of training VLMs. Having a diverse and balanced dataset is important for learning good world models that can span enough concepts. It is also important to remove duplicates which occur a lot within large-scale datasets, this will save a lot of compute time and mitigate the risks of memorization. In addition, pruning the data is also an important component since we want to be sure that the captions are indeed related to the image content. Lastly, improving the caption quality is crucial to enhance VLMs performance. 
    \textbf{Grounding} VLMs is another important step to ensure that the VLMs correctly associate words with specific concepts. Two common grounding methods leverage either bounding boxes or negative captions. Lastly, \textbf{alignment} is a much-needed step to ensure that the model is producing answers that are expected from a human point of view.}
    \label{fig:clip}
\end{figure}

\subsection{Training data}

To evaluate the quality of pretraining datasets, DataComp~\citep{gadre2023datacomp} proposes a benchmark where the model architecture and pretraining hyperparameters of CLIP are fixed. The focus is on designing image-text datasets that achieve strong zero-shot and retrieval performance on 38 downstream tasks. DataComp provides multiple pools of noisy web datasets, ranging from small (1.28 million) to extra-large (12.8 billion) image-text pairs. For each pool, multiple filtering strategies are proposed and evaluated. DataComp demonstrates that data pruning is a crucial step in training highly efficient and performant VLMs. Data-pruning methods for VLMs can be categorized into three categories: (1) heuristics that eliminate low-quality pairs; (2) bootstrapping methods that utilize pretrained VLMs to rank image-text pairs based on their multimodal alignment, discarding poorly aligned pairs; and finally, (3) methods that aim to create diverse and balanced datasets.\\

\noindent \textbf{Heuristics:} Filters based on heuristics can be further categorized into unimodal and multimodal filters. Unimodal heuristics include removing captions with low text complexity as measured by the number of objects, attributes, and actions~\citep{DiHT}, eliminating non-English alt-text using fastText~\citep{fasttext}, and removing images based on their resolution and aspect ratio~\citep{gadre2023datacomp}. Multimodal heuristics involve methods that employ image classifiers to filter out image-text pairs for which none of the objects detected in the image map to any of the text tokens~\citep{conceptualcaptions}. Additionally, since web-scale datasets often display part of the caption as text in the image, multimodal heuristics, such as text spotting, aim to eliminate image-text pairs with high overlap using off-the-shelf text-spotters~\citep{kuang2021mmocr}. This results in models that learn to extract high-level visual semantics rather than focusing on optical character recognition, thereby preventing low performance on object-centric and scene-centric downstream zero-shot tasks~\citep{DiHT}.\\

\noindent \textbf{Ranking based on Pretrained VLMs:} One of the most effective pruning methods, CLIPScore~\citep{hessel2021clipscore,schuhmann2021laion}, computes the cosine similarity between image and text embeddings using a pretrained CLIP model. This score is then used to rank the alignment of image-text pairs. LAION filtering~\citep{schuhmann2021laion} employs an OpenAI CLIP model~\citep{clip_openai_radford21a} pretrained on 400 million image-text pairs to evaluate the image-text alignment of large web-scale datasets and filter out samples with the lowest CLIPScore. Inspired by text spotting~\citep{DiHT}, T-MARS \citep{tmars} detects and masks text regions in images before computing the CLIPScore, resulting in a more accurate alignment score. Sieve by \citet{mahmoud2023sieve} demonstrates that false positives and negatives resulting from CLIPScore ranking can be minimized by relying on generative image captioning models pretrained on small but curated datasets.\\

\noindent \textbf{Diversity and Balancing:} Pretraining Vision-Language Models using a diverse and well-balanced dataset can enhance their generalization capabilities~\citep{clip_openai_radford21a}. To create such a dataset, DataComp~\citep{gadre2023datacomp} suggests sampling image-text pairs that are semantically similar to diverse and curated datasets like ImageNet~\citep{imagenet}. Text-based sampling retains image-text pairs whose captions overlap with one of the ImageNet classes. Meanwhile, image-based sampling methods encode noisy web-scale images using the OpenAI CLIP ViT-L/14 vision encoder and cluster the images into 100,000 groups using FAISS~\citep{faiss}. Subsequently, embeddings of ImageNet training samples are used to select the closest cluster to each sample. While this approach can result in a diverse dataset, sampling images semantically similar to ImageNet images could bias the CLIP model, potentially limiting its generalization to new downstream tasks. MetaCLIP~\citep{xu2023metaclip} utilizes 500,000 queries from Wikipedia/WordNet as metadata to create a pretraining data distribution that captures a wide range of concepts. 
Their ``balanced'' sampling algorithm (similar to the one described in \citet{clip_openai_radford21a}) aims to strike a balance between well-represented and under-represented concepts, by limiting the number of samples for each query to 20,000. Nonetheless, collecting a perfectly balanced dataset is impractical due to the natural long-tailed distribution of web data. Consequently, all these CLIP variants still exhibit imbalanced performances across downstream visual concepts~\citep{parashar2024neglected}.
Having a \textbf{wide range of training data concepts seems to be one of the most important components behind the ``zero-shot abilities''} of VLMs. Actually, \citet{udandarao2024zeroshot} demonstrate that the zero-shot performances of VLMs depend mostly on how much those zero-shot downstream concepts are present in the training data.

\subsubsection{Improving the training data with synthetic data}

A line of research focuses on improving the quality of VLM's training data by improving the captions through filtering and synthetic data generation. Specifically, \acrfull{BLIP}~\citep{li2022blip} performs bootstrapping by generating synthetic samples and filtering out noisy captions. Subsequently, in \citet{santurkar2022caption}, authors leverage BLIP to approximate the descriptiveness of a caption and show that models trained on consistent and complete synthetic captions generated by BLIP outperform a model trained on human-written captions. 
\citet{nguyen2023improving} use large image-captioning models like BLIP2~\citep{blip2} to replace poorly aligned alt-text labels with descriptive synthetic captions. They demonstrate that pretraining CLIP with a mixture of real and synthetic captions is effective. However, they also show that at scale, the improvement provided by synthetic captions is capped by the limited diversity of generated captions compared to the high diversity of noisy text labels. More recently, \citet{chen2023pixartalpha} demonstrate that by using \acrfull{LLaVA}~\citep{liu2023visual_tuning,liu2023improved,liu2024llavanext} as captioning model, it is possible to train very efficiently a text-to-image generative model.\\

Inspired by the great progress of large-scale diffusion models \citep{rombach2022high,dai2023emu} and considering the promise of using synthetic image samples in other applications such as classification~\citep{hemmat2023feedback, azizi2023synthetic, bansal2023leaving}, another line of research is to use generated images from text-to-image generative models. \citet{tian2023stablerep} demonstrate improved performance of using synthetic data compared to CLIP~\citep{clip_openai_radford21a} and SimCLR~\citep{chen2020simple} using only synthetic samples. Specifically, they use multiple synthetic samples of the same text prompt as multi-positives pairs for the the contrastive representation learning objective. Furthermore, SynCLR~\citep{tian2023learning} and SynthCLIP~\citep{hammoud2024synthclip} also train a VLM without any real datapoints and only leverage synthetic samples. They use an LLM to generate captions, and then give them to a text-to-image model to generate images based on those captions.

\subsubsection{Using data augmentation}

Can we exploit data augmentation similarly to self-supervised visual models? SLIP~\citep{mu2022slip} addresses this question by introducing an auxiliary self-supervised loss term on the vision encoder. As in SimCLR~\citep{chen2020simple}, the input image is used to generate two augmentations that create a positive pair to be contrasted with all other images in the batch. The overhead of this addition is rather small, while providing a regularization term that improves the learned the representations. However, the use of the SSL loss only for the visual encoder does not fully exploit the important signal coming from text. To this end, CLIP-rocket~\citep{fini2023improved} suggests converting SSL losses to be cross-modal. In particular, it shows that the CLIP contrastive loss can be used in presence of multiple augmentations of the image-text pair, and it is better than other non-contrastive alternatives inspired from SSL, e.g., \citet{grill2020bootstrap}, \citet{caron2020unsupervised}, and \citet{zbontar2021barlow}. In CLIP-rocket, the input image-text pair is augmented in an asymmetrical way, with one weak and one strong set of augmentations. The two resulting augmented pairs are embedded with the standard CLIP encoder and then projected to the multimodal embedding space using two different projectors. The projector of the weakly augmented pair is kept the same as in the original CLIP, i.e., a linear layer, while the projector of the strongly augmented pair is a 2-layer MLP to cope with the noisier embeddings. As highlighted in \citet{bordes2022high} it is crucial to separate the two projectors as the \emph{strong} one is learning more invariant, too invariant, representations for downstream tasks. At inference time, weak and strong learnt representations are interpolated to get a single vector.

\subsubsection{Interleaved data curation}

Autoregressive language models like Flamingo~\citep{alayrac2022flamingo} and MM1~\citep{mckinzie2024mm1} have shown including interleaved text and image data during training improves few-shot performance of the model. The interleaved datasets used for pre-training are usually crawled from the internet and are curated to improve quality and safety. There are two types of curation strategies that can be used to collect interleaved datasets:\\

\textbf{Natural interleaved data}: 
\acrfull{OBELICS}~\citep{laurençon2023obelics} dataset is a good example of this category of datasets; OBELICS is constructed by preserving the intrinsic structure and context in which text and images co-occur within web documents offering a more authentic representation of multimodal web content. Multiple curation steps are used to curate this dataset where English data is collected from common crawl and deduplicated followed by pre-processing HTML document where useful DOM nodes are identified and retained, then for each DOM node we apply image filtering to remove logos followed by a paragraph, and we apply document-level text filtering using various heuristics to handle text that is not well-formed or coherent.\\

\textbf{Synthetic interleaved data}: MMC4~\citep{zhu2023multimodal} is a good example of this type of dataset where text only dataset is retrofitted with images collected from the internet, in this process images are paired with text based on contextual relevance enabled by calculating the CLIP based similarity scores. This method provides a means to retrofit existing vast text corpora with visual information, thereby extending their utility for multimodal learning. While this approach may lack the contextual nuance of naturally interleaved datasets, it allows for the scalable creation of multimodal data from well-established text-only resources.

\subsubsection{Assessing multimodal data quality}

A very active area for research when it comes to VLMs is to identify the quality of the underlying data used to train it. Since quality is a subjective metric, it's hard to determine apriori what qualifies as good data to train these models. Previous works like Flamingo~\citep{alayrac2022flamingo}, MM1~\citep{mckinzie2024mm1}, and OBELICS~\citep{laurençon2023obelics} have demonstrated that high-quality interleaved multimodal data is a critical requirement for obtaining optimal performance for these VLM models which makes it essential to quantify the quality of the data in a fast and scalable manner. The quality itself could be assessed on multiple fronts incorporating the quality of the text itself, the image itself, and the alignment information between the image and text. Methods like QuRating~\citep{wettig2024qurating}, Data efficient LMs~\citep{sachdeva2024train}, and text-quality-based pruning~\citep{sharma2024text} have explored ways to quantify textual data quality and use that to identify high-quality data subsets to train LM models in a data efficient manner. Similarly methods like VILA~\citep{ke2023vila} and LAION-aesthetics~\citep{laion-aesthetics} attempt to quantify the aesthetic quality of an image to select high-quality subsets of image data to improve image generation models. For alignment, the CLIP family of approaches \citep{clip_openai_radford21a,xu2023metaclip,gao2024clip} have been the models of choice to evaluate how coherent the textual data is with respect the provided image. Despite having some relevant work on evaluating text, image, and alignment quality, we lack a holistic way of evaluating the quality of multimodal and interleaved data, which remains an active area of research to further improve training of VLM models.

\subsubsection{Harnessing human expertise: the power of data annotation}

In recent years, the importance of leveraging human data annotation has become increasingly evident in advancing the field of vision-language modeling. This approach involves strategically selecting images and having humans provide labels or descriptions that capture the intricate relationship between visual elements and language. By learning from more subtle and detailed information, models can better comprehend complex scenes and generate more accurate descriptions. Although there are several popular multimodal datasets available, such as OKVQA~\citep{marino2019ok}, A-OKVQA~\citep{schwenk2022okvqa}, Image Paragraph Captioning~\citep{krause2017hierarchical}, VisDial~\citep{das2017visual}, Visual Spatial Reasoning~\citep{liu2023visual}, and MagicBrush~\citep{zhang2024magicbrush}, many of these rely on older image benchmarks like COCO~\citep{lin2014microsoft} or Visual Genome~\citep{krishna2017visual}, which highlights the need for more diverse and contemporary imagery sources. More recently, \citet{urbanek2023picture} introduce the DCI dataset which contains fine-grained human annotations for some images from the SA-1B dataset~\citep{kirillov2023segany}. A limitation of human-annotated data is that it is often costly to get, especially when requesting fine-grained annotations. In consequence, the number of images with highly detailed annotations is often low which makes often those datasets more suited for evaluation or fine-tuning than for large-scale pre-training.

\subsection{Software}

In this section, we discuss some of the existing software that people can leverage to evaluate and train VLMs as well as the resources needed to train them. 

\subsubsection{Using existing public software repositories}

There exist several software such as OpenCLIP (\url{https://github.com/mlfoundations/open_clip}) or transformers (\url{https://github.com/huggingface/transformers}) that implement most VLMs. Those tools are extremely useful when making benchmarks or comparing different models. If one's goal is to try and compare different pre-trained VLM on a given downstream task, then those software provide a good platform to do that.

\subsubsection{How many GPUs do I need?}

The question around the compute resources needed is very important since it will mostly determine the budget one will need to train such model. CLIP~\citep{clip_openai_radford21a} and OpenCLIP~\citep{openclip} have leveraged more than 500 GPUs to train their models. When looking at the public cloud prices for such resources, they are equivalent to hundreds of thousands of dollars which is inaccessible to most companies or academic labs. But, when using the right ingredients such as having a high-quality dataset and leveraging masking strategies when using bigger models, training a contrastive model like CLIP on hundreds of millions of images from scratch should not require more than 64 GPUs (which should be equivalent to spending around 10K USD in compute). If the VLM that is used for training leverages existing pre-trained image or text encoder, or LLM, the cost of learning a mapping should be much lower. 

\subsubsection{Speeding up training}

There were recent software developments such as the introduction of torch.compile by the PyTorch team~(\url{https://pytorch.org/tutorials/intermediate/torch_compile_tutorial.html}) that significantly speed up model training. By using more efficient attention mechanisms, the xformers library~\citep{xFormers2022} is also often used to give an additional speed up. However, there is an area that is often overlooked when training vision models which is data loading. By having to load large mini-batch of images, data loading often becomes a bottleneck that significantly slows down training. In addition, because of space constraint, large-scale datasets are often saved in chunks of tar files that have to be uncompressed on the fly (and thus slowing down training). The main recommendation we have is to store as many uncompressed files as possible to speed up training. In addition, one can leverage the \acrfull{FFCV} library~\citep{leclerc2023ffcv} to create data files that are much faster to load. Using FFCV instead of webdataset can significantly speed up VLM training. The only drawback of storing uncompressed files with either webdataset or FFCV is that the storage might be more costly than storing compressed files. However since the training speed will be much faster, the additional storage cost should be compensated quickly by the lower amount of compute needed.

\paragraph{Masking.} Masking is another way to quickly improve the training efficiency of large models. When using models with hundreds of millions or billions of parameters, the cost of a single forward and backward might be high. \citet{li2023scaling} show that by randomly masking image tokens one can significantly speed up training time while improving model performances.

\subsubsection{Importance of other hyper-parameters.}
\citet{mckinzie2024mm1} study the most important design choices for training VLMs showing image resolution, visual encoder capacity, and visual pretraining data are the choices that most impact model performance. They also show while there are many ways to connect modalities, this choice is much less important. The authors also discuss the importance of various types of training data from text-only data to interleaved and image-caption paired data, demonstrating the right mix achieves the best performance across both zero-shot classification and visual-question answering tasks.

\subsection{Which model to use?}

As highlighted in the first part of this introduction, there are several methods to train VLMs. Some of them leverage simple contrastive training criteria, others use masking strategies to predict missing texts or image patches, while some models are using generative paradigms such as autoregression or diffusion. It is also possible to leverage a pre-trained vision or text backbones like Llama or GPT. In that instance, building a VLM model requires learning only a mapping between the LLM and vision encoder representations. So, from all those methods, which one should someone choose? Do we need to train vision and text encoder from scratch like CLIP or is it better to start with pretrained LLM such as Flamingo or MiniGPT? 

\subsubsection{When to use contrastive models like CLIP?}
Contrastive models like CLIP associate text with visual concepts while keeping a simple training paradigm by pushing text and image representation to be matched in the representation space. By doing so, CLIP learns representations that have both \textit{meaning} in the image and text space, which makes it possible to prompt the CLIP text encoder with words such that we can retrieve the images that map to the corresponding text representations. For example, many data curation pipelines such as MetaCLIP~\citep{xu2023metaclip} are using metadata string matching to build datasets to ensure that each word or concept has enough images associated with them. CLIP models are also a good base for building more complex models, especially when trying to improve grounding. For researchers who are looking at trying additional training criteria or different model architectures to better capture relations or a better understanding of concepts, CLIP is a particularly good starting point. However, one should keep in mind that CLIP is not a generative model, thus it is not possible to generate a caption given a specific image. It is only possible to retrieve the best \textit{caption} within a list of already existing captions. In consequence, current CLIP models cannot be used to provide high-level descriptions of a given image. Another drawback is that CLIP usually needs a very large dataset as well as large batch sizes to offer decent performances, which implies that CLIP usually needs significant resources to be trained from scratch.

\subsubsection{When to use masking?}
Masking is an alternative strategy to train VLMs. By learning to reconstruct data from both masked images and text, it is possible to jointly model their distributions. In contrast to contrastive models which operate in a representation space, models based on masking might need to leverage a decoder to map back the representation to the input space (and thus to apply a reconstruction loss). Training an additional decoder might add an additional bottleneck which might make these methods less efficient than a purely contrastive one. However, the advantage is that there is no batch dependency anymore since each example can be considered separately (because we do not need negative examples). Removing negative examples can enable the use of smaller mini-batches without the need to fine-tune additional hyper-parameters such as the softmax temperature. Many VLM methods leverage a mix of masking strategies along with some contrastive loss.

\subsubsection{When to use a generative model?}
Generative models based on diffusion or autoregressive criteria have demonstrated impressive abilities in generating photorealistic images based on text prompt. Most large-scale training efforts on VLM are also starting to integrate image generation components. Some researchers argue that having the ability to generate images given words is an important step towards creating a good world model while other researchers argue that such a reconstruction step is not needed~\citep{balestriero2024learning}. However from an application perspective, it might be easier to understand and assess what the model has learned when it is able to decode abstract representations in the input data space\footnote{It is also possible to learn a decoder on top of a trained join-embedding architecture~\citep{bordes2022high}.}. While models like CLIP would need extensive $k$-NN evaluations using millions of image data points to show what the images closest to a given word embedding look like, generative models can just output the most probable image directly without such an expensive pipeline. In addition, generative models can learn an implicit joint distribution between text and images which might be more suited for learning good representations than leveraging pretrained unimodal encoders. However, they are more computationally expensive to train than their contrastive learning counterpart.

\subsubsection{When to use LLM on pretrained backbone?}
Using already pretrained text or vision encoder can be a good alternative when having access to limited resources. In that case, only the mapping between the text representation and vision representation should be learned. However, the main issue with this approach is that the VLM will be impacted by the potential hallucination of the LLM. It could also be impacted by any bias coming from the pretrained models. In consequence, there might be an additional overhead in trying to correct the defect of the vision model or of the LLM. Some might argue that it is important to leverage independent image and text encoder to project the information into a lower dimension manifold on which we can learn a mapping while others might argue that it is important to learn the distribution of image and text jointly. To summarize leveraging a pre-trained model is interesting when having limited access to compute resources and when researchers are interested in learning mapping in representation spaces.

\subsection{Improving grounding}
Grounding is an important challenge in the VLM and generative model literature. It mostly aims to solve the problem of models not understanding well the text prompt which could either lead to ignoring some part of the prompt or to hallucinating something that is not even part of the prompt. Some of those challenges are related to understanding relations such as an object being on the left or right, negations, counting, or understanding attributes (such as colors or textures). Improving grounding is an active area of research and for now there isn't a single simple method that can solve that. Nevertheless, in this section, we present some of the tricks that are typically used to improve grounding performances.

\subsubsection{Using bounding boxes annotations}
Models like X-VLM~\citep{zeng2021multi} leverage bounding box annotations and incorporate box regression and \acrfull{IoU} loss to accurately locate and align visual concepts with their corresponding textual descriptions. By knowing where the objects are on the images and what are the captions associated with each object, it is easier for the model to associate text to the right visual clues, and thus improve grounding. 
X-VLM is trained on a comprehensive collection of datasets, including COCO~\citep{lin2014microsoft}, Visual Genome~\citep{krishna2017visual}, SBU, and Conceptual Captions~\citep{changpinyo2021conceptual}, amassing up to 16 million images. This extensive training catalog of data with bounding boxes annotations enables X-VLM to outperform existing methods across a variety of vision-language tasks such as image-text retrieval, visual reasoning, visual grounding, and image captioning.\\ 

Instead of using already annotated data, some methods like Kosmos-2~\citep{peng2024grounding} rely on public models to create their own image-text datasets. They make a web-scale grounded image-text pairs from web-crawl data by first extracting the nouns from the text captions using spaCy~\citep{spacy2} and then use the grounded model GLIP~\citep{li2022grounded} to predict bounding boxes associated with the nouns extracted from the captions. Then they use spaCy to extract the expression associated with a given words such that to produce captions that can be associated with each of the bounding boxes that have been detected. Doing so enable the use of very large-scale web-annotated datasets. However such an approach is limited by how strong the grounding model for bounding box detection is. It is likely that if this base model fails on some rare nouns or instances, the downstream model would make similar mistakes. 

\subsubsection{Negative captioning}
Negative samples within the realm of contrastive objectives have been extensively used to mitigate collapse, enhance generalization, and discriminative feature learning~\citep{chen2020simple, liu2023improved, grill2020bootstrap, he2020momentum, caron2021emerging}. By contrasting positive pairs (similar or related samples) with negative pairs (dissimilar or unrelated samples), models are forced to develop a nuanced understanding of the data, going beyond mere superficial features to grasp the underlying patterns that distinguish different classes or categories.\\

In the same vein, recent works on VLMs have shown that similar techniques (negative samples) can be adopted to mitigate various problems in vision-language models~\citep{aro_yuksekgonul2023when, li2021supervision, goel2022cyclip, radenovic2023filtering}. For instance, the ARO benchmark~\citep{aro_yuksekgonul2023when} evaluates VLMs on their ability to correctly associate images with captions, using negative samples to test the model's understanding of incorrect or nonsensical pairings. This approach has demonstrated that VLMs can significantly benefit from the nuanced differentiation capabilities fostered by exposure to negative samples, leading to more accurate and contextually aware models.

\subsection{Improving alignment}

Motivated by the success of instruction tuning in the language domain \citep{chung2022scaling}, vision-language models have also begun to incorporate instruction-fine-tuning and \acrfull{RLHF} in vision-language models to improve multimodal chat capabilities and align outputs with desired responses.\\

Instruction-tuning involves fine-tuning a vision-language model on supervised data containing instructions, inputs, and the desired response. Typically instruction tuning datasets are much smaller compared to pretraining data---with instruction tuning data sizes ranging from a few to one hundred thousand samples (see \citet{li2023visionlanguage} for further discussion of instruction tuning). LLaVa, InstructBLIP \citep{liu2023visual_tuning}, and OpenFlamingo~\citep{awadalla2023openflamingo} are three prominent vision-language models that incorporate instruction tuning.\\

RLHF also aims to align model outputs with human preferences. 
For RLHF a reward model is trained to match human preferences for what humans consider a good or bad model response. 
While instruction tuning requires supervised training samples, which can be costly to gather, RLHF takes advantage of an auxiliary reward model to mimic human preferences.
The primary model, whether a language-only or a vision-language model, is then fine-tuned with the reward model to align outputs with human preferences. LLaVa-RLFH is one prominent example of vision-language models incorporating RLHF to improve model output alignment with factual information \citep{sun2023aligning}.

\subsubsection{A LLaVA story}

Motivated by the success of instruction tuning in the language domain, \acrshort{LLaVA}~\citep{liu2023visual_tuning} was among the first models to incorporate instruction-fine-tuning in vision-langauge models to improve multimodal chat capabilities. 
The authors generate 150k synthetically generated visual instruction samples for fine-tuning. The original LLava model incorporates a pretrained Vicuna language model encoder and a pretrained CLIP ViT-L/14 vision encoder. The encoder outputs are fused into the same dimensional space with a linear projector. Along with improved qualitative chat interactions, LLaVA also shows improvements on synthetic instruction following and Science QA benchmarks~\citep{lu2022learn}.

\paragraph{LLaVA 1.5.} \citet{liu2023improved} improves on LLava's instruction fine-tuning by using a cross-modal fully connected multi-layer perceptron (MLP) layer and incorporating academic VQA instruction data. LLava 1.5 is trained on 600k image-text pairs making it much more efficient to train compared to other instruction-tuned models such as InstructBLIP or Qwen-VL. Training takes approximately one day on 8-A100 GPUs. LLava 1.5 performs well on a suite of academic VQA and instruction-following benchmarks.

\paragraph{LLaVA-RLHF.}
Due to the scarcity of high-quality visual instruction tuning data for vision language model training, VLLMs such as LLaVA~\citep{liu2023visual_tuning} may misalign the vision and language modalities and generate hallucinated outputs. To address this issue, LLaVA-RLHF~\citep{sun2023aligning} was proposed to improve multimodal alignment with a novel \acrshort{RLHF} algorithm, Factually Augmented RLHF. The idea is based on adapting RLHF from text domain to vision-language task and augmenting the reward model with extra factual information of image captions and ground-truth multi-choice to reduce reward hacking. LLaVA-RLHF also uses GPT4-generated training data and human-written image-text pairs for further improving its general capabilities. On LLaVA-Bench, LLaVA-RLHF achieves 94\% performance level of GPT-4~\citep{openai2024gpt4}. On MMHAL-BENCH with a special focus on penalizing hallucinations, LLaVA-RLHF outperforms baselines by 60\%.

\paragraph{LLaVA-NeXT (v1.6).}
LLaVA-NeXT~\citep{liu2024llavanext} improves over LLaVA-v1.5 on several fronts. First, the image resolution is increased by concatenating visual features from the full image and smaller image patches, which are separately fed through the vision encoder. Second, the visual instruction tuning data mixture is improved with better visual reasoning, OCR, world knowledge, and logical reasoning examples. Third, the largest model variant uses a 34B-parameter LLM backbone (\texttt{Nous-Hermes-2-Yi-34B}). LLaVA-NeXT achieves state-of-the-art performance compared to open-source multimodal LLMs such as CogVLM~\citep{hong2023cogagent, wang2023cogvlm} or Yi-VL~\citep{ai2024yi}, and closes the gap with commercial models such as Gemini Pro~\citep{geminiteam2024gemini}.

\subsubsection{Multimodal in-context learning}
Otter~\citep{li2023otter} shows that \emph{multimodal in-context learning} is possible: A few examples (e.g., instruction-image-answer tuples) are provided as the context, and the model could successfully follow instructions in the test examples without extra fine-tuning. This ability is analogous to text-only LLM in-context learning. The multimodal in-context learning ability can be attributed to fine-tuning on the newly proposed multimodal instruction tuning dataset MIMIC-IT~\citep{li2023mimicit} that contains around 2.8M multimodal instruction-response pairs with in-context examples. Each sample in MIMIC-IT contains in-context instruction-image-answer tuples  as well as a test example (where given the instruction and an image, the goal is to generate the answer in the test example). The in-context tuples are relevant to the test example in one of the three ways: (1) the in-context instructions are similar but the images are different; (2) the images are the same but the instructions are different; (3) the images are in a sequential fashion but the instructions are different, where the sequential images are taken from video repositories like \citet{yang2023panoptic}. Fine-tuning OpenFlamingo~\citep{awadalla2023openflamingo} on MIMIC-IT results in the model Otter, and Otter exhibits stronger instruction following ability as well as multimodal in-context learning ability. 

\subsection{Improving text-rich image understanding}
Understanding text is a crucial aspect of visual perception in our daily lives. The success of \acrfull{MLLMs} paved the way for the ability to handle extraordinary applications of VLMs in zero-shot tasks transferred to many real-world scenarios. \cite{liu2024} show that MLLMs exhibit excellent zero-shot \acrfull{OCR} performance in the wild, without explicitly training on the OCR domain-specific data. However, these models often struggle with interpreting texts within images when presented with complex relationships between the datatypes, possibly due to the prevalence of natural images in their training data (for instance, Conceptual Captions ~\citep{changpinyo2021conceptual} and COCO~\citep{lin2014microsoft}. Some common, non-exhaustive challenges with text understanding and models tackling them:\\

\textbf{Instruction tuning with fine-grained text-rich data : LLaVAR \citep{zhang2024llavar}}
To address issues with comprehending textual details within an image, LLavaR enhances the current visual instruction tuning pipeline with text-rich images such as movie posters and book covers. The authors used publicly available OCR tools to collect results on 422K text-rich images from the LAION dataset \citep{schuhmann2022laion5b}. They then prompted text-only GPT-4 \citep{openai2024gpt4} with recognized text and image captions to generate 16K conversations, each containing question-answer pairs for text-rich images. By combining this collected data with previous multimodal instruction-following data, the LLaVAR model was able to substantially improve the capability of the LLaVA model \citep{liu2023visual_tuning}. with up to a 20\% accuracy improvement on text-based VQA datasets and a slight improvement on natural images.\\

\textbf{Dealing with fine-grained text in high resolution images : Monkey~\citep{li2024monkey}}
Currently, most MM-LLMs have their input images limited to a resolution of 224 x 224, consistent with the input size of the visual encoder used in their architecture.
These models struggle to extract detailed information in complex text-centric tasks such as Scene Text-Centric Visual Question Answering (VQA), Document-Oriented VQA, and Key Information Extraction (KIE) with high-resolution input and detailed scene understanding. To address these challenges, a new approach, Monkey~\citep{li2024monkey}, has been introduced.\\

Monkey's architecture is designed to enhance the capabilities of LLMs by processing input images in uniform patches using a sliding window method, each matching the size used in the original training of the well-trained vision encoder.  Each patch is processed independently by a static visual encoder, enhanced with LoRA adjustments and a trainable visual resampler. This allows Monkey to handle higher resolutions up to 1344×896 pixels, enabling the detailed capture of complex visual information. It also employs a multi-level description generation method, enriching the context for scene-object associations. This two-part strategy ensures more effective learning from generated data. By integrating the unique capabilities of these systems, Monkey offers a comprehensive and layered approach to caption generation, capturing a wide spectrum of visual details.\\

\textbf{Decoupled Scene Text Recognition Module and MM-LLM : Lumos~\citep{shenoy2024lumos}}
Lumos proposes a multimodal assistant with text understanding capabilities that leverages a combination of on-device and cloud computation.
Lumos uses a decoupled \acrfull{STR} module which then feeds into the multimodal LLM. Lumos' STR module contains four sub-components: \acrfull{ROI} detection, Text detection, Text recognition, and Reading-order reconstruction. ROI detection effectively detects salient areas in the visual, and then crops the salient area as STR input. Text detection takes the cropped image from ROI detection as input, detects words, and outputs the identified bounding box coordinates for each word. Text recognition takes the cropped image from ROI detection and the word bounding box coordinates from Text detection as input, and returns the recognized words. Reading-order reconstruction organizes recognized words into paragraphs and in reading order within each paragraph based on the layout.\\

The cloud hosts a multimodal LLM module, which takes in the recognized text and coordinates from the STR module. This decoupled STR module can be run on-device, reducing power and latency from transferring high-resolution images to the cloud. As mentioned above, one of the key challenges has been capturing fine-grained text from  the scene due to limitations of LLM’s encoders. Lumos’s STR module works on 3kx4k sized image which would yield enhanced performance in complex text understanding tasks similar to Monkey. 

\subsection{Parameter-Efficient Fine-Tuning}
Training VLMs has shown great effectiveness in cross-domain vision and language tasks. However, as the size of pre-trained models continues to grow, fine-tuning the entire parameter set of these models becomes impractical due to computational constraints. To address this challenge, \acrfull{PEFT} methods have been developed to address the high computational cost associated with fine-tuning large-scale models. These methods focus on training a subset of parameters, rather than the entire model, to adapt to downstream tasks. Existing PEFT methods can be categorized into four main groups, namely \acrfull{LoRa} based methods, Prompt-based methods, Adapter-based methods, and Mapping-based methods.\\

\textbf{LoRA-based methods.}
LoRA~\citep{hu2021lora} is recognized as a popular method for parameter fine-tuning. LoRA can be applied to both pure language models and vision-language models. Several variants of LoRA have been developed to enhance its functionality and efficiency. One such variant is QLoRA~\citep{dettmers2024qlora}, which integrates LoRA with a quantized backbone and enables the back-propagation of gradients through a frozen, 4-bit quantized pre-trained language model into LoRA.
Another variant is VeRA~\citep{kopiczko2023vera}, which is designed to reduce the number of trainable parameters in comparison to LoRA, while maintaining equivalent performance levels. This is achieved by utilizing a single pair of low-rank matrices shared across all layers and learning small scaling vectors instead. Lastly, DoRA~\citep{liu2024dora} decomposes the pre-trained weight into two components, magnitude and direction, for fine-tuning. DoRA has demonstrated the capability to generalize Low-rank adaptation methods from language models to Vision-Language benchmarks through empirical experiments.\\

\textbf{Prompt-based methods.}
The process of vision-language pre-training involves the alignment of images and texts within a shared feature space, enabling zero-shot transfer to subsequent tasks via prompting. Consequently, another method for efficient fine-tuning is linked with prompting.
\citet{zhou2022learning} introduce Context Optimization (CoOp), a technique designed to adapt large pre-trained vision-language models, such as CLIP, for downstream image recognition tasks, eliminating the need for manual prompt engineering. CoOp optimizes the context words of the prompt using learnable vectors during the training process. The method provides two implementations: unified context and class-specific context.
Experimental results from 11 datasets indicate that CoOp outperforms hand-crafted prompts and linear probe models in few-shot learning. Additionally, it exhibits superior domain generalization capabilities compared to zero-shot models that utilize manual prompts.
Then \citet{jia2022visual} present Visual Prompt Tuning (VPT), for adapting large-scale Transformer models in vision. Contrary to the conventional approach of full fine-tuning, which updates all backbone parameters, VPT introduces a minimal amount (less than 1\% of model parameters) of trainable parameters in the input space. This is achieved while keeping the model backbone frozen, and in many instances. VPT demonstrates comparable or even superior accuracy to full fine-tuning. \\

\textbf{Adapter-based methods.}
Adapters refer to new modules added between layers of a pre-trained network~\citep{houlsby2019parameter}. Specifically, in the vision-language model domain, CLIP-Adapter~\citep{gao2024clip} fine-tunes with feature adapters on either the visual or language branch. It adopts an additional bottleneck layer to learn new features and performs residual-style feature blending with the original pre-trained features. In addition, VL-adapter~\citep{sung2022vl} evaluates various adapter-based methodologies, within a unified multi-task framework across a diverse range of image-text and video-text benchmark tasks. The study further delves into the concept of weight-sharing between tasks as a strategy to augment the efficiency and performance of these adapters. Empirical results indicate that the application of the weight-sharing technique in conjunction with adapters can effectively rival the performance of full fine-tuning, while necessitating updates to only a minimal fraction of the total parameters (4.18\% for image-text tasks and 3.39\% for video-text tasks).
Subsequently, LLaMA-Adapter V2~\citep{gao2023llama} proposes a parameter-efficient visual instruction model that enhances large language models' multi-modal reasoning capabilities without requiring extensive parameters or multi-modal training data. It proposes unlocking more learnable parameters (e.g., norm, bias, and scale) and an early fusion method to incorporate visual tokens into LLM layers. Compared to other full-fine-tuning approaches like MiniGPT-4 and LLaVA, LLaMA-Adapter V2 involves much fewer additional parameters. \\

\textbf{Mapping-based methods.}
Injecting trainable modules into pretrained models through adapters or LoRA requires some knowledge of the network's architecture to decide where to insert or adapt parameters. In the context of VLMs, \citet{manas2023mapl} and \citet{merullo2022linearly} propose a simpler approach which only requires training a mapping between pretrained unimodal modules (i.e., vision encoders and LLMs), while keeping them completely frozen and free of adapter layers. In addition, this method requires fewer trainable parameters and leads to increased data-efficiency~\citep{vallaeys2024improved}. LiMBeR~\citep{merullo2022linearly} uses a linear layer that projects visual features to have the same LLM hidden state dimension. This projection is independently applied to each feature vector, which means the length of the sequence passed to the LLM is the same as the number of visual feature vectors, increasing the computational cost of training and inference. MAPL~\citep{manas2023mapl} designs a mapping network which addresses this issue by aggregating the visual feature vectors into a smaller set. The input feature vectors are projected and concatenated to a sequence of learnable query tokens, and only the outputs of the query tokens are fed to the LLM.

\section{Approaches for Responsible VLM Evaluation}

As the main ability of VLMs is to map text with images, it is crucial to measure visio-linguistic abilities so as to ensure that the words are actually mapping to visual clues. Early tasks used to evaluate VLMs were image captioning and \acrfull{VQA}~\citep{VQA}. In this section, we also discuss the task of text-centric VQA that assesses the ability of the model to understand and read text from images. Another common evaluation introduced by \citet{clip_openai_radford21a} is based on zero-shot predictions such as the ImageNet~\citep{imagenet} classification task. Such classification tasks are important to assess if a VLM has a good enough knowledge of the world. More recent benchmarks such as Winoground~\citep{thrush2022winoground} measure visio-linguistic compositional reasoning. Since VLM models are known to display biases or hallucinations, it is important to assess those two components. 

\begin{figure}
    \centering
    \includegraphics[width=0.9\linewidth]{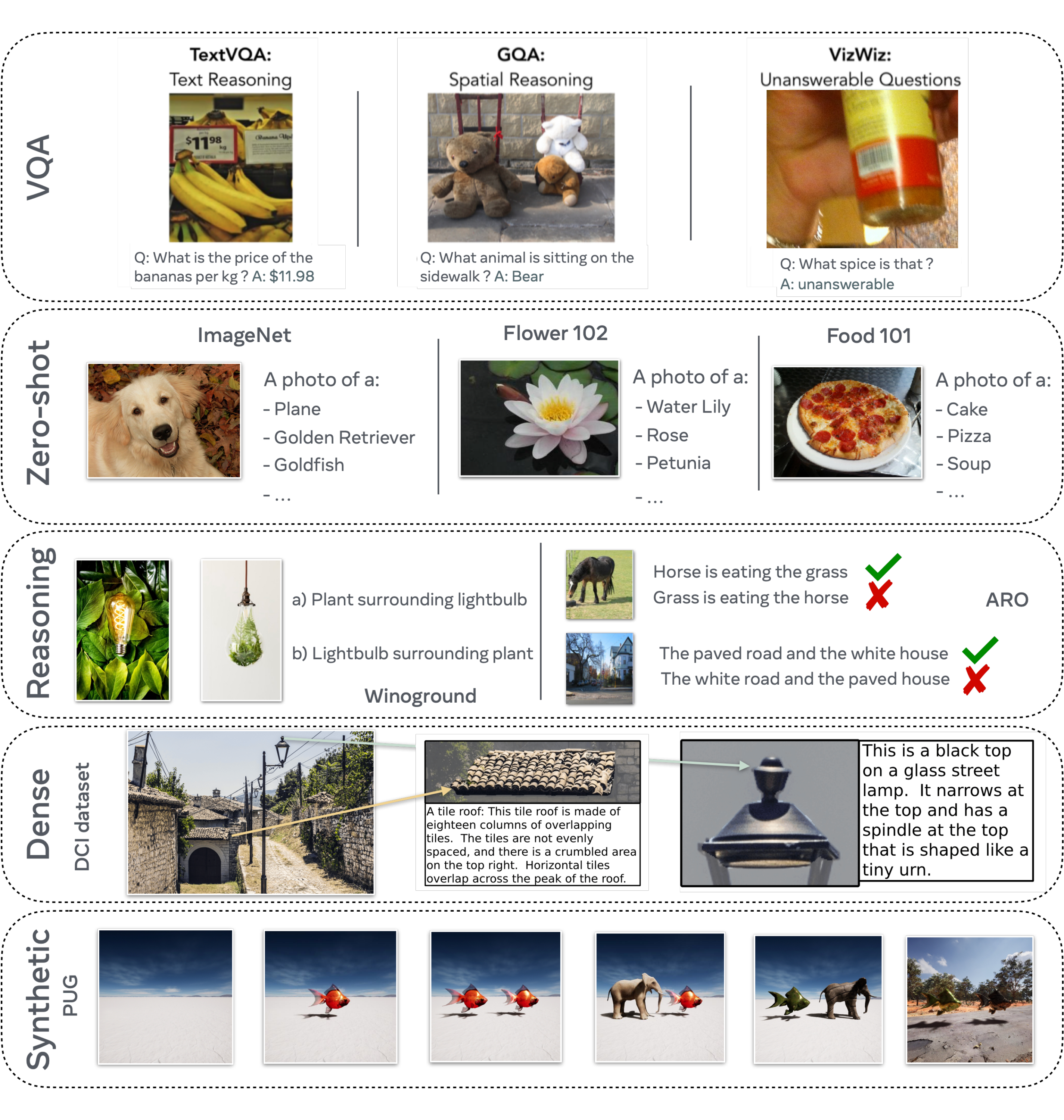}
    \caption{Different methods to evaluate VLMs. \textbf{Visual Question Answering (VQA)} has been one of the most common methods, though the model and ground truth answers are compared via exact string matching, which may underestimate the model performance. \textbf{Reasoning} consists of giving VLMs a list of captions and making it select the most probable one within this list. Two popular benchmarks in this category are Winoground~\citep{diwan2022winoground} and ARO~\citep{aro_yuksekgonul2023when}. More recently, \textbf{Dense} human annotations can be used to evaluate how well the model is able to map the captions to the correct parts of an image~\citep{urbanek2023picture}. Lastly, one can use synthetic data like PUG~\citep{bordes2023pug} to generate images in different configurations to evaluate VLM robustness to specific variations.}
    \label{fig:eval_summary}
\end{figure}

\subsection{Benchmarking visio-linguistic abilities}

A first way to evaluate VLMs is to leverage visio-linguistic benchmarks. These are designed in such a way to assess whether the VLMs are able to associate specific words or phrases with the corresponding visual clues. These benchmarks are at the forefront of VLM evaluation since they assess how well a visio-linguistic mapping is learned. From visual question answering to zero-shot classification, there are many methods that are often used to evaluate VLMs. Some of them focus on the detection of simple visual clues such as ``Is a dog visible in the image?'' to much more complex scenes in which we would try to assess whether the VLM is able to give the correct answer to questions such as ``How many dogs are in the images, and what are they looking at?'' By starting from simple captions that highlight clear visual clues to more complex captions that require some level of spatial understanding and reasoning, these benchmarks allow us to assess the strengths and weaknesses of most VLMs. 

\subsubsection{Image captioning}
 Introduced by \citet{COCO_eval}, the COCO captioning dataset and challenge evaluate the caption quality generated by a given VLM. By leveraging an external evaluation server, researchers could send the caption generated by their models and have them evaluated by the server that used scores like BLEU~\citep{papineni2002bleu} or ROUGE~\citep{lin-2004-rouge} to compare the generated caption to a set of reference captions. However, such scores are still heuristics that try to approximate the similarity of those captions. Many works such as \citet{mathur-etal-2020-tangled} have advocated for the retirement of scores like BLEU. \\

To avoid the issue of having to compare a caption with a set of reference captions, \citet{hessel2021clipscore} introduce the CLIPScore that leverages CLIP to predict how close a caption is to an image. The higher the score, the more likely the caption is to actually describe the image content. However, there is a significant limitation of CLIPScore which is the underlying performances of the CLIP model used.

\subsubsection{Text-to-image consistency}
In addition to evaluating the ability to generate a caption for a given image, one might also want to evaluate the ability to generate an image given a  caption.
There are end-to-end approaches that use a single model to produce a consistency score. Though it was initially proposed for image captioning, CLIPScore is also used in image generation to measure the alignment between a generated image and a text prompt.
\citet{lin2024evaluating} and \citet{li2024evaluating} apply another approach that formats the text prompt as a question (e.g., ``Does this figure show \{text caption\}'') and gets the probability of a VQA model answering \textit{yes}.
There are also a series of metrics that leverage a \acrfull{LM} to generate questions given a text caption. TIFA \citep{hu2023tifa} and Davidsonian Scene Graph (DSG) \citep{cho2023davidsonian} both use an \acrshort{LM} to generate natural language binary and multiple choice questions, and a \acrfull{VQA} model to evaluate the questions. DSG additionally addresses hallucinations in LLMs and VLMs -- the generated questions are organized into a scene graph based on their dependencies and a question is counted as correct if and only if the questions it depends on are also correct. For example, assume a VQA model is given the questions ``Is there a car?'', ``What color is the car?'' and ``How many wheels does the car have?''. If the model incorrectly answers \textit{``no''} to the first question, the rest of the questions are deemed incorrect regardless of their answers because the model did not recognize the car. VPEval \citep{cho2023visual} is another metric that also generates questions but instead of being in natural language, the questions are visual programs. These visual programs are executable by different visual modules, such as a counting module, a VQA module or an \acrfull{OCR} module. \citet{lin2024evaluating} and \citet{li2024evaluating} introduce VQAScore, another VQA-based method for text-to-image evaluation. Instead of generating questions using an LM, they instead take the text prompt and pass that directly to a VQA model. For instance, given a prompt \textit{a red dog next to blue flower}, VQAScore computes the probability of a VQA model generating \textit{yes} given the question \texttt{Does this figure show a red dog next to a blue flower?}.

\subsubsection{Visual question answering}
\acrfull{VQA} is the task of answering natural language questions about images. Due to its simplicity and generality, VQA is one of the main tasks used to evaluate VLMs. In fact, most VLM tasks can be reformulated as VQA (e.g., ``what is in the image?'' for captioning, ``where is this?'' for phrase grounding, etc.). The task was originally proposed~\citep{VQA} in two flavors: multiple-choice and open-ended answers. Popular benchmarks based on the VQA task include VQAv2~\citep{vqa2}, TextVQA~\citep{singh2019towards}, GQA~\citep{hudson2019gqa}, Visual Genome QA~\citep{krishna2017visual}, VizWiz-QA~\citep{gurari2018vizwiz}, OK-VQA~\citep{marino2019ok}, ScienceQA~\citep{lu2022learn}, MMMU~\citep{yue2023mmmu} (see Figure~\ref{fig:eval_summary}). VQA is traditionally evaluated with VQA Accuracy, which is based on exact string match between a candidate answer generated by a model and a set of reference answers annotated by humans. This metric has worked well so far in the multiple-choice and IID training settings. However, the community is transitioning towards generative models (capable of generating free-form, open-ended answers) and OOD evaluation (e.g., zero-shot transfer). In these new settings, the traditional VQA Accuracy metric is too stringent and tends to underestimate the performance of current VLM systems \citep{agrawal2023reassessing}. To overcome this limitation, some works have resorted to artificially constraining~\citep{blip2} or rephrasing~\citep{awal2023investigating} the output of VLM to match the format of reference answers. However, this precludes a fair comparison among VLM as their perceived performance is largely dependent on answer formatting tricks. To enable a truthful and fair evaluation of VLM, \citet{manas2023improving} propose to leverage LLMs as judges for VQA.\\

\textbf{Selective prediction.} Besides answer correctness, another dimension of evaluation is \textit{selective prediction} for VQA -- how well a VLM can abstain from answering questions it would otherwise get incorrect, and achieve high accuracy on questions it chooses to answer. This is important for applications where accuracy is critical, and incorrect answers could mislead users who place trust in the model.
\citet{whitehead2022reliable} formalize this framework for VQA, defining evaluation in terms of coverage (the fraction of questions answered) at a specified risk (level of error tolerated), as well as a cost-based metric (Effective Reliability) that penalizes incorrect answers more than abstentions. The decision to abstain can be determined by thresholding an uncertainty measure, such as the answer probability directly, a learned correctness function \citep{whitehead2022reliable,dancette2023improving}, or agreement among expert models (e.g., \citet{Si:Shi:Zhao:Zettlemoyer:Boyd-Graber-2023} in the unimodal language space).\\

\textbf{Visual Dialog.} \citet{das2017visual} introduced VisDial, a dataset and benchmark that extends VQA by using a series of questions about an image. Its goal is to measure the ability of an agent to hold a discussion about a given image. In contrast to traditional VQA in which questions can be considered as independent, visual dialog benchmarks evaluate more general intelligence abilities such as being able to understand context from the discussion history.

\subsubsection{Text-centric Visual Question Answering}
Text-Based VQA is a task that involves providing responses to natural language inquiries about textual content in the image. Beyond understanding the correlation between textual and visual content in generic VQA, these queries require the model to 1) read the text in the scene accurately and devise how it’s structured and ordered and 2) reason about the text in the image in correlation to each other as well as other visual elements in the image.
Text-centric evaluations can be done using a broad spectrum of tasks like Text Recognition, Scene Text-Centric Visual Question Answering (VQA), Document-Oriented VQA, Key Information Extraction (KIE), and Handwritten Mathematical Expression Recognition (HMER). Each of these tasks presents unique challenges and requirements, providing a comprehensive overview of the capabilities and limitations of the LLMs.\\

Text Recognition is a fundamental task in \acrfull{OCR}, requiring the model to accurately identify and transcribe text from a variety of sources. Scene Text-Centric VQA extends this challenge by requiring the model to not only recognize text within a scene, but also to answer questions about it. Document-Oriented VQA further complicates this by introducing structured documents, such as forms and invoices, into the mix. KIE is a task that focuses on extracting key pieces of information from a document, such as names, dates, or specific values. Finally, HMER is a specialized task that involves recognizing and transcribing handwritten mathematical expressions, a particularly challenging task due to the complexity and variability of handwritten notation. Some popular benchmarks include IIIT5K \citep{mishra2012}, COCOText \citep{veit2016cocotext}, SVT \citep{svt}, IC13 \citep{IC13} for text recognition, STVQA \citep{biten2019icdar}, Text VQA \citep{singh2019towards}, OCR VQA \citep{mishraICDAR19} and EST VQA \citep{wang2020general} for scene text-centric VQA, DocVQA \citep{mathew2021docvqa}, Info VQA \citep{mathew2021infographicvqa} and ChartQA \citep{masry2022chartqa} for document-oriented VQA, SROIE \citep{Huang_2019}, FUNSD \citep{jaume2019funsd} and POIE \citep{kuang2023visual} for KIE and HME100k \citep{yuan2022syntaxaware} for HMER. The composition of the datasets varies widely and should be chosen primarily based on purpose of the evaluation -- some focus on specific types of text (such as handwritten or artistic text),  while others include a mix of text types. Some datasets were specifically designed to challenge the models' ability to handle multilingual text, handwritten text, non-semantic text, and mathematical expression recognition. Some datasets purely focus on a plethora of different infographics and tabular representations. 

\subsubsection{Zero-shot image classification}
Zero-shot classification consists of evaluating a model on a classification task for which the model was not explicitly trained. This should be contrasted with few-shot learning which requires few training data samples of the downstream task of interest for model fine-tuning.
~\citet{clip_openai_radford21a} demonstrate that zero-shot classification performance of CLIP can be significantly improved with different types of prompt structures, especially when customized for specific tasks. They were able to show competitive performances on the well-known ImageNet classification benchmark~\citep{imagenet}. This was the first work to show that VLM approaches might be able to compete with standard classification training. In addition to ImageNet, it is standard to evaluate VLMs on additional classification datasets such as CIFAR10/100~\citep{krizhevsky2009learning}, Caltech 101~\citep{li_andreeto_ranzato_perona_2022}, Food101~\citep{food101}, CUB~\citep{WahCUB_200_2011}, StanfordCars~\citep{StanfordCars}, Eurosat~\citep{Eurosat}, Flowers102~\citep{Flower102}, OxfordPets~\citep{OxfordPet}, FGVC-Aircraft~\citep{aircraft} and Pascal VOC~\citep{pascal_voc}.\\

Since prompt engineering, e.g., using concept names within human-engineered prompt templates, such as ``a photo of a \{{\tt class}\}'' and 
``a demonstration of a \{{\tt class}\}'', can substantially enhance zero-shot performance, recent studies introduce novel approaches~\citep{menon2022visual, pratt2023does, parashar2023prompting} that employ LLMs like ChatGPT to automatically generate prompts, often with rich visual descriptions, e.g., ``a {\tt tiger}, which has sharp claws''. While these methods adopt label names as originally written by CLIP~\citep{clip_openai_radford21a}, \citet{parashar2024neglected} substitutes these names with their most frequently used synonyms (e.g., replacing {\tt cash machine} with  {\tt ATM}) to improve accuracy, irrespective of the prompt templates employed. As highlighted by \citet{udandarao2024zeroshot}, zero-shot abilities of a VLM depend mostly on whether those concepts are present or not in the training data. Thus, it is not clear whether we should still consider such evaluations as zero-shot since the model might be already trained in some indirect way to solve the downstream task. 

\paragraph{Generalization on \acrfull{OOD} tasks.} Using zero-shot evaluation for CLIP on tasks like ImageNet and achieving good performances is only possible because the CLIP training data is large enough that it may contain much of the concepts and class labels that are present in the ImageNet dataset. In consequence, when there are some downstream tasks for which the training CLIP distribution might be too different, it can lead to poor generalization. \citet{samadh2023align} suggests modifying the token distribution of test examples such that they align with ImageNet data distribution (since the original CLIP training data is unknown). They show that such alignment can help improve performances on various OOD benchmarks as well as on different downstream tasks. 

\subsubsection{Visio-linguistic compositional reasoning}
Several recent benchmarks introduce artificially created captions that are designed with ambiguity to attack the model. One easy way to create such captions can be by reordering the words in the ground-truth caption. Then the model is evaluated on its ability to discriminate the correct caption from the perturbed one (which makes this evaluation equivalent to a binary classification problem). In this section, we are presenting some of the benchmarks that are often used that leverage such binary classification setups.\\

Winoground~\citep{thrush2022winoground} is a task for evaluating the visiolinguistic abilities of VLMs. Each example in the dataset contains two images and two captions. Each image matches exactly one caption, with the captions differing only in their word order. For example, in Figure \ref{fig:eval_summary}, there are two captions \textit{``some plants surrounding a lightbulb''} and \textit{``a lightbulb surrounding some plants''}.
Models are tasked with scoring the correct image-caption pairs higher than the incorrect pairs. \citet{diwan2022winoground} additionally explore Winoground and provide insight on why this task is so challenging for VLMs.\\

More recently, \acrfull{ARO} was introduced by \citet{aro_yuksekgonul2023when} to assess relation, attribute, and order understanding by VLMs. The dataset was built using GQA, COCO and Flickr30k. Then, negative captions were generated by swapping either the relations, attribute or order from the original caption. By doing so, a caption describing ``A horse eating grass'' becomes ``grass eating a horse'' (Figure \ref{fig:eval_summary}). Then the model is evaluated on its ability to predict a lower probability to the negative caption. In contrast to Winoground that finds real images that correspond to the negative caption, ARO does not come with true ``negative'' images. Such an approach has the advantage that it is possible to generate a lot of negative captions; however, some of them might not make any sense in the real world.\\

\citet{sugarcrepe} have observed that recently developed image-to-text retrieval benchmarks \citep{aro_yuksekgonul2023when,vlchecklist,crepe}, which are designed to assess the detailed compositional abilities of VLMs, can be manipulated. These benchmarks indeed depend on procedurally-generated hard negatives that often lack logical coherence or fluency due to grammatical inaccuracies. To mitigate these issues, \citet{sugarcrepe} instead suggest leveraging ChatGPT to generate more plausible and linguistically correct hard negatives. They have divided the SUGARCREPE dataset~\citep{sugarcrepe}, similar to the approach taken in ARO to evaluate different forms of hard-negatives, each measuring a specific compositional aspect (e.g., attribute, relationship, object understanding).

\paragraph{Warning!} A major issue with many of the benchmarks relying on the binary classification problem of discriminating the correct caption from the negative one is that they often do not consider the case in which the model outputs an equal probability for both captions. This can occur if the model collapses the information to the same representation vector for both captions. If the model outputs the same probabilities, then the argmax operation used by frameworks like PyTorch will always return the first element of the vector. It happens that many benchmarks put the correct caption as the first element. Thus, a model whose parameters are all equal to zero could achieve 100\% accuracy in these benchmarks. We recommend adding a small epsilon random number or keeping track if the captions are assigned the same probabilities.

\subsubsection{Dense captioning and crop-caption matching}
The current generation of VLMs is often limited to short text descriptions as input due to text tokenizers. The popular Clip Tokenizer (used to train CLIP-based models) generates only a maximum of 77 tokens, equivalent to fifty English words or a small paragraph. Even if it is possible to summarize an image with few words, images are often much richer than that. When using short captions, we lose information about the background and the fine-grained specifics of the object we want to describe. The \acrfull{DCI} dataset~\citep{urbanek2023picture} was introduced to provide complete image descriptions. By dividing an image into distinct parts using Segment Anything~\citep{kirillov2023segany}, the authors asked human annotators to provide detailed descriptions for each segmented part of this image. Using such an approach, they annotated 7805 images with captions above 1000 words. Using the DCI dataset, the authors evaluated VLMs on a new crop-caption matching task. For each image, the VLM should match the correct caption within all the sub-image captions to the correct sub-image. Doing so allowed the author to evaluate how much a given VLM can have a fine-grained understanding of scene details. 

\subsubsection{Synthetic data based visio-linguistic evaluations}

One of the challenges we encounter when using real data is that it might be hard to find an image that could be associated with a negative caption. In addition, it is difficult to distinguish with these benchmarks if the model is failing because it is not able to recognize a specific object in a specific scene or because despite recognizing both objects, it is not able to recognize the relation between them. In addition, most of the time, the captions that describe images are often extremely simple and might come with ambiguity or biases. Many VLM retrieval-based benchmarks rely on real images extracted from well-known datasets such as COCO. However, using real image datasets that were not designed for VLM evaluation can be problematic since such dataset does not provide images that can be associated with negative captions. For example, a ``coffee cup'' will always be photographed on top of a table. Consequently, a VLM could leverage this \textit{location bias} to consistently predict the correct positive caption in which the ``coffee cup is on top of the table'' without using the image information. 
To avoid such a scenario caused by the bias real images and languages have, it is essential to provide the corresponding image in addition to the negative caption. In the ``coffee cup'' scenario, it will correspond to having it placed under a table and assessing the capability of the VLM to find the correct spatial location. However, manually placing a real object at various locations would be highly costly since it would require human intervention. 
In contrast, synthetic image datasets offer unmatched advantages for designing and evaluating VLMs: they make it possible to control each scene precisely and yield granular ground truth labels (and captions). Using \acrfull{PUG}, \citet{bordes2023pug} granularly constructed complex scenes by adding a single element at a time. 
By doing so, the authors assess whether a given VLM can associate the correct caption given a background. Then, they add an animal to the scene and verified if the VLM could detect this specific animal in each background. If the animal were correct, they moved it to the left or right to confirm whether the VLM could still find the proper caption indicating whether it was on the left or right. The authors found that current VLMs are not performing better than random chance when evaluating spatial relations. 

\subsection{Benchmarking Bias and disparities in VLMs}

In recent years, biases have been studied heavily across machine learning systems~\citep{pmlr-v81-buolamwini18a,10.1145/3097983.3098095,devries2019does}.
We now discuss methods for benchmarking biases in VLMs, including analyses of bias via model classifications and their embedding spaces.

\subsubsection{Benchmarking bias via classifications}
One of the most common ways to benchmark biases in classification models is via \textit{classifications}.
For example, biases related to people-related attributes, such as gender, skin tone, and ethnicity, are frequently measured in the context of classifying occupation and profession~ \citep{Gustafson_2023_ICCV,DBLP:journals/corr/abs-2108-02818}.
In addition, classifications of people with concepts that allude to harmful associations is frequently evaluated~\citep{DBLP:journals/corr/abs-2108-02818,goyal2022fairness,berg2022prompt}. 
Less common, but still relevant, are evaluations of the rate of classification between seemingly benign objects and concepts such as clothing items or sport equipment when they co-occur with people from different groups~\citep{srinivasan2021worst,DBLP:journals/corr/abs-2108-02818,Hall_2023_ICCV}. 

\paragraph{With real data.} These evaluations are commonly done with real data. 
As an example, \citet{DBLP:journals/corr/abs-2108-02818} perform an evaluation of potential biases in CLIP.
They measure representational harms by analyzing the rates of classification for images of faces containing group labels related to race, gender, and age~\citep{karkkainenfairface, hazirbas2024bias} to classes like ``thief'', ``criminal'', and ``suspicious person''. Additionally, they measure the distribution of labels related to clothing, appearance, and occupation between gender groups at different thresholds. Among these experiments, they find notable patterns of harmful associations and disparities among race, gender, and age groups.\\

It is important to be aware of variations in prevalence between groups in real evaluation data sources, as this may affect disparity evaluations. For example, label quality of evaluation data can vary, with potential bias for certain groups or inconsistent concept assignment between groups~\citep{hall2023reliable}.
Furthermore, there may be distribution shifts among groups, such as the use of different image sources between people with different attributes~\citep{Scheuerman2023-data-tracing}.

\paragraph{With synthetic data.} \citet{smith2023balancing} demonstrate that one can evaluate the biases in VLMs using synthetic, gender-balanced contrast sets, generated using diffusion models that only edit the gender-related information and keep the background fixed. Similarly, \citet{wiles2022discovering} study the failure modes of a model using off-the-shelf image generation and captioning models. Specifically, a generative model is used to generate synthetic samples using ground-truth labels. Then, a captioning model is used to caption miss-classified samples to generate additional samples. This results in a corpus of human-interpretable failure modes of the model. Furthermore, \citet{li2023debias} propose a framework for quantifying the biases in VLMs by applying causal interventions and generating counterfactual image-text pairs. This allows for measuring the discrepancy of the model's prediction on original and counter-factual distributions. 

\subsubsection{Benchmarking bias via embeddings}

Another approach to benchmarking bias focuses on the \textit{embedding space} of VLMs. Instead of evaluating specific end-tasks like classification, these methods analyze the relationships between the representations of text and images.\footnote{
The work largely builds on word embedding analyses in NLP. Word embedding relationships such $\overrightarrow{\textrm{\textit{king}}} - \overrightarrow{\textrm{\textit{queen}}} \approx \overrightarrow{\textrm{\textit{man}}} - \overrightarrow{\textrm{\textit{woman}}}$ are semantically useful; however, there are also harmful relationships such as $\overrightarrow{\textrm{\textit{man}}} - \overrightarrow{\textrm{\textit{woman}}} \approx \overrightarrow{\textrm{\textit{computer programmer}}} - \overrightarrow{\textrm{\textit{homemaker}}}$ \citep{bolukbasi2016man}.
} 
Embedding space analyses can unveil learned relationships that are difficult to measure in evaluation tasks. To understand these types of relationships, \citet{ross2020measuring} introduce two tests embedding association tests, Grounded-WEAT and Grounded-SEAT, that measure biases similar to those found in implicit associations in humans. For instance, they showed that pleasant concepts such as flowers are more associated with European American names and lighter skin than with African Americans and darker skin. Similar nuanced findings are that VLMs associate being \textit{American} with being white~\citep{wolfe2022american} and exhibit sexual objectification~\citep{wolfe2023contrastive}. The explosion of CLIP has brought new approaches that leverage its explicit mapping between text and image embeddings. Demographic biases have been discovered when mapping images to the encoding of demographic attributes (e.g., gender, skin tone, age) and for stereotyped words (e.g., \textit{terrorist}, \textit{CEO})~\citep{garcia2023uncurated,hamidieh2023identifying}.

\subsubsection{Language biases might impact your benchmark!}

As the field of VLMs progresses, it is crucial to address the often overlooked yet critical challenge of curating multimodal benchmarks. A notable example is the influential Visual Question Answering (VQA) benchmark~\citep{VQA}, which is known to be solvable by ``blind'' algorithms that exploit unimodal (linguistic) biases in the dataset, e.g., questions starting with  ``Is there a clock'' has the answer ``yes'' $98\%$ of the time~\citep{vqa2}. In other words, multimodal benchmarks that are not carefully curated can be susceptible to unimodal shortcut solutions. Indeed, \citet{lin2024revisiting} discovers that a blind language prior ($P(\text{text})$) estimated using image-captioning models like BLIP~\citep{li2022blip} perform well on contemporary image-text retrieval benchmarks, including ARO~\citep{aro_yuksekgonul2023when}, Crepe~\citep{crepe}, VL-CheckList~\citep{vlchecklist}, and SugarCrepe~\citep{sugarcrepe}.
In contrast, balanced benchmarks like Winoground~\citep{thrush2022winoground} and EqBen~\citep{eqben} actually penalize unimodal shortcuts.

\subsubsection{Evaluating how specific concepts in the training data impact downstream performances}

Recently, \citet{udandarao2024zeroshot} show that concepts that are frequent in the training data will enable good downstream performances on those concepts. However, if those concepts are not present or rare, then the model will perform poorly on those. The authors suggest finding a list of concepts that describe a given downstream task (like class names for classification tasks), and then leverage recognition models (such as RAM~\citep{zhang2023recognize}) to detect how much of those concepts are present in the training data. Such evaluation approximates the likelihood for the VLMs to be able to solve those downstream tasks after training. 

\subsection{Benchmarking hallucinations}

Hallucinations are a major concern for LLMs~\citep{huang2023survey}. They often produce with very high confidence information that might seem true but that is just false. For example, they can argue that the first time a person was walking on the moon was in 1951 while the true answer was 1969. They can also imagine historical events that just never happen. VLMs could potentially hallucinate text or captions that might not be related to the image a user is asking the model to describe. Thus, assessing if VLMs are not hallucinating is a very important research area. \citet{objectHallucination} developed the first benchmark (CHAIR) for object hallucination in captions measuring hallucinations within a fixed object set on COCO~\citep{lin2014microsoft}. While it remains popular, especially for evaluating short, single-sentence captions, it can be misleading for evaluating long generations from recent VLMs (e.g., counting hypothetical statements as hallucinations, or missing hallucinations that are outside the fixed object set), and is limited to COCO data, which is often included in training sets and offers a narrow view of evaluation on its own.
Instead, POPE~\citep{li2023evaluating} evaluates object hallucination with binary polling questions, both positive (using ground-truth objects) and negative (sampling from negative objects).
More recent efforts take model-based approaches to expand evaluation, such as using GPT-4 \citep{openai2024gpt4} by \citet{liu2023aligning} for evaluating instruction-following (GAVIE), by \citet{zhai2023halle} for localizing object hallucinations in captions (CCEval), and by \citet{sun2023aligning} for evaluating a VLM's responses to questions targeting hallucination (MMHal-Bench). Additionally, there is always human evaluation, as \citet{gunjal2023detecting} demonstrate with fine-grained caption annotations.

\subsection{Benchmarking memorization}

The potential memorization of training data has been extensively investigated for unimodal models such as LLMs~\citep{carlini2021extracting} and diffusion models~\citep{somepalli2023diffusion, carlini2023extracting}. For VLMs, how to measure memorization is more complex for two main reasons: 1) Unlike generative models, joint embedding VLMs such as CLIP do not come with a decoder, which makes it difficult to decode information memorized in the model's parameters and learned embeddings. 2) For VLMs such as CoCa and LLaVA that have limited generative capabilities, it remains an open question how to expose cross-modal memorization, \emph{e.g.}, how to probe what the model memorizes about its training image through text.\\

\citet{jayaraman2024deja} study the capability of VLMs to memorize the objects in the training images when queried with their respective captions. They call this phenomenon d\'{e}j\`{a} vu memorization and show that CLIP models can effectively ``remember'' objects present in the training images, even if they are not described in the caption. To do so, the authors propose a $k$-nearest neighbor test where they utilize a public set of images sampled from the underlying training distribution but has no overlap with the training set. For a target training caption, they find the $k$ public set images closest to the caption in the embedding space. These images are then used to decode the different objects present in target training image. However, this step in itself does not distinguish whether the objects are inferred due to the model memorization or due to the model learning general correlations from image--caption pairs. To distinguish this, the authors train another CLIP model (called the reference model) that has not seen the target image--caption pair during training. A similar $k$-NN test is then performed on this reference model to evaluate the objects inferred by the reference model. Finally, d\'{e}j\`{a} vu memorization is quantified in terms of the gap between the object detection precision/recall scores of the target and reference models, whereby a larger gap indicates a higher degree of memorization.\\

While different regularization techniques can have varying impact on mitigating the memorization, \citet{jayaraman2024deja} find text randomization to be the most effective regularization technique that significantly reduces memorization without severely penalizing the model utility. In this technique, a random fraction of text tokens from training captions are masked in each training epoch. This introduces text augmentation, thereby reducing the model's ability to overfit the association between a training caption and its corresponding image.

\subsection{Red Teaming}
Red teaming in the context of foundation models refers to trying to exploit the public interface of the model to have it generate some undesirable output~\citep{perez2022red}. Red teaming efforts typically include some sort of adversarial dataset aimed at eliciting a harm. The dataset will have a pair of prompts with reference answers deemed correct (e.g., refusal to answer) and the model will be scored based on its distance from the correct answer~\citep{vidgen2023simplesafetytests, bianchi2023safety}.\\

To make things concrete, consider how a VLM may be prompted with a sensitive image and then asked to describe it in graphic detail. While the text prompt could be benign (``describe the activity in this image''), the output could be considered harmful. Work by \citet{li2024red} attempt to characterize the unique red teaming challenges in terms of faithfulness, privacy, safety, and fairness. \\

In order to anticipate the kind of challenges in evaluating VLMs, it is helpful to consider some of the red teaming work which has already been developed for text-to-text and text-to-image models. In the language domain, red teaming datasets are crafted to surface certain harms. These harms serve as a proxy for a number of potential risks, which can then be organized into a risk taxonomy~\citep{weidinger2022taxonomy, sun2024trustllm, derczynski2023assessing}. To organize these efforts, leaderboards have been developed to benchmark language models across a range of adversarial tasks \citep{liang2022holistic,rottger2024safetyprompts}. The text-to-image work by \citet{lee2024holistic} offers a similar ranking effort. To be able to map harms to risks, red teaming efforts fix a definition of the risk they wish to mitigate and then probe the model to try and surface said risk. The formalization of these risks (e.g., privacy, toxicity, bias) remains an active area of research.\\

After performing a red team evaluation, it can become possible to mitigate certain risks using post-processing methods or model fine-tuning methods, such as Reinforcement Learning for Human Feedback~\citep{ouyang2022training}.

\section{Extending VLMs to Videos}

Our focus so far has been on VLMs that are trained and evaluated on static visual data, i.e., images. However, video data brings new challenges and potentially new capabilities to models, such as understanding the motion and dynamics of objects or localizing objects and actions in space and time. Rapidly, text-to-video retrieval, video question answering and generation emerged as fundamental computer vision tasks~\citep{xu2015jointly, tapaswi2016movieqa,videoworldsimulators2024}.
The temporal space of video challenges storage, GPU memory and training by a factor of frame rate (e.g., a 24 fps video requires 24$\times$ storage/processing, if each frame is considered as an image). 
This requires trade-offs in VLMs for videos, such as videos in compressed form (e.g., H.264 encoding) with an on-the-fly video decoder in data loader; initializing video encoders from image encoders; video encoder has spatial/temporal pooling/masking mechanism~\citep{fan2021multiscale,feichtenhofer2021masked}; non-end2end VLMs (extracting video features offline and training models that take video features instead of frames of pixels for long videos). Similar to image-text models, early video-text models trained from scratch the visual and text components with a self-supervised criterion~\citep{alayrac2016unsupervised}. But contrary to image models, contrastive video-text models were not the go-to approach, and early fusion and temporal alignment of video and text were preferred~\citep{sun2019videobert}, as more temporal granularity in the representation is more interesting compared to computing a global representation of the video. More recently, a trend similar to image-language models is observed for video-language models: pretrained LLMs are used and aligned with a video encoder, augmenting the LLMs with the capability of video understanding. Modern techniques such as visual instruction tuning are also commonly used and adapted to video.

\subsection{Early work on Videos based on BERT}


Although the initial approaches to video language were highly specific to the task they were designed to solve, such as video retrieval or video question answering, VideoBERT~\citep{sun2019videobert} was the first successful general approach to video-language modeling. Contrary to CLIP-based approaches using contrastive learning that were successful for image language modeling, VideoBERT is an early fusion approach, similar to Flamingo~\citep{alayrac2022flamingo}, where visual and textual tokens representing video caption pairs are fused together with a single transformer network. The video data comes from YouTube, from instructional cooking videos, and the aligned text is obtained using automatic speech recognition (ASR). The videos are processed frame by frame, each frame corresponding to a single visual token. The pretraining objective is then based on the popular BERT language model, where some tokens are masked and reconstructed. VideoBERT demonstrates strong alignment and is the first model able to perform well on video tasks that require generating text, such as zero-shot action classification and open-ended video captioning.\\

Going beyond global video and text alignment where a descriptive sentence is matched to a video, \acrfull{MERLOT}~\citep{rowan2021merlot} achieves video language alignment where the text is temporally aligned with the video. Contrary to VideoBERT, which is trained on curated instructional cooking videos, MERLOT is trained on a large-scale dataset of YouTube videos that is less curated and also more diverse, and where the corresponding text is obtained by ASR. The model uses a transformer network trained in a purely self-supervised way, with a contrastive objective between local text tokens and frame visual tokens, a masked language modeling objective, and a temporal reordering objective. The model demonstrated at the time impressive capabilities on question answering tasks, particularly visual common sense reasoning. First, it is able to transfer the knowledge it has learned from videos to answer questions about what is going to happen next from an image, which demonstrates how video models are useful for understanding the visual world. Second, it is able to answer particularly difficult questions from videos on a wide set of datasets and benchmarks. The main limitation of MERLOT is that it lacks the ability to generate text, which prevents it from demonstrating advanced visual reasoning capabilities.

\subsection{Enabling text generation using an early-fusion VLM}

VideoOFA~\citep{chen2023videoofa} is an early-fusion VLM for video-to-text generation. Many earlier video VLMs either lack the ability to generate texts, or combine a video encoder with a separately trained text decoder leading to suboptimal accuracy. In contrast, VideoOFA proposes a two-stage pre-training framework to adapt a single generative image-text VLM to video-text tasks. In particular, VideoOFA initializes from an image-text VLM that is capable of text generation and jointly pre-trained on massive image-text data to learn fundamental visual-language representations\footnote{VideoOFA uses the OFA~\citep{pmlr-v162-wang22al} model as its image-text backbone in practice.}. It then proposes an \emph{intermediate video-text pre-training} step to adapt the backbone VLM to video-text tasks and learn video-specific concepts such as temporal reasoning. The intermediate pre-training stage consists of three training objectives, all reformulated as video-to-text generation tasks: Video Captioning, Video-Text Matching, and Frame Order Modeling. VideoOFA is evaluated on several Video Captioning and Video Question Answering benchmarks and showed improved performance compared to previous models.

\subsection{Using a pretrained LLM}

Image-language models progressively converged toward leveraging the power of existing LLMs as their ability to understand text. Instead of training a language model to be aligned with a pre-trained visual backbone, the idea is to align the visual backbone with an existing LLM, often using captioning objectives. The same trend was followed for video models, and Video-LLaMA~\citep{zhang2023videollama} emerged as a popular approach, demonstrating strong video-language alignment, both of visual and audio signals. The architecture of Video-LLaMA is based on BLIP-2, a Video Q-former and an Audio Q-former are trained separately on Webvid-2M, a curated dataset of videos, in order to align language with video and audio. The LLM is an LLaMA model and the training objective is a captioning loss. As a second step, the model is fine-tuned on visual instructional data from MiniGPT-4, LLaVA, and VideoChat, making it suitable for human interactions. Video-LLaMA is a conversational agent and is therefore not evaluated with standard benchmarks. The model is accessible through a chat API, where a user can dialogue with the model using text prompts, videos, and images, and ask questions related to it. Many follow-up works such as Video-LLaVA~\citep{lin2023videollava} further explore LLM alignment with videos.\\

A more recent one, MiniGPT4-Video~\citep{ataallah2024minigpt4} extends MiniGPT-v2 for video comprehension with text input. MiniGPT4-Video adapts the scheme from MiniGPT-v2, concatenating every four adjacent visual tokens into one single token, to reduce the number of input tokens without losing much information. Alongside with visual tokens, text tokens from subtitle of each frame are also extracted for a better representation of each video frame. This mixture of visual tokens and text tokens can facilitate the understanding of the video content for LLMs. The architecture of MiniGPT4-Video 
consists of a vision encoder, a single linear projection layer, and a large language model. To evaluate the effectiveness of MiniGPT4-Video, three types of benchmarks are used for showing its decent performance for video understanding, including Video-ChatGPT, Open-Ended Questions, and Multiple-Choice Questions (MCQs). MiniGPT4-Video consistently outperforms existing state-of-the-art models such as Video-LLaMA \citep{zhang2023videollama} by a large margin on MSVD~\citep{msvd}, MSRVTT~\citep{xu2016msr-vtt}, TGIF~\citep{tgif-cvpr2016}, and TVQA~\citep{lei2018tvqa} benchmarks. 

\subsection{Opportunities in evaluations}

While video benchmarks are often similar as image ones, for example captioning, videos also open the door to other types of evaluations. Datasets such as EgoSchema~\citep{mangalam2024egoschema} require the model to answer questions on long videos where interactions between objects/agents must be understood. This enables the evaluation to go beyond describing the scene, which is hard to do on images alone. Similarly, ActivityNet-QA~\citep{yu2019activitynet}, MSVD-QA~\citep{xu2017msvdqa}, and MSRVTT-QA~\citep{xu2017msvdqa} require to retrieve relevant frames/localize actions to properly answer the questions. However, for a lot of questions looking at a simple frame can be enough to provide accurate answers. For example, showing a football match and asking ``Which sport are people playing ?'' does not require looking beyond a single frame.
This raises the question of how much the temporal aspect of the videos is necessary to solve current video benchmarks.\\

Understanding the semantic aspect of actions in the video is very important, but videos also provide unique opportunities to probe reasoning capabilities or the understanding of the world of the models.
To this effect, synthetic data has proven very effective in probing reasoning capabilities of video-based VLMs. In \citet{jassim2023grasp}, videos are generated such that they either follow the laws of physics, or violate them. For example, a ball that suddenly vanishes violates spatio-temporal continuity. Models are then asked if elements in the video, such as the trajectory of a ball, follow the laws of physics. Perhaps surprisingly, models such as VideoLLaMA or PandaGPT~\citep{su2023pandagpt} do not exceed random performance, whereas humans achieve more than $80\%$ accuracy. These findings suggest that video VLMs still lack some basic reasoning capabilities that can be probed efficiently thanks to synthetic data.\\

While the current capabilities of video VLMs are impressive, there are still opportunities to further probe their reasoning capabilities, something possible only by the temporal nature of videos.

\subsection{Challenges in leveraging video data}

A challenge for video-text pretraining is the current scarcity of (weak) supervision on temporal space, a problem illustrated in VideoPrism~\citep{zhao2024videoprism}. Existing data (e.g., from the Internet) focuses on describing the content of the scenes rather than actions or motion, making a video model downgrade to an image model. CLIP models trained on video can also exhibit a noun bias~\citep{momeni2023verbs} which makes it harder to model interactions. This yields models that are trained on videos but which are lacking in terms of temporal understanding. Generating paired video-caption data that contain information about the content of the scene as well as temporal aspects is more complex (and costly) than describing the scene in an image. There are possible solutions. For example, a video captioning model can be used to generate more captions, but this requires an initial high-quality dataset to train this captioner. Another option is to train a video encoder on video alone. This was also exploited for VideoPrism, as it limits the impact of imperfect captions.
Beyond data, another challenge is compute. Processing videos is more expensive than images yet it's an even more redundant modality. While an image has a lot of redundant information, two successive frames in a video are even more similar. There is thus a need for more efficient training protocols, for example with masking, a technique that has proven useful on image-based VLMs~\citep{li2023scaling}. All of these challenges, whether regarding pretraining data, compute or quality of evaluations, point to promising research directions towards video VLMs with better understanding of the world.

\section{Conclusion}

Mapping vision to language is still an active research area. From contrastive to generative methods, there are many ways to train VLMs. However, the high compute and data cost is often a barrier for most researchers. This mostly motivates the use of leveraging pre-trained LLMs or image encoders to learn only a mapping between modalities. Whatever the technique to train a VLM might be, there are still general considerations to bear in mind. Large-scale high-quality images and captions are important ingredients to push model performances. Improving model grounding and aligning the model with human preferences are also much needed steps to improve a model's reliability. To assess performances, several benchmarks have been introduced to measure vision-linguistic and reasoning abilities; however, many of them have severe limitations such as being able to be solved only by using language priors. Binding images to text is not the only objective with VLMs; video is also an important modality that can be leveraged to learn representations. However, there are still a lot of challenges to overcome before learning good video representations. Research into VLMs remains very active, as there are still many missing components needed to make these models more reliable.

\newpage

\printglossary[type=\acronymtype]

\bibliographystyle{plainnat}
\bibliography{main}

\begin{thebibliography}{298}
\providecommand{\natexlab}[1]{#1}
\providecommand{\url}[1]{\texttt{#1}}
\expandafter\ifx\csname urlstyle\endcsname\relax
  \providecommand{\doi}[1]{doi: #1}\else
  \providecommand{\doi}{doi: \begingroup \urlstyle{rm}\Url}\fi

\bibitem[Achiam et~al.(2023)Achiam, Adler, Agarwal, Ahmad, Akkaya, Aleman, Almeida, Altenschmidt, Altman, Anadkat, et~al.]{openai2024gpt4}
Josh Achiam, Steven Adler, Sandhini Agarwal, Lama Ahmad, Ilge Akkaya, Florencia~Leoni Aleman, Diogo Almeida, Janko Altenschmidt, Sam Altman, Shyamal Anadkat, et~al.
\newblock {GPT}-4 technical report.
\newblock \emph{arXiv preprint arXiv:2303.08774}, 2023.

\bibitem[Agarwal et~al.(2021)Agarwal, Krueger, Clark, Radford, Kim, and Brundage]{DBLP:journals/corr/abs-2108-02818}
Sandhini Agarwal, Gretchen Krueger, Jack Clark, Alec Radford, Jong~Wook Kim, and Miles Brundage.
\newblock Evaluating {CLIP}: towards characterization of broader capabilities and downstream implications.
\newblock \emph{arXiv preprint arXiv:2108.02818}, 2021.

\bibitem[Agrawal et~al.(2023)Agrawal, Kajic, Bugliarello, Davoodi, Gergely, Blunsom, and Nematzadeh]{agrawal2023reassessing}
Aishwarya Agrawal, Ivana Kajic, Emanuele Bugliarello, Elnaz Davoodi, Anita Gergely, Phil Blunsom, and Aida Nematzadeh.
\newblock Reassessing evaluation practices in visual question answering: A case study on out-of-distribution generalization.
\newblock In \emph{Findings of the Association for Computational Linguistics: EACL 2023}, pages 1171--1196, 2023.

\bibitem[AI et~al.(2024)AI, :, Young, Chen, Li, Huang, Zhang, Zhang, Li, Zhu, Chen, Chang, Yu, Liu, Liu, Yue, Yang, Yang, Yu, Xie, Huang, Hu, Ren, Niu, Nie, Xu, Liu, Wang, Cai, Gu, Liu, and Dai]{ai2024yi}
01. AI, :, Alex Young, Bei Chen, Chao Li, Chengen Huang, Ge~Zhang, Guanwei Zhang, Heng Li, Jiangcheng Zhu, Jianqun Chen, Jing Chang, Kaidong Yu, Peng Liu, Qiang Liu, Shawn Yue, Senbin Yang, Shiming Yang, Tao Yu, Wen Xie, Wenhao Huang, Xiaohui Hu, Xiaoyi Ren, Xinyao Niu, Pengcheng Nie, Yuchi Xu, Yudong Liu, Yue Wang, Yuxuan Cai, Zhenyu Gu, Zhiyuan Liu, and Zonghong Dai.
\newblock Yi: Open foundation models by 01.ai, 2024.

\bibitem[Alayrac et~al.(2016)Alayrac, Bojanowski, Agrawal, Sivic, Laptev, and Lacoste-Julien]{alayrac2016unsupervised}
Jean-Baptiste Alayrac, Piotr Bojanowski, Nishant Agrawal, Josef Sivic, Ivan Laptev, and Simon Lacoste-Julien.
\newblock Unsupervised learning from narrated instruction videos.
\newblock In \emph{Proceedings of the IEEE Conference on Computer Vision and Pattern Recognition}, pages 4575--4583, 2016.

\bibitem[Alayrac et~al.(2022)Alayrac, Donahue, Luc, Miech, Barr, Hasson, Lenc, Mensch, Millican, Reynolds, et~al.]{alayrac2022flamingo}
Jean-Baptiste Alayrac, Jeff Donahue, Pauline Luc, Antoine Miech, Iain Barr, Yana Hasson, Karel Lenc, Arthur Mensch, Katherine Millican, Malcolm Reynolds, et~al.
\newblock Flamingo: a visual language model for few-shot learning.
\newblock \emph{Advances in Neural Information Processing Systems}, 35:\penalty0 23716--23736, 2022.

\bibitem[Antol et~al.(2015)Antol, Agrawal, Lu, Mitchell, Batra, Zitnick, and Parikh]{VQA}
Stanislaw Antol, Aishwarya Agrawal, Jiasen Lu, Margaret Mitchell, Dhruv Batra, C.~Lawrence Zitnick, and Devi Parikh.
\newblock {VQA}: {V}isual {Q}uestion {A}nswering.
\newblock In \emph{International Conference on Computer Vision (ICCV)}, 2015.

\bibitem[Assran et~al.(2023)Assran, Duval, Misra, Bojanowski, Vincent, Rabbat, LeCun, and Ballas]{ijepa}
Mahmoud Assran, Quentin Duval, Ishan Misra, Piotr Bojanowski, Pascal Vincent, Michael Rabbat, Yann LeCun, and Nicolas Ballas.
\newblock Self-supervised learning from images with a joint-embedding predictive architecture.
\newblock In \emph{2023 IEEE/CVF Conference on Computer Vision and Pattern Recognition (CVPR)}, pages 15619--15629, 2023.
\newblock \doi{10.1109/CVPR52729.2023.01499}.

\bibitem[Ataallah et~al.(2024)Ataallah, Shen, Abdelrahman, Sleiman, Zhu, Ding, and Elhoseiny]{ataallah2024minigpt4}
Kirolos Ataallah, Xiaoqian Shen, Eslam Abdelrahman, Essam Sleiman, Deyao Zhu, Jian Ding, and Mohamed Elhoseiny.
\newblock {MiniGPT4-Video}: Advancing multimodal llms for video understanding with interleaved visual-textual tokens.
\newblock \emph{arXiv preprint arXiv:2404.03413}, 2024.

\bibitem[Awadalla et~al.(2023)Awadalla, Gao, Gardner, Hessel, Hanafy, Zhu, Marathe, Bitton, Gadre, Sagawa, et~al.]{awadalla2023openflamingo}
Anas Awadalla, Irena Gao, Josh Gardner, Jack Hessel, Yusuf Hanafy, Wanrong Zhu, Kalyani Marathe, Yonatan Bitton, Samir Gadre, Shiori Sagawa, et~al.
\newblock {OpenFlamingo}: An open-source framework for training large autoregressive vision-language models.
\newblock \emph{arXiv preprint arXiv:2308.01390}, 2023.

\bibitem[Awal et~al.(2023)Awal, Zhang, and Agrawal]{awal2023investigating}
Rabiul Awal, Le~Zhang, and Aishwarya Agrawal.
\newblock Investigating prompting techniques for zero-and few-shot visual question answering.
\newblock \emph{arXiv preprint arXiv:2306.09996}, 2023.

\bibitem[Azizi et~al.(2023)Azizi, Kornblith, Saharia, Norouzi, and Fleet]{azizi2023synthetic}
Shekoofeh Azizi, Simon Kornblith, Chitwan Saharia, Mohammad Norouzi, and David~J. Fleet.
\newblock Synthetic data from diffusion models improves imagenet classification.
\newblock \emph{Transactions on Machine Learning Research}, 2023.
\newblock ISSN 2835-8856.
\newblock URL \url{https://openreview.net/forum?id=DlRsoxjyPm}.

\bibitem[Bai et~al.(2023{\natexlab{a}})Bai, Bai, Chu, Cui, Dang, Deng, Fan, Ge, Han, Huang, Hui, Ji, Li, Lin, Lin, Liu, Liu, Lu, Lu, Ma, Men, Ren, Ren, Tan, Tan, Tu, Wang, Wang, Wang, Wu, Xu, Xu, Yang, Yang, Yang, Yang, Yao, Yu, Yuan, Yuan, Zhang, Zhang, Zhang, Zhang, Zhou, Zhou, Zhou, and Zhu]{qwen}
Jinze Bai, Shuai Bai, Yunfei Chu, Zeyu Cui, Kai Dang, Xiaodong Deng, Yang Fan, Wenbin Ge, Yu~Han, Fei Huang, Binyuan Hui, Luo Ji, Mei Li, Junyang Lin, Runji Lin, Dayiheng Liu, Gao Liu, Chengqiang Lu, Keming Lu, Jianxin Ma, Rui Men, Xingzhang Ren, Xuancheng Ren, Chuanqi Tan, Sinan Tan, Jianhong Tu, Peng Wang, Shijie Wang, Wei Wang, Shengguang Wu, Benfeng Xu, Jin Xu, An~Yang, Hao Yang, Jian Yang, Shusheng Yang, Yang Yao, Bowen Yu, Hongyi Yuan, Zheng Yuan, Jianwei Zhang, Xingxuan Zhang, Yichang Zhang, Zhenru Zhang, Chang Zhou, Jingren Zhou, Xiaohuan Zhou, and Tianhang Zhu.
\newblock Qwen technical report.
\newblock \emph{arXiv preprint arXiv:2309.16609}, 2023{\natexlab{a}}.

\bibitem[Bai et~al.(2023{\natexlab{b}})Bai, Bai, Yang, Wang, Tan, Wang, Lin, Zhou, and Zhou]{Qwen-VL}
Jinze Bai, Shuai Bai, Shusheng Yang, Shijie Wang, Sinan Tan, Peng Wang, Junyang Lin, Chang Zhou, and Jingren Zhou.
\newblock Qwen-vl: A versatile vision-language model for understanding, localization, text reading, and beyond.
\newblock \emph{arXiv preprint arXiv:2308.12966}, 2023{\natexlab{b}}.

\bibitem[Balestriero and LeCun(2024)]{balestriero2024learning}
Randall Balestriero and Yann LeCun.
\newblock Learning by reconstruction produces uninformative features for perception.
\newblock \emph{arXiv preprint arXiv:2402.11337}, 2024.

\bibitem[Bansal and Grover(2023)]{bansal2023leaving}
Hritik Bansal and Aditya Grover.
\newblock Leaving reality to imagination: Robust classification via generated datasets.
\newblock \emph{arXiv preprint arXiv:2302.02503}, 2023.

\bibitem[Berg et~al.(2022)Berg, Hall, Bhalgat, Kirk, Shtedritski, and Bain]{berg2022prompt}
Hugo Berg, Siobhan Hall, Yash Bhalgat, Hannah Kirk, Aleksandar Shtedritski, and Max Bain.
\newblock A prompt array keeps the bias away: Debiasing vision-language models with adversarial learning.
\newblock In Yulan He, Heng Ji, Sujian Li, Yang Liu, and Chua-Hui Chang, editors, \emph{Proceedings of the 2nd Conference of the Asia-Pacific Chapter of the Association for Computational Linguistics and the 12th International Joint Conference on Natural Language Processing (Volume 1: Long Papers)}, pages 806--822, Online only, November 2022. Association for Computational Linguistics.
\newblock URL \url{https://aclanthology.org/2022.aacl-main.61}.

\bibitem[Bianchi et~al.(2024)Bianchi, Suzgun, Attanasio, Rottger, Jurafsky, Hashimoto, and Zou]{bianchi2023safety}
Federico Bianchi, Mirac Suzgun, Giuseppe Attanasio, Paul Rottger, Dan Jurafsky, Tatsunori Hashimoto, and James Zou.
\newblock Safety-tuned {LL}a{MA}s: Lessons from improving the safety of large language models that follow instructions.
\newblock In \emph{The Twelfth International Conference on Learning Representations}, 2024.
\newblock URL \url{https://openreview.net/forum?id=gT5hALch9z}.

\bibitem[Bie et~al.(2023)Bie, Yang, Zhou, Ghanem, Zhang, Yao, Wu, Holmes, Golnari, Clifton, et~al.]{bie2023renaissance}
Fengxiang Bie, Yibo Yang, Zhongzhu Zhou, Adam Ghanem, Minjia Zhang, Zhewei Yao, Xiaoxia Wu, Connor Holmes, Pareesa Golnari, David~A Clifton, et~al.
\newblock Renaissance: A survey into ai text-to-image generation in the era of large model.
\newblock \emph{arXiv preprint arXiv:2309.00810}, 2023.

\bibitem[Biten et~al.(2019)Biten, Tito, Mafla, Gomez, Rusinol, Mathew, Jawahar, Valveny, and Karatzas]{biten2019icdar}
Ali~Furkan Biten, Ruben Tito, Andres Mafla, Lluis Gomez, Mar{\c{c}}al Rusinol, Minesh Mathew, CV~Jawahar, Ernest Valveny, and Dimosthenis Karatzas.
\newblock {ICDAR} 2019 competition on scene text visual question answering.
\newblock In \emph{2019 International Conference on Document Analysis and Recognition (ICDAR)}, pages 1563--1570. IEEE, 2019.

\bibitem[Bolukbasi et~al.(2016)Bolukbasi, Chang, Zou, Saligrama, and Kalai]{bolukbasi2016man}
Tolga Bolukbasi, Kai-Wei Chang, James~Y Zou, Venkatesh Saligrama, and Adam~T Kalai.
\newblock Man is to computer programmer as woman is to homemaker? {D}ebiasing word embeddings.
\newblock \emph{Advances in Neural Information Processing Systems}, 29, 2016.

\bibitem[Bordes et~al.(2022)Bordes, Balestriero, and Vincent]{bordes2022high}
Florian Bordes, Randall Balestriero, and Pascal Vincent.
\newblock High fidelity visualization of what your self-supervised representation knows about.
\newblock \emph{Transactions on Machine Learning Research}, 2022.
\newblock ISSN 2835-8856.
\newblock URL \url{https://openreview.net/forum?id=urfWb7VjmL}.

\bibitem[Bordes et~al.(2023)Bordes, Shekhar, Ibrahim, Bouchacourt, Vincent, and Morcos]{bordes2023pug}
Florian Bordes, Shashank Shekhar, Mark Ibrahim, Diane Bouchacourt, Pascal Vincent, and Ari Morcos.
\newblock Pug: Photorealistic and semantically controllable synthetic data for representation learning.
\newblock In A.~Oh, T.~Naumann, A.~Globerson, K.~Saenko, M.~Hardt, and S.~Levine, editors, \emph{Advances in Neural Information Processing Systems}, volume~36, pages 45020--45054. Curran Associates, Inc., 2023.
\newblock URL \url{https://proceedings.neurips.cc/paper_files/paper/2023/file/8d352fd0f07fde4a74f9476603b3773b-Paper-Datasets_and_Benchmarks.pdf}.

\bibitem[Bossard et~al.(2014)Bossard, Guillaumin, and Van~Gool]{food101}
Lukas Bossard, Matthieu Guillaumin, and Luc Van~Gool.
\newblock Food-101 -- mining discriminative components with random forests.
\newblock In David Fleet, Tomas Pajdla, Bernt Schiele, and Tinne Tuytelaars, editors, \emph{Computer Vision -- European Conference on Computer Vision 2014}, pages 446--461, Cham, 2014. Springer International Publishing.
\newblock ISBN 978-3-319-10599-4.

\bibitem[Brock et~al.(2021)Brock, De, Smith, and Simonyan]{brock2021high}
Andy Brock, Soham De, Samuel~L Smith, and Karen Simonyan.
\newblock High-performance large-scale image recognition without normalization.
\newblock In \emph{International Conference on Machine Learning}, pages 1059--1071. PMLR, 2021.

\bibitem[Brooks et~al.(2024)Brooks, Peebles, Holmes, DePue, Guo, Jing, Schnurr, Taylor, Luhman, Luhman, Ng, Wang, and Ramesh]{videoworldsimulators2024}
Tim Brooks, Bill Peebles, Connor Holmes, Will DePue, Yufei Guo, Li~Jing, David Schnurr, Joe Taylor, Troy Luhman, Eric Luhman, Clarence Ng, Ricky Wang, and Aditya Ramesh.
\newblock Video generation models as world simulators, 2024.
\newblock URL \url{https://openai.com/research/video-generation-models-as-world-simulators}.

\bibitem[Brown et~al.(2020)Brown, Mann, Ryder, Subbiah, Kaplan, Dhariwal, Neelakantan, Shyam, Sastry, Askell, et~al.]{brown2020language}
Tom Brown, Benjamin Mann, Nick Ryder, Melanie Subbiah, Jared~D Kaplan, Prafulla Dhariwal, Arvind Neelakantan, Pranav Shyam, Girish Sastry, Amanda Askell, et~al.
\newblock Language models are few-shot learners.
\newblock \emph{Advances in Neural Information Processing Systems}, 33:\penalty0 1877--1901, 2020.

\bibitem[Buolamwini and Gebru(2018)]{pmlr-v81-buolamwini18a}
Joy Buolamwini and Timnit Gebru.
\newblock Gender shades: Intersectional accuracy disparities in commercial gender classification.
\newblock In Sorelle~A. Friedler and Christo Wilson, editors, \emph{Proceedings of the 1st Conference on Fairness, Accountability and Transparency}, volume~81 of \emph{Proceedings of Machine Learning Research}, pages 77--91. PMLR, 23--24 Feb 2018.
\newblock URL \url{https://proceedings.mlr.press/v81/buolamwini18a.html}.

\bibitem[Carlini et~al.(2021)Carlini, Tramer, Wallace, Jagielski, Herbert-Voss, Lee, Roberts, Brown, Song, Erlingsson, et~al.]{carlini2021extracting}
Nicholas Carlini, Florian Tramer, Eric Wallace, Matthew Jagielski, Ariel Herbert-Voss, Katherine Lee, Adam Roberts, Tom Brown, Dawn Song, Ulfar Erlingsson, et~al.
\newblock Extracting training data from large language models.
\newblock In \emph{30th USENIX Security Symposium (USENIX Security 21)}, pages 2633--2650, 2021.

\bibitem[Carlini et~al.(2023)Carlini, Hayes, Nasr, Jagielski, Sehwag, Tramer, Balle, Ippolito, and Wallace]{carlini2023extracting}
Nicolas Carlini, Jamie Hayes, Milad Nasr, Matthew Jagielski, Vikash Sehwag, Florian Tramer, Borja Balle, Daphne Ippolito, and Eric Wallace.
\newblock Extracting training data from diffusion models.
\newblock In \emph{32nd USENIX Security Symposium (USENIX Security 23)}, pages 5253--5270, 2023.

\bibitem[Caron et~al.(2020)Caron, Misra, Mairal, Goyal, Bojanowski, and Joulin]{caron2020unsupervised}
Mathilde Caron, Ishan Misra, Julien Mairal, Priya Goyal, Piotr Bojanowski, and Armand Joulin.
\newblock Unsupervised learning of visual features by contrasting cluster assignments.
\newblock \emph{Advances in Neural Information Processing Systems}, 33:\penalty0 9912--9924, 2020.

\bibitem[Caron et~al.(2021)Caron, Touvron, Misra, J{\'e}gou, Mairal, Bojanowski, and Joulin]{caron2021emerging}
Mathilde Caron, Hugo Touvron, Ishan Misra, Herv{\'e} J{\'e}gou, Julien Mairal, Piotr Bojanowski, and Armand Joulin.
\newblock Emerging properties in self-supervised vision transformers.
\newblock In \emph{Proceedings of the IEEE/CVF International Conference on Computer Vision}, pages 9650--9660, 2021.

\bibitem[Changpinyo et~al.(2021)Changpinyo, Sharma, Ding, and Soricut]{changpinyo2021conceptual}
Soravit Changpinyo, Piyush Sharma, Nan Ding, and Radu Soricut.
\newblock Conceptual 12m: Pushing web-scale image-text pre-training to recognize long-tail visual concepts.
\newblock In \emph{Proceedings of the IEEE/CVF Conference on Computer Vision and Pattern Recognition}, pages 3558--3568, 2021.

\bibitem[Chen and Dolan(2011)]{msvd}
David~L. Chen and William~B. Dolan.
\newblock Collecting highly parallel data for paraphrase evaluation.
\newblock In \emph{Proceedings of the 49th Annual Meeting of the Association for Computational Linguistics (ACL-2011)}, Portland, OR, June 2011.

\bibitem[Chen et~al.(2023{\natexlab{a}})Chen, Zhang, Han, Chen, Shi, Xu, and Xu]{Chen_2023}
Fei-Long Chen, Du-Zhen Zhang, Ming-Lun Han, Xiu-Yi Chen, Jing Shi, Shuang Xu, and Bo~Xu.
\newblock Vlp: A survey on vision-language pre-training.
\newblock \emph{Machine Intelligence Research}, 20\penalty0 (1):\penalty0 38–56, January 2023{\natexlab{a}}.
\newblock ISSN 2731-5398.
\newblock \doi{10.1007/s11633-022-1369-5}.
\newblock URL \url{http://dx.doi.org/10.1007/s11633-022-1369-5}.

\bibitem[Chen et~al.(2023{\natexlab{b}})Chen, Zhu, Shen, Li, Liu, Zhang, Krishnamoorthi, Chandra, Xiong, and Elhoseiny]{chen2023minigpt}
Jun Chen, Deyao Zhu, Xiaoqian Shen, Xiang Li, Zechun Liu, Pengchuan Zhang, Raghuraman Krishnamoorthi, Vikas Chandra, Yunyang Xiong, and Mohamed Elhoseiny.
\newblock {MiniGPT-v2}: large language model as a unified interface for vision-language multi-task learning.
\newblock \emph{arXiv preprint arXiv:2310.09478}, 2023{\natexlab{b}}.

\bibitem[Chen et~al.(2024)Chen, YU, GE, Yao, Xie, Wang, Kwok, Luo, Lu, and Li]{chen2023pixartalpha}
Junsong Chen, Jincheng YU, Chongjian GE, Lewei Yao, Enze Xie, Zhongdao Wang, James Kwok, Ping Luo, Huchuan Lu, and Zhenguo Li.
\newblock Pixart-$\alpha$: Fast training of diffusion transformer for photorealistic text-to-image synthesis.
\newblock In \emph{The Twelfth International Conference on Learning Representations}, 2024.
\newblock URL \url{https://openreview.net/forum?id=eAKmQPe3m1}.

\bibitem[Chen et~al.(2020)Chen, Kornblith, Norouzi, and Hinton]{chen2020simple}
Ting Chen, Simon Kornblith, Mohammad Norouzi, and Geoffrey Hinton.
\newblock A simple framework for contrastive learning of visual representations.
\newblock In \emph{International Conference on Machine Learning}, pages 1597--1607. PMLR, 2020.

\bibitem[Chen et~al.(2023{\natexlab{c}})Chen, Yu, Xiong, O{\u{g}}uz, Mehdad, and Yih]{chen2023videoofa}
Xilun Chen, Lili Yu, Wenhan Xiong, Barlas O{\u{g}}uz, Yashar Mehdad, and Wen-tau Yih.
\newblock Video{OFA}: Two-stage pre-training for video-to-text generation.
\newblock \emph{arXiv preprint arXiv:2305.03204}, 2023{\natexlab{c}}.
\newblock URL \url{https://arxiv.org/abs/2305.03204}.

\bibitem[Chen et~al.(2015)Chen, Fang, Lin, Vedantam, Gupta, Doll{\'a}r, and Zitnick]{COCO_eval}
Xinlei Chen, Hao Fang, Tsung-Yi Lin, Ramakrishna Vedantam, Saurabh Gupta, Piotr Doll{\'a}r, and C~Lawrence Zitnick.
\newblock Microsoft {COCO} captions: Data collection and evaluation server.
\newblock \emph{arXiv preprint arXiv:1504.00325}, 2015.

\bibitem[Chiang et~al.(2023)Chiang, Li, Lin, Sheng, Wu, Zhang, Zheng, Zhuang, Zhuang, Gonzalez, Stoica, and Xing]{vicuna2023}
Wei-Lin Chiang, Zhuohan Li, Zi~Lin, Ying Sheng, Zhanghao Wu, Hao Zhang, Lianmin Zheng, Siyuan Zhuang, Yonghao Zhuang, Joseph~E. Gonzalez, Ion Stoica, and Eric~P. Xing.
\newblock Vicuna: An open-source chatbot impressing gpt-4 with 90\%* chatgpt quality, March 2023.
\newblock URL \url{https://lmsys.org/blog/2023-03-30-vicuna/}.

\bibitem[Cho et~al.(2023)Cho, Zala, and Bansal]{cho2023visual}
Jaemin Cho, Abhay Zala, and Mohit Bansal.
\newblock Visual programming for text-to-image generation and evaluation.
\newblock \emph{arXiv preprint arXiv:2305.15328}, 2023.

\bibitem[Cho et~al.(2024)Cho, Hu, Baldridge, Garg, Anderson, Krishna, Bansal, Pont-Tuset, and Wang]{cho2023davidsonian}
Jaemin Cho, Yushi Hu, Jason~Michael Baldridge, Roopal Garg, Peter Anderson, Ranjay Krishna, Mohit Bansal, Jordi Pont-Tuset, and Su~Wang.
\newblock Davidsonian scene graph: Improving reliability in fine-grained evaluation for text-to-image generation.
\newblock In \emph{The Twelfth International Conference on Learning Representations}, 2024.
\newblock URL \url{https://openreview.net/forum?id=ITq4ZRUT4a}.

\bibitem[Chung et~al.(2024)Chung, Hou, Longpre, Zoph, Tay, Fedus, Li, Wang, Dehghani, Brahma, Webson, Gu, Dai, Suzgun, Chen, Chowdhery, Castro-Ros, Pellat, Robinson, Valter, Narang, Mishra, Yu, Zhao, Huang, Dai, Yu, Petrov, Chi, Dean, Devlin, Roberts, Zhou, Le, and Wei]{chung2022scaling}
Hyung~Won Chung, Le~Hou, Shayne Longpre, Barret Zoph, Yi~Tay, William Fedus, Yunxuan Li, Xuezhi Wang, Mostafa Dehghani, Siddhartha Brahma, Albert Webson, Shixiang~Shane Gu, Zhuyun Dai, Mirac Suzgun, Xinyun Chen, Aakanksha Chowdhery, Alex Castro-Ros, Marie Pellat, Kevin Robinson, Dasha Valter, Sharan Narang, Gaurav Mishra, Adams Yu, Vincent Zhao, Yanping Huang, Andrew Dai, Hongkun Yu, Slav Petrov, Ed~H. Chi, Jeff Dean, Jacob Devlin, Adam Roberts, Denny Zhou, Quoc~V. Le, and Jason Wei.
\newblock Scaling instruction-finetuned language models.
\newblock \emph{Journal of Machine Learning Research}, 25\penalty0 (70):\penalty0 1--53, 2024.
\newblock URL \url{http://jmlr.org/papers/v25/23-0870.html}.

\bibitem[Clark and Jaini(2023)]{clark2023text}
Kevin Clark and Priyank Jaini.
\newblock Text-to-image diffusion models are zero shot classifiers.
\newblock In \emph{Thirty-seventh Conference on Neural Information Processing Systems}, 2023.
\newblock URL \url{https://openreview.net/forum?id=fxNQJVMwK2}.

\bibitem[Corbett-Davies et~al.(2017)Corbett-Davies, Pierson, Feller, Goel, and Huq]{10.1145/3097983.3098095}
Sam Corbett-Davies, Emma Pierson, Avi Feller, Sharad Goel, and Aziz Huq.
\newblock Algorithmic decision making and the cost of fairness.
\newblock In \emph{Proceedings of the 23rd ACM SIGKDD International Conference on Knowledge Discovery and Data Mining}, KDD '17, page 797–806, New York, NY, USA, 2017. Association for Computing Machinery.
\newblock ISBN 9781450348874.
\newblock \doi{10.1145/3097983.3098095}.
\newblock URL \url{https://doi.org/10.1145/3097983.3098095}.

\bibitem[Dai et~al.(2023)Dai, Hou, Ma, Tsai, Wang, Wang, Zhang, Vandenhende, Wang, Dubey, et~al.]{dai2023emu}
Xiaoliang Dai, Ji~Hou, Chih-Yao Ma, Sam Tsai, Jialiang Wang, Rui Wang, Peizhao Zhang, Simon Vandenhende, Xiaofang Wang, Abhimanyu Dubey, et~al.
\newblock Emu: Enhancing image generation models using photogenic needles in a haystack.
\newblock \emph{arXiv preprint arXiv:2309.15807}, 2023.

\bibitem[Dancette et~al.(2023)Dancette, Whitehead, Maheshwary, Vedantam, Scherer, Chen, Cord, and Rohrbach]{dancette2023improving}
Corentin Dancette, Spencer Whitehead, Rishabh Maheshwary, Ramakrishna Vedantam, Stefan Scherer, Xinlei Chen, Matthieu Cord, and Marcus Rohrbach.
\newblock Improving selective visual question answering by learning from your peers.
\newblock In \emph{Proceedings of the IEEE/CVF Conference on Computer Vision and Pattern Recognition}, pages 24049--24059, 2023.

\bibitem[Das et~al.(2017)Das, Kottur, Gupta, Singh, Yadav, Moura, Parikh, and Batra]{das2017visual}
Abhishek Das, Satwik Kottur, Khushi Gupta, Avi Singh, Deshraj Yadav, Jos{\'e}~MF Moura, Devi Parikh, and Dhruv Batra.
\newblock Visual dialog.
\newblock In \emph{Proceedings of the IEEE Conference on Computer Vision and Pattern Recognition}, pages 326--335, 2017.

\bibitem[de~Vries et~al.(2019)de~Vries, Misra, Wang, and van~der Maaten]{devries2019does}
Terrance de~Vries, Ishan Misra, Changhan Wang, and Laurens van~der Maaten.
\newblock Does object recognition work for everyone?
\newblock In \emph{Proceedings of the IEEE/CVF Conference on Computer Vision and Pattern Recognition (CVPR) Workshops}, June 2019.

\bibitem[Deng et~al.(2009)Deng, Dong, Socher, Li, Li, and Fei-Fei]{imagenet}
Jia Deng, Wei Dong, Richard Socher, Li-Jia Li, Kai Li, and Li~Fei-Fei.
\newblock Imagenet: A large-scale hierarchical image database.
\newblock In \emph{2009 IEEE Conference on Computer Vision and Pattern Recognition}, pages 248--255, 2009.
\newblock \doi{10.1109/CVPR.2009.5206848}.

\bibitem[Derczynski et~al.(2023)Derczynski, Kirk, Balachandran, Kumar, Tsvetkov, Leiser, and Mohammad]{derczynski2023assessing}
Leon Derczynski, Hannah~Rose Kirk, Vidhisha Balachandran, Sachin Kumar, Yulia Tsvetkov, MR~Leiser, and Saif Mohammad.
\newblock Assessing language model deployment with risk cards.
\newblock \emph{arXiv preprint arXiv:2303.18190}, 2023.

\bibitem[Dettmers et~al.(2023)Dettmers, Pagnoni, Holtzman, and Zettlemoyer]{dettmers2024qlora}
Tim Dettmers, Artidoro Pagnoni, Ari Holtzman, and Luke Zettlemoyer.
\newblock {QL}o{RA}: Efficient finetuning of quantized {LLM}s.
\newblock In \emph{Thirty-seventh Conference on Neural Information Processing Systems}, 2023.
\newblock URL \url{https://openreview.net/forum?id=OUIFPHEgJU}.

\bibitem[Devlin et~al.(2019)Devlin, Chang, Lee, and Toutanova]{devlin-etal-2019-bert}
Jacob Devlin, Ming-Wei Chang, Kenton Lee, and Kristina Toutanova.
\newblock {BERT}: Pre-training of deep bidirectional transformers for language understanding.
\newblock In Jill Burstein, Christy Doran, and Thamar Solorio, editors, \emph{Proceedings of the 2019 Conference of the North {A}merican Chapter of the Association for Computational Linguistics: Human Language Technologies, Volume 1 (Long and Short Papers)}, pages 4171--4186, Minneapolis, Minnesota, June 2019. Association for Computational Linguistics.
\newblock \doi{10.18653/v1/N19-1423}.
\newblock URL \url{https://aclanthology.org/N19-1423}.

\bibitem[Diwan et~al.(2022)Diwan, Berry, Choi, Harwath, and Mahowald]{diwan2022winoground}
Anuj Diwan, Layne Berry, Eunsol Choi, David Harwath, and Kyle Mahowald.
\newblock Why is winoground hard? investigating failures in visuolinguistic compositionality.
\newblock In Yoav Goldberg, Zornitsa Kozareva, and Yue Zhang, editors, \emph{Proceedings of the 2022 Conference on Empirical Methods in Natural Language Processing}, pages 2236--2250, Abu Dhabi, United Arab Emirates, December 2022. Association for Computational Linguistics.
\newblock \doi{10.18653/v1/2022.emnlp-main.143}.
\newblock URL \url{https://aclanthology.org/2022.emnlp-main.143}.

\bibitem[Dosovitskiy et~al.(2021)Dosovitskiy, Beyer, Kolesnikov, Weissenborn, Zhai, Unterthiner, Dehghani, Minderer, Heigold, Gelly, Uszkoreit, and Houlsby]{dosovitskiy2020image}
Alexey Dosovitskiy, Lucas Beyer, Alexander Kolesnikov, Dirk Weissenborn, Xiaohua Zhai, Thomas Unterthiner, Mostafa Dehghani, Matthias Minderer, Georg Heigold, Sylvain Gelly, Jakob Uszkoreit, and Neil Houlsby.
\newblock An image is worth 16x16 words: Transformers for image recognition at scale.
\newblock In \emph{International Conference on Learning Representations}, 2021.
\newblock URL \url{https://openreview.net/forum?id=YicbFdNTTy}.

\bibitem[Du et~al.(2022)Du, Liu, Li, and Zhao]{du2022survey}
Yifan Du, Zikang Liu, Junyi Li, and Wayne~Xin Zhao.
\newblock A survey of vision-language pre-trained models, 2022.

\bibitem[Dubois et~al.(2021)Dubois, Bloem-Reddy, Ullrich, and Maddison]{dubois2021lossy}
Yann Dubois, Benjamin Bloem-Reddy, Karen Ullrich, and Chris~J Maddison.
\newblock Lossy compression for lossless prediction.
\newblock \emph{Advances in Neural Information Processing Systems}, 34:\penalty0 14014--14028, 2021.

\bibitem[Esser et~al.(2021)Esser, Rombach, and Ommer]{esser2021taming}
Patrick Esser, Robin Rombach, and Bjorn Ommer.
\newblock Taming transformers for high-resolution image synthesis.
\newblock In \emph{Proceedings of the IEEE/CVF Conference on Computer Vision and Pattern Recognition}, pages 12873--12883, 2021.

\bibitem[Everingham et~al.(2010)Everingham, Van~Gool, Williams, Winn, and Zisserman]{pascal_voc}
M.~Everingham, L.~Van~Gool, C.~K.~I. Williams, J.~Winn, and A.~Zisserman.
\newblock The pascal visual object classes (voc) challenge.
\newblock \emph{International Journal of Computer Vision}, 88\penalty0 (2):\penalty0 303--338, June 2010.

\bibitem[Fan et~al.(2021)Fan, Xiong, Mangalam, Li, Yan, Malik, and Feichtenhofer]{fan2021multiscale}
Haoqi Fan, Bo~Xiong, Karttikeya Mangalam, Yanghao Li, Zhicheng Yan, Jitendra Malik, and Christoph Feichtenhofer.
\newblock Multiscale vision transformers.
\newblock In \emph{Proceedings of the IEEE/CVF International Conference on Computer Vision}, pages 6824--6835, 2021.

\bibitem[Federici et~al.(2020)Federici, Dutta, Forré, Kushman, and Akata]{federici2020learning}
Marco Federici, Anjan Dutta, Patrick Forré, Nate Kushman, and Zeynep Akata.
\newblock Learning robust representations via multi-view information bottleneck.
\newblock In \emph{International Conference on Learning Representations}, 2020.
\newblock URL \url{https://openreview.net/forum?id=B1xwcyHFDr}.

\bibitem[Feichtenhofer et~al.(2022)Feichtenhofer, Fan, Li, and He]{feichtenhofer2021masked}
Christoph Feichtenhofer, Haoqi Fan, Yanghao Li, and Kaiming He.
\newblock Masked autoencoders as spatiotemporal learners.
\newblock In Alice~H. Oh, Alekh Agarwal, Danielle Belgrave, and Kyunghyun Cho, editors, \emph{Advances in Neural Information Processing Systems}, 2022.
\newblock URL \url{https://openreview.net/forum?id=UaXD4Al3mdb}.

\bibitem[Fini et~al.(2023)Fini, Astolfi, Romero-Soriano, Verbeek, and Drozdzal]{fini2023improved}
Enrico Fini, Pietro Astolfi, Adriana Romero-Soriano, Jakob Verbeek, and Michal Drozdzal.
\newblock Improved baselines for vision-language pre-training.
\newblock \emph{Transactions on Machine Learning Research}, 2023.
\newblock ISSN 2835-8856.
\newblock URL \url{https://openreview.net/forum?id=a7nvXxNmdV}.
\newblock Featured Certification.

\bibitem[Fisher(1936)]{fisher1936use}
Ronald~A Fisher.
\newblock The use of multiple measurements in taxonomic problems.
\newblock \emph{Annals of Eugenics}, 7\penalty0 (2):\penalty0 179--188, 1936.

\bibitem[Gadre et~al.(2023)Gadre, Ilharco, Fang, Hayase, Smyrnis, Nguyen, Marten, Wortsman, Ghosh, Zhang, Orgad, Entezari, Daras, Pratt, Ramanujan, Bitton, Marathe, Mussmann, Vencu, Cherti, Krishna, Koh, Saukh, Ratner, Song, Hajishirzi, Farhadi, Beaumont, Oh, Dimakis, Jitsev, Carmon, Shankar, and Schmidt]{gadre2023datacomp}
Samir~Yitzhak Gadre, Gabriel Ilharco, Alex Fang, Jonathan Hayase, Georgios Smyrnis, Thao Nguyen, Ryan Marten, Mitchell Wortsman, Dhruba Ghosh, Jieyu Zhang, Eyal Orgad, Rahim Entezari, Giannis Daras, Sarah~M Pratt, Vivek Ramanujan, Yonatan Bitton, Kalyani Marathe, Stephen Mussmann, Richard Vencu, Mehdi Cherti, Ranjay Krishna, Pang~Wei Koh, Olga Saukh, Alexander Ratner, Shuran Song, Hannaneh Hajishirzi, Ali Farhadi, Romain Beaumont, Sewoong Oh, Alex Dimakis, Jenia Jitsev, Yair Carmon, Vaishaal Shankar, and Ludwig Schmidt.
\newblock Datacomp: In search of the next generation of multimodal datasets.
\newblock In \emph{Thirty-seventh Conference on Neural Information Processing Systems Datasets and Benchmarks Track}, 2023.
\newblock URL \url{https://openreview.net/forum?id=dVaWCDMBof}.

\bibitem[Gafni et~al.(2022)Gafni, Polyak, Ashual, Sheynin, Parikh, and Taigman]{gafni2022make}
Oran Gafni, Adam Polyak, Oron Ashual, Shelly Sheynin, Devi Parikh, and Yaniv Taigman.
\newblock Make-a-scene: Scene-based text-to-image generation with human priors.
\newblock In \emph{European Conference on Computer Vision}, pages 89--106. Springer, 2022.

\bibitem[Gao et~al.(2023)Gao, Han, Zhang, Lin, Geng, Zhou, Zhang, Lu, He, Yue, et~al.]{gao2023llama}
Peng Gao, Jiaming Han, Renrui Zhang, Ziyi Lin, Shijie Geng, Aojun Zhou, Wei Zhang, Pan Lu, Conghui He, Xiangyu Yue, et~al.
\newblock Llama-adapter v2: Parameter-efficient visual instruction model.
\newblock \emph{arXiv preprint arXiv:2304.15010}, 2023.

\bibitem[Gao et~al.(2024)Gao, Geng, Zhang, Ma, Fang, Zhang, Li, and Qiao]{gao2024clip}
Peng Gao, Shijie Geng, Renrui Zhang, Teli Ma, Rongyao Fang, Yongfeng Zhang, Hongsheng Li, and Yu~Qiao.
\newblock Clip-adapter: Better vision-language models with feature adapters.
\newblock \emph{International Journal of Computer Vision}, 132\penalty0 (2):\penalty0 581--595, 2024.

\bibitem[Garcia et~al.(2023)Garcia, Hirota, Wu, and Nakashima]{garcia2023uncurated}
Noa Garcia, Yusuke Hirota, Yankun Wu, and Yuta Nakashima.
\newblock Uncurated image-text datasets: Shedding light on demographic bias.
\newblock In \emph{Proceedings of the IEEE/CVF Conference on Computer Vision and Pattern Recognition}, pages 6957--6966, 2023.

\bibitem[Ghosh et~al.(2024)Ghosh, Acharya, Saha, Jain, and Chadha]{ghosh2024exploring}
Akash Ghosh, Arkadeep Acharya, Sriparna Saha, Vinija Jain, and Aman Chadha.
\newblock Exploring the frontier of vision-language models: A survey of current methodologies and future directions, 2024.

\bibitem[Goel et~al.(2022)Goel, Bansal, Bhatia, Rossi, Vinay, and Grover]{goel2022cyclip}
Shashank Goel, Hritik Bansal, Sumit Bhatia, Ryan Rossi, Vishwa Vinay, and Aditya Grover.
\newblock Cyclip: Cyclic contrastive language-image pretraining.
\newblock \emph{Advances in Neural Information Processing Systems}, 35:\penalty0 6704--6719, 2022.

\bibitem[Goyal et~al.(2022)Goyal, Soriano, Hazirbas, Sagun, and Usunier]{goyal2022fairness}
Priya Goyal, Adriana~Romero Soriano, Caner Hazirbas, Levent Sagun, and Nicolas Usunier.
\newblock Fairness indicators for systematic assessments of visual feature extractors.
\newblock In \emph{Proceedings of the 2022 ACM Conference on Fairness, Accountability, and Transparency}, pages 70--88, 2022.

\bibitem[Goyal et~al.(2017)Goyal, Khot, Summers-Stay, Batra, and Parikh]{vqa2}
Yash Goyal, Tejas Khot, Douglas Summers-Stay, Dhruv Batra, and Devi Parikh.
\newblock Making the {V} in {VQA} matter: Elevating the role of image understanding in visual question answering.
\newblock In \emph{Proceedings of the IEEE Conference on Computer Vision and Pattern Recognition}, pages 6904--6913, 2017.

\bibitem[Grill et~al.(2020)Grill, Strub, Altch{\'e}, Tallec, Richemond, Buchatskaya, Doersch, Avila~Pires, Guo, Gheshlaghi~Azar, et~al.]{grill2020bootstrap}
Jean-Bastien Grill, Florian Strub, Florent Altch{\'e}, Corentin Tallec, Pierre Richemond, Elena Buchatskaya, Carl Doersch, Bernardo Avila~Pires, Zhaohan Guo, Mohammad Gheshlaghi~Azar, et~al.
\newblock Bootstrap your own latent-a new approach to self-supervised learning.
\newblock \emph{Advances in Neural Information Processing Systems}, 33:\penalty0 21271--21284, 2020.

\bibitem[Gunjal et~al.(2024)Gunjal, Yin, and Bas]{gunjal2023detecting}
Anisha Gunjal, Jihan Yin, and Erhan Bas.
\newblock Detecting and preventing hallucinations in large vision language models.
\newblock In \emph{Proceedings of the AAAI Conference on Artificial Intelligence}, volume~38, pages 18135--18143, 2024.

\bibitem[Gurari et~al.(2018)Gurari, Li, Stangl, Guo, Lin, Grauman, Luo, and Bigham]{gurari2018vizwiz}
Danna Gurari, Qing Li, Abigale~J Stangl, Anhong Guo, Chi Lin, Kristen Grauman, Jiebo Luo, and Jeffrey~P Bigham.
\newblock Vizwiz grand challenge: Answering visual questions from blind people.
\newblock In \emph{Proceedings of the IEEE Conference on Computer Vision and Pattern Recognition}, pages 3608--3617, 2018.

\bibitem[Gustafson et~al.(2023)Gustafson, Rolland, Ravi, Duval, Adcock, Fu, Hall, and Ross]{Gustafson_2023_ICCV}
Laura Gustafson, Chloe Rolland, Nikhila Ravi, Quentin Duval, Aaron Adcock, Cheng-Yang Fu, Melissa Hall, and Candace Ross.
\newblock Facet: Fairness in computer vision evaluation benchmark.
\newblock In \emph{Proceedings of the IEEE/CVF International Conference on Computer Vision (ICCV)}, pages 20370--20382, October 2023.

\bibitem[Gutmann and Hyv{\"a}rinen(2010)]{nce}
Michael Gutmann and Aapo Hyv{\"a}rinen.
\newblock Noise-contrastive estimation: {{A}} new estimation principle for unnormalized statistical models.
\newblock In \emph{Proceedings of the {{Thirteenth International Conference}} on {{Artificial Intelligence}} and {{Statistics}}}, pages 297--304. {JMLR Workshop and Conference Proceedings}, March 2010.

\bibitem[Hall et~al.(2023{\natexlab{a}})Hall, Chern, Gustafson, Ventura, Kulkarni, Ross, and Usunier]{hall2023reliable}
Melissa Hall, Bobbie Chern, Laura Gustafson, Denisse Ventura, Harshad Kulkarni, Candace Ross, and Nicolas Usunier.
\newblock Towards reliable assessments of demographic disparities in multi-label image classifiers, 2023{\natexlab{a}}.

\bibitem[Hall et~al.(2023{\natexlab{b}})Hall, Gustafson, Adcock, Misra, and Ross]{Hall_2023_ICCV}
Melissa Hall, Laura Gustafson, Aaron Adcock, Ishan Misra, and Candace Ross.
\newblock Vision-language models performing zero-shot tasks exhibit disparities between gender groups.
\newblock In \emph{Proceedings of the IEEE/CVF International Conference on Computer Vision (ICCV) Workshops}, pages 2778--2785, October 2023{\natexlab{b}}.

\bibitem[Hamidieh et~al.(2023)Hamidieh, Zhang, Hartvigsen, and Ghassemi]{hamidieh2023identifying}
Kimia Hamidieh, Haoran Zhang, Thomas Hartvigsen, and Marzyeh Ghassemi.
\newblock Identifying implicit social biases in vision-language models, 2023.

\bibitem[Hammoud et~al.(2024)Hammoud, Itani, Pizzati, Torr, Bibi, and Ghanem]{hammoud2024synthclip}
Hasan Abed Al~Kader Hammoud, Hani Itani, Fabio Pizzati, Philip Torr, Adel Bibi, and Bernard Ghanem.
\newblock Synthclip: Are we ready for a fully synthetic clip training?
\newblock \emph{arXiv preprint arXiv:2402.01832}, 2024.

\bibitem[Hazirbas et~al.(2024)Hazirbas, Sun, Efroni, and Ibrahim]{hazirbas2024bias}
Caner Hazirbas, Alicia Sun, Yonathan Efroni, and Mark Ibrahim.
\newblock The bias of harmful label associations in vision-language models.
\newblock \emph{arXiv preprint arXiv: 2402.07329}, 2024.

\bibitem[He et~al.(2015)He, Zhang, Ren, and Sun]{resnet}
Kaiming He, Xiangyu Zhang, Shaoqing Ren, and Jian Sun.
\newblock Deep residual learning for image recognition.
\newblock \emph{arXiv preprint arXiv:1512.03385}, 2015.

\bibitem[He et~al.(2020)He, Fan, Wu, Xie, and Girshick]{he2020momentum}
Kaiming He, Haoqi Fan, Yuxin Wu, Saining Xie, and Ross Girshick.
\newblock Momentum contrast for unsupervised visual representation learning.
\newblock In \emph{Proceedings of the IEEE/CVF Conference on Computer Vision and Pattern Recognition}, pages 9729--9738, 2020.

\bibitem[He et~al.(2022)He, Chen, Xie, Li, Doll{\'a}r, and Girshick]{he2022masked}
Kaiming He, Xinlei Chen, Saining Xie, Yanghao Li, Piotr Doll{\'a}r, and Ross Girshick.
\newblock Masked autoencoders are scalable vision learners.
\newblock In \emph{Proceedings of the IEEE/CVF Conference on Computer Vision and Pattern Recognition}, pages 16000--16009, 2022.

\bibitem[Helber et~al.(2019)Helber, Bischke, Dengel, and Borth]{Eurosat}
Patrick Helber, Benjamin Bischke, Andreas Dengel, and Damian Borth.
\newblock Eurosat: A novel dataset and deep learning benchmark for land use and land cover classification.
\newblock \emph{IEEE Journal of Selected Topics in Applied Earth Observations and Remote Sensing}, 12\penalty0 (7):\penalty0 2217--2226, 2019.

\bibitem[Hemmat et~al.(2023)Hemmat, Pezeshki, Bordes, Drozdzal, and Romero-Soriano]{hemmat2023feedback}
Reyhane~Askari Hemmat, Mohammad Pezeshki, Florian Bordes, Michal Drozdzal, and Adriana Romero-Soriano.
\newblock Feedback-guided data synthesis for imbalanced classification.
\newblock \emph{arXiv preprint arXiv:2310.00158}, 2023.

\bibitem[Henighan et~al.(2020{\natexlab{a}})Henighan, Kaplan, Katz, Chen, Hesse, Jackson, Jun, Brown, Dhariwal, Gray, Hallacy, Mann, Radford, Ramesh, Ryder, Ziegler, Schulman, Amodei, and McCandlish]{scaling_law_gen}
Tom Henighan, Jared Kaplan, Mor Katz, Mark Chen, Christopher Hesse, Jacob Jackson, Heewoo Jun, Tom~B. Brown, Prafulla Dhariwal, Scott Gray, Chris Hallacy, Benjamin Mann, Alec Radford, Aditya Ramesh, Nick Ryder, Daniel~M. Ziegler, John Schulman, Dario Amodei, and Sam McCandlish.
\newblock Scaling laws for autoregressive generative modeling.
\newblock \emph{ArXiv}, abs/2010.14701, 2020{\natexlab{a}}.
\newblock URL \url{https://api.semanticscholar.org/CorpusID:225094178}.

\bibitem[Henighan et~al.(2020{\natexlab{b}})Henighan, Kaplan, Katz, Chen, Hesse, Jackson, Jun, Brown, Dhariwal, Gray, et~al.]{scaling_law}
Tom Henighan, Jared Kaplan, Mor Katz, Mark Chen, Christopher Hesse, Jacob Jackson, Heewoo Jun, Tom~B Brown, Prafulla Dhariwal, Scott Gray, et~al.
\newblock Scaling laws for autoregressive generative modeling.
\newblock \emph{arXiv preprint arXiv:2010.14701}, 2020{\natexlab{b}}.

\bibitem[Hessel et~al.(2021)Hessel, Holtzman, Forbes, Le~Bras, and Choi]{hessel2021clipscore}
Jack Hessel, Ari Holtzman, Maxwell Forbes, Ronan Le~Bras, and Yejin Choi.
\newblock {CLIPS}core: A reference-free evaluation metric for image captioning.
\newblock In Marie-Francine Moens, Xuanjing Huang, Lucia Specia, and Scott Wen-tau Yih, editors, \emph{Proceedings of the 2021 Conference on Empirical Methods in Natural Language Processing}, pages 7514--7528, Online and Punta Cana, Dominican Republic, November 2021. Association for Computational Linguistics.
\newblock \doi{10.18653/v1/2021.emnlp-main.595}.
\newblock URL \url{https://aclanthology.org/2021.emnlp-main.595}.

\bibitem[Hong et~al.(2023)Hong, Wang, Lv, Xu, Yu, Ji, Wang, Wang, Dong, Ding, and Tang]{hong2023cogagent}
Wenyi Hong, Weihan Wang, Qingsong Lv, Jiazheng Xu, Wenmeng Yu, Junhui Ji, Yan Wang, Zihan Wang, Yuxiao Dong, Ming Ding, and Jie Tang.
\newblock Cogagent: A visual language model for gui agents, 2023.

\bibitem[Honnibal and Montani(2017)]{spacy2}
Matthew Honnibal and Ines Montani.
\newblock {spaCy 2}: Natural language understanding with {B}loom embeddings, convolutional neural networks and incremental parsing, 2017.

\bibitem[Houlsby et~al.(2019)Houlsby, Giurgiu, Jastrzebski, Morrone, De~Laroussilhe, Gesmundo, Attariyan, and Gelly]{houlsby2019parameter}
Neil Houlsby, Andrei Giurgiu, Stanislaw Jastrzebski, Bruna Morrone, Quentin De~Laroussilhe, Andrea Gesmundo, Mona Attariyan, and Sylvain Gelly.
\newblock Parameter-efficient transfer learning for {NLP}.
\newblock In \emph{International Conference on Machine Learning}, pages 2790--2799. PMLR, 2019.

\bibitem[Hsieh et~al.(2023)Hsieh, Zhang, Ma, Kembhavi, and Krishna]{sugarcrepe}
Cheng-Yu Hsieh, Jieyu Zhang, Zixian Ma, Aniruddha Kembhavi, and Ranjay Krishna.
\newblock Sugarcrepe: Fixing hackable benchmarks for vision-language compositionality.
\newblock In \emph{Thirty-seventh Conference on Neural Information Processing Systems Datasets and Benchmarks Track}, 2023.
\newblock URL \url{https://openreview.net/forum?id=Jsc7WSCZd4}.

\bibitem[Hu et~al.(2022)Hu, Shen, Wallis, Allen-Zhu, Li, Wang, Wang, and Chen]{hu2021lora}
Edward~J Hu, Yelong Shen, Phillip Wallis, Zeyuan Allen-Zhu, Yuanzhi Li, Shean Wang, Lu~Wang, and Weizhu Chen.
\newblock Lo{RA}: Low-rank adaptation of large language models.
\newblock In \emph{International Conference on Learning Representations}, 2022.
\newblock URL \url{https://openreview.net/forum?id=nZeVKeeFYf9}.

\bibitem[Hu et~al.(2023)Hu, Liu, Kasai, Wang, Ostendorf, Krishna, and Smith]{hu2023tifa}
Yushi Hu, Benlin Liu, Jungo Kasai, Yizhong Wang, Mari Ostendorf, Ranjay Krishna, and Noah~A Smith.
\newblock Tifa: Accurate and interpretable text-to-image faithfulness evaluation with question answering.
\newblock In \emph{Proceedings of the IEEE/CVF International Conference on Computer Vision}, pages 20406--20417, 2023.

\bibitem[Huang et~al.(2023)Huang, Yu, Ma, Zhong, Feng, Wang, Chen, Peng, Feng, Qin, et~al.]{huang2023survey}
Lei Huang, Weijiang Yu, Weitao Ma, Weihong Zhong, Zhangyin Feng, Haotian Wang, Qianglong Chen, Weihua Peng, Xiaocheng Feng, Bing Qin, et~al.
\newblock A survey on hallucination in large language models: Principles, taxonomy, challenges, and open questions.
\newblock \emph{arXiv preprint arXiv:2311.05232}, 2023.

\bibitem[Huang et~al.(2019)Huang, Chen, He, Bai, Karatzas, Lu, and Jawahar]{Huang_2019}
Zheng Huang, Kai Chen, Jianhua He, Xiang Bai, Dimosthenis Karatzas, Shijian Lu, and C.~V. Jawahar.
\newblock {ICDAR} 2019 competition on scanned receipt ocr and information extraction.
\newblock In \emph{2019 International Conference on Document Analysis and Recognition (ICDAR)}. IEEE, September 2019.
\newblock \doi{10.1109/icdar.2019.00244}.
\newblock URL \url{http://dx.doi.org/10.1109/ICDAR.2019.00244}.

\bibitem[Hudson and Manning(2019)]{hudson2019gqa}
Drew~A Hudson and Christopher~D Manning.
\newblock {GQA}: A new dataset for real-world visual reasoning and compositional question answering.
\newblock In \emph{Proceedings of the IEEE/CVF Conference on Computer Vision and Pattern Recognition}, pages 6700--6709, 2019.

\bibitem[Hyv{\"a}rinen(2005)]{hyvarinen2005estimation}
Aapo Hyv{\"a}rinen.
\newblock Estimation of non-normalized statistical models by score matching.
\newblock \emph{Journal of Machine Learning Research}, 6\penalty0 (Apr):\penalty0 695--709, 2005.

\bibitem[Ilharco et~al.(2021)Ilharco, Wortsman, Wightman, Gordon, Carlini, Taori, Dave, Shankar, Namkoong, Miller, Hajishirzi, Farhadi, and Schmidt]{openclip}
Gabriel Ilharco, Mitchell Wortsman, Ross Wightman, Cade Gordon, Nicholas Carlini, Rohan Taori, Achal Dave, Vaishaal Shankar, Hongseok Namkoong, John Miller, Hannaneh Hajishirzi, Ali Farhadi, and Ludwig Schmidt.
\newblock Openclip, July 2021.
\newblock URL \url{https://doi.org/10.5281/zenodo.5143773}.
\newblock If you use this software, please cite it as below.

\bibitem[Jaini et~al.(2024)Jaini, Clark, and Geirhos]{jaini2023intriguing}
Priyank Jaini, Kevin Clark, and Robert Geirhos.
\newblock Intriguing properties of generative classifiers.
\newblock In \emph{The Twelfth International Conference on Learning Representations}, 2024.
\newblock URL \url{https://openreview.net/forum?id=rmg0qMKYRQ}.

\bibitem[Jassim et~al.(2023)Jassim, Holubar, Richter, Wolff, Ohmer, and Bruni]{jassim2023grasp}
Serwan Jassim, Mario Holubar, Annika Richter, Cornelius Wolff, Xenia Ohmer, and Elia Bruni.
\newblock Grasp: A novel benchmark for evaluating language grounding and situated physics understanding in multimodal language models.
\newblock \emph{arXiv preprint arXiv:2311.09048}, 2023.

\bibitem[Jaume et~al.(2019)Jaume, Ekenel, and Thiran]{jaume2019funsd}
Guillaume Jaume, Hazim~Kemal Ekenel, and Jean-Philippe Thiran.
\newblock Funsd: A dataset for form understanding in noisy scanned documents.
\newblock In \emph{2019 International Conference on Document Analysis and Recognition Workshops (ICDARW)}, volume~2, pages 1--6. IEEE, 2019.

\bibitem[Jayaraman et~al.(2024)Jayaraman, Guo, and Chaudhuri]{jayaraman2024deja}
Bargav Jayaraman, Chuan Guo, and Kamalika Chaudhuri.
\newblock D\'ej\`a vu memorization in vision-language models.
\newblock \emph{arXiv preprint arXiv:2402.02103}, 2024.

\bibitem[Jia et~al.(2022)Jia, Tang, Chen, Cardie, Belongie, Hariharan, and Lim]{jia2022visual}
Menglin Jia, Luming Tang, Bor-Chun Chen, Claire Cardie, Serge Belongie, Bharath Hariharan, and Ser-Nam Lim.
\newblock Visual prompt tuning.
\newblock In \emph{European Conference on Computer Vision}, pages 709--727. Springer, 2022.

\bibitem[Johnson et~al.(2019)Johnson, Douze, and J{\'e}gou]{faiss}
Jeff Johnson, Matthijs Douze, and Herv{\'e} J{\'e}gou.
\newblock Billion-scale similarity search with gpus.
\newblock \emph{IEEE Transactions on Big Data}, 7\penalty0 (3):\penalty0 535--547, 2019.

\bibitem[Joulin et~al.(2017)Joulin, Grave, Bojanowski, and Mikolov]{fasttext}
Armand Joulin, Edouard Grave, Piotr Bojanowski, and Tomas Mikolov.
\newblock Bag of tricks for efficient text classification.
\newblock In Mirella Lapata, Phil Blunsom, and Alexander Koller, editors, \emph{Proceedings of the 15th Conference of the {E}uropean Chapter of the Association for Computational Linguistics: Volume 2, Short Papers}, pages 427--431, Valencia, Spain, April 2017. Association for Computational Linguistics.
\newblock URL \url{https://aclanthology.org/E17-2068}.

\bibitem[Karatzas et~al.(2013)Karatzas, Shafait, Uchida, Iwamura, i~Bigorda, Mestre, Romeu, Mota, Almaz{\'a}n, and de~las Heras]{IC13}
Dimosthenis Karatzas, Faisal Shafait, Seiichi Uchida, M.~Iwamura, Llu{\'i}s~G{\'o}mez i~Bigorda, Sergi~Robles Mestre, Joan~Mas Romeu, David~Fern{\'a}ndez Mota, Jon Almaz{\'a}n, and Llu{\'i}s-Pere de~las Heras.
\newblock {ICDAR} 2013 robust reading competition.
\newblock In \emph{2013 12th International Conference on Document Analysis and Recognition}, pages 1484--1493, 2013.

\bibitem[Karkkainen and Joo(2021)]{karkkainenfairface}
Kimmo Karkkainen and Jungseock Joo.
\newblock Fairface: Face attribute dataset for balanced race, gender, and age for bias measurement and mitigation.
\newblock In \emph{Proceedings of the IEEE/CVF Winter Conference on Applications of Computer Vision}, pages 1548--1558, 2021.

\bibitem[Ke et~al.(2023)Ke, Ye, Yu, Wu, Milanfar, and Yang]{ke2023vila}
Junjie Ke, Keren Ye, Jiahui Yu, Yonghui Wu, Peyman Milanfar, and Feng Yang.
\newblock {VILA}: Learning image aesthetics from user comments with vision-language pretraining.
\newblock In \emph{Proceedings of the IEEE/CVF Conference on Computer Vision and Pattern Recognition}, pages 10041--10051, 2023.

\bibitem[Kirillov et~al.(2023)Kirillov, Mintun, Ravi, Mao, Rolland, Gustafson, Xiao, Whitehead, Berg, Lo, Dollar, and Girshick]{kirillov2023segany}
Alexander Kirillov, Eric Mintun, Nikhila Ravi, Hanzi Mao, Chloe Rolland, Laura Gustafson, Tete Xiao, Spencer Whitehead, Alexander~C. Berg, Wan-Yen Lo, Piotr Dollar, and Ross Girshick.
\newblock Segment anything.
\newblock In \emph{Proceedings of the IEEE/CVF International Conference on Computer Vision (ICCV)}, pages 4015--4026, October 2023.

\bibitem[Kopiczko et~al.(2024)Kopiczko, Blankevoort, and Asano]{kopiczko2023vera}
Dawid~Jan Kopiczko, Tijmen Blankevoort, and Yuki~M Asano.
\newblock Ve{RA}: Vector-based random matrix adaptation.
\newblock In \emph{The Twelfth International Conference on Learning Representations}, 2024.
\newblock URL \url{https://openreview.net/forum?id=NjNfLdxr3A}.

\bibitem[Krause et~al.(2013)Krause, Stark, Deng, and Fei-Fei]{StanfordCars}
Jonathan Krause, Michael Stark, Jia Deng, and Li~Fei-Fei.
\newblock 3d object representations for fine-grained categorization.
\newblock In \emph{Proceedings - 2013 IEEE International Conference on Computer Vision Workshops, ICCVW 2013}, Proceedings of the IEEE International Conference on Computer Vision, pages 554--561, United States, 2013. Institute of Electrical and Electronics Engineers Inc.
\newblock ISBN 9781479930227.
\newblock \doi{10.1109/ICCVW.2013.77}.
\newblock 2013 14th IEEE International Conference on Computer Vision Workshops, ICCVW 2013 ; Conference date: 01-12-2013 Through 08-12-2013.

\bibitem[Krause et~al.(2017)Krause, Johnson, Krishna, and Fei-Fei]{krause2017hierarchical}
Jonathan Krause, Justin Johnson, Ranjay Krishna, and Li~Fei-Fei.
\newblock A hierarchical approach for generating descriptive image paragraphs.
\newblock In \emph{Proceedings of the IEEE Conference on Computer Vision and Pattern Recognition}, pages 317--325, 2017.

\bibitem[Krishna et~al.(2017)Krishna, Zhu, Groth, Johnson, Hata, Kravitz, Chen, Kalantidis, Li, Shamma, et~al.]{krishna2017visual}
Ranjay Krishna, Yuke Zhu, Oliver Groth, Justin Johnson, Kenji Hata, Joshua Kravitz, Stephanie Chen, Yannis Kalantidis, Li-Jia Li, David~A Shamma, et~al.
\newblock Visual genome: Connecting language and vision using crowdsourced dense image annotations.
\newblock \emph{International Journal of Computer Vision}, 123:\penalty0 32--73, 2017.

\bibitem[Krizhevsky(2009)]{krizhevsky2009learning}
Alex Krizhevsky.
\newblock Learning multiple layers of features from tiny images, 2009.
\newblock URL \url{https://www.cs.toronto.edu/~kriz/learning-features-2009-TR.pdf}.

\bibitem[Kuang et~al.(2023)Kuang, Hua, Liang, Yang, Jiang, Ren, and Bai]{kuang2023visual}
Jianfeng Kuang, Wei Hua, Dingkang Liang, Mingkun Yang, Deqiang Jiang, Bo~Ren, and Xiang Bai.
\newblock Visual information extraction in the wild: Practical dataset and end-to-end solution.
\newblock In Gernot~A. Fink, Rajiv Jain, Koichi Kise, and Richard Zanibbi, editors, \emph{Document Analysis and Recognition -- ICDAR 2023}, pages 36--53, Cham, 2023. Springer Nature Switzerland.
\newblock ISBN 978-3-031-41731-3.

\bibitem[Kuang et~al.(2021)Kuang, Sun, Li, Yue, Lin, Chen, Wei, Zhu, Gao, Zhang, et~al.]{kuang2021mmocr}
Zhanghui Kuang, Hongbin Sun, Zhizhong Li, Xiaoyu Yue, Tsui~Hin Lin, Jianyong Chen, Huaqiang Wei, Yiqin Zhu, Tong Gao, Wenwei Zhang, et~al.
\newblock {MMOCR}: a comprehensive toolbox for text detection, recognition and understanding.
\newblock In \emph{Proceedings of the 29th ACM International Conference on Multimedia}, pages 3791--3794, 2021.

\bibitem[Kwon et~al.(2023)Kwon, Cai, Ravichandran, Bas, Bhotika, and Soatto]{kwon2023masked}
Gukyeong Kwon, Zhaowei Cai, Avinash Ravichandran, Erhan Bas, Rahul Bhotika, and Stefano Soatto.
\newblock Masked vision and language modeling for multi-modal representation learning.
\newblock In \emph{The Eleventh International Conference on Learning Representations}, 2023.
\newblock URL \url{https://openreview.net/forum?id=ZhuXksSJYWn}.

\bibitem[Lauren{\c{c}}on et~al.(2023)Lauren{\c{c}}on, Saulnier, Tronchon, Bekman, Singh, Lozhkov, Wang, Karamcheti, Rush, Kiela, Cord, and Sanh]{laurençon2023obelics}
Hugo Lauren{\c{c}}on, Lucile Saulnier, Leo Tronchon, Stas Bekman, Amanpreet Singh, Anton Lozhkov, Thomas Wang, Siddharth Karamcheti, Alexander~M Rush, Douwe Kiela, Matthieu Cord, and Victor Sanh.
\newblock {OBELICS}: An open web-scale filtered dataset of interleaved image-text documents.
\newblock In \emph{Thirty-seventh Conference on Neural Information Processing Systems Datasets and Benchmarks Track}, 2023.
\newblock URL \url{https://openreview.net/forum?id=SKN2hflBIZ}.

\bibitem[Lavoie et~al.(2024)Lavoie, Kirichenko, Ibrahim, Assran, Wildon, Courville, and Ballas]{lavoie2024modeling}
Samuel Lavoie, Polina Kirichenko, Mark Ibrahim, Mahmoud Assran, Andrew~Gordon Wildon, Aaron Courville, and Nicolas Ballas.
\newblock Modeling caption diversity in contrastive vision-language pretraining.
\newblock \emph{arXiv preprint arXiv:2405.00740}, 2024.

\bibitem[Leclerc et~al.(2023)Leclerc, Ilyas, Engstrom, Park, Salman, and M{\k{a}}dry]{leclerc2023ffcv}
Guillaume Leclerc, Andrew Ilyas, Logan Engstrom, Sung~Min Park, Hadi Salman, and Aleksander M{\k{a}}dry.
\newblock {FFCV}: Accelerating training by removing data bottlenecks.
\newblock In \emph{Proceedings of the IEEE/CVF Conference on Computer Vision and Pattern Recognition}, pages 12011--12020, 2023.

\bibitem[LeCun and Bengio(1998)]{CNN}
Yann LeCun and Yoshua Bengio.
\newblock \emph{Convolutional Networks for Images, Speech, and Time Series}, page 255–258.
\newblock MIT Press, Cambridge, MA, USA, 1998.
\newblock ISBN 0262511029.

\bibitem[LeCun et~al.(2006)LeCun, Chopra, Hadsell, Ranzato, and Huang]{lecun_energy}
Yann LeCun, Sumit Chopra, Raia Hadsell, {Marc Aurelio} Ranzato, and {Fu Jie} Huang.
\newblock A tutorial on energy-based learning, 2006.

\bibitem[Lee et~al.(2024)Lee, Yasunaga, Meng, Mai, Park, Gupta, Zhang, Narayanan, Teufel, Bellagente, et~al.]{lee2024holistic}
Tony Lee, Michihiro Yasunaga, Chenlin Meng, Yifan Mai, Joon~Sung Park, Agrim Gupta, Yunzhi Zhang, Deepak Narayanan, Hannah Teufel, Marco Bellagente, et~al.
\newblock Holistic evaluation of text-to-image models.
\newblock \emph{Advances in Neural Information Processing Systems}, 36, 2024.

\bibitem[Lefaudeux et~al.(2022)Lefaudeux, Massa, Liskovich, Xiong, Caggiano, Naren, Xu, Hu, Tintore, Zhang, Labatut, Haziza, Wehrstedt, Reizenstein, and Sizov]{xFormers2022}
Benjamin Lefaudeux, Francisco Massa, Diana Liskovich, Wenhan Xiong, Vittorio Caggiano, Sean Naren, Min Xu, Jieru Hu, Marta Tintore, Susan Zhang, Patrick Labatut, Daniel Haziza, Luca Wehrstedt, Jeremy Reizenstein, and Grigory Sizov.
\newblock xformers: A modular and hackable transformer modelling library.
\newblock \url{https://github.com/facebookresearch/xformers}, 2022.

\bibitem[Lei et~al.(2018)Lei, Yu, Bansal, and Berg]{lei2018tvqa}
Jie Lei, Licheng Yu, Mohit Bansal, and Tamara~L Berg.
\newblock Tvqa: Localized, compositional video question answering.
\newblock In \emph{EMNLP}, 2018.

\bibitem[Li et~al.(2023{\natexlab{a}})Li, Prabhudesai, Duggal, Brown, and Pathak]{li2023diffusion}
Alexander~C. Li, Mihir Prabhudesai, Shivam Duggal, Ellis Brown, and Deepak Pathak.
\newblock Your diffusion model is secretly a zero-shot classifier.
\newblock In \emph{Proceedings of the IEEE/CVF International Conference on Computer Vision (ICCV)}, pages 2206--2217, October 2023{\natexlab{a}}.

\bibitem[Li et~al.(2024{\natexlab{a}})Li, Lin, Pathak, Li, Fei, Wu, Xia, Zhang, Neubig, and Ramanan]{li2024evaluating}
Baiqi Li, Zhiqiu Lin, Deepak Pathak, Jiayao Li, Yixin Fei, Kewen Wu, Xide Xia, Pengchuan Zhang, Graham Neubig, and Deva Ramanan.
\newblock Evaluating and improving compositional text-to-visual generation.
\newblock In \emph{Proceedings of the IEEE/CVF Conference on Computer Vision and Pattern Recognition (CVPR) Workshops}, 2024{\natexlab{a}}.

\bibitem[Li et~al.(2023{\natexlab{b}})Li, Zhang, Chen, Wang, Pu, Yang, Li, and Liu]{li2023mimicit}
Bo~Li, Yuanhan Zhang, Liangyu Chen, Jinghao Wang, Fanyi Pu, Jingkang Yang, Chunyuan Li, and Ziwei Liu.
\newblock {MIMIC-IT}: Multi-modal in-context instruction tuning.
\newblock \emph{arXiv preprint arXiv:2306.05425}, 2023{\natexlab{b}}.

\bibitem[Li et~al.(2023{\natexlab{c}})Li, Zhang, Chen, Wang, Yang, and Liu]{li2023otter}
Bo~Li, Yuanhan Zhang, Liangyu Chen, Jinghao Wang, Jingkang Yang, and Ziwei Liu.
\newblock Otter: A multi-modal model with in-context instruction tuning.
\newblock \emph{arXiv preprint arXiv:2305.03726}, 2023{\natexlab{c}}.

\bibitem[Li et~al.(2023{\natexlab{d}})Li, Ge, Li, and Shan]{li2023visionlanguage}
Chen Li, Yixiao Ge, Dian Li, and Ying Shan.
\newblock Vision-language instruction tuning: A review and analysis.
\newblock \emph{arXiv preprint arXiv: 2311.08172}, 2023{\natexlab{d}}.

\bibitem[Li et~al.(2022{\natexlab{a}})Li, Andreeto, Ranzato, and Perona]{li_andreeto_ranzato_perona_2022}
Fei-Fei Li, Marco Andreeto, Marc'Aurelio Ranzato, and Pietro Perona.
\newblock Caltech 101, avr 2022{\natexlab{a}}.

\bibitem[Li et~al.(2022{\natexlab{b}})Li, Li, Xiong, and Hoi]{li2022blip}
Junnan Li, Dongxu Li, Caiming Xiong, and Steven Hoi.
\newblock {BLIP}: Bootstrapping language-image pre-training for unified vision-language understanding and generation.
\newblock In \emph{International Conference on Machine Learning}, pages 12888--12900. PMLR, 2022{\natexlab{b}}.

\bibitem[Li et~al.(2023{\natexlab{e}})Li, Li, Savarese, and Hoi]{blip2}
Junnan Li, Dongxu Li, Silvio Savarese, and Steven Hoi.
\newblock {BLIP}-2: Bootstrapping language-image pre-training with frozen image encoders and large language models.
\newblock In Andreas Krause, Emma Brunskill, Kyunghyun Cho, Barbara Engelhardt, Sivan Sabato, and Jonathan Scarlett, editors, \emph{Proceedings of the 40th International Conference on Machine Learning}, volume 202 of \emph{Proceedings of Machine Learning Research}, pages 19730--19742. PMLR, 23--29 Jul 2023{\natexlab{e}}.
\newblock URL \url{https://proceedings.mlr.press/v202/li23q.html}.

\bibitem[Li et~al.(2019)Li, Yatskar, Yin, Hsieh, and Chang]{li2019visualbert}
Liunian~Harold Li, Mark Yatskar, Da~Yin, Cho-Jui Hsieh, and Kai-Wei Chang.
\newblock {VisualBERT}: A simple and performant baseline for vision and language.
\newblock \emph{arXiv preprint arXiv:1908.03557}, 2019.

\bibitem[Li et~al.(2022{\natexlab{c}})Li, Zhang, Zhang, Yang, Li, Zhong, Wang, Yuan, Zhang, Hwang, et~al.]{li2022grounded}
Liunian~Harold Li, Pengchuan Zhang, Haotian Zhang, Jianwei Yang, Chunyuan Li, Yiwu Zhong, Lijuan Wang, Lu~Yuan, Lei Zhang, Jenq-Neng Hwang, et~al.
\newblock Grounded language-image pre-training.
\newblock In \emph{Proceedings of the IEEE/CVF Conference on Computer Vision and Pattern Recognition}, pages 10965--10975, 2022{\natexlab{c}}.

\bibitem[Li et~al.(2024{\natexlab{b}})Li, Li, Yin, Ahmed, Liu, and Liu]{li2024red}
Mukai Li, Lei Li, Yuwei Yin, Masood Ahmed, Zhenguang Liu, and Qi~Liu.
\newblock Red teaming visual language models.
\newblock \emph{arXiv preprint arXiv:2401.12915}, 2024{\natexlab{b}}.

\bibitem[Li et~al.(2021)Li, Liang, Zhao, Cui, Ouyang, Shao, Yu, and Yan]{li2021supervision}
Yangguang Li, Feng Liang, Lichen Zhao, Yufeng Cui, Wanli Ouyang, Jing Shao, Fengwei Yu, and Junjie Yan.
\newblock Supervision exists everywhere: A data efficient contrastive language-image pre-training paradigm.
\newblock \emph{arXiv preprint arXiv:2110.05208}, 2021.

\bibitem[Li et~al.(2023{\natexlab{f}})Li, Fan, Hu, Feichtenhofer, and He]{li2023scaling}
Yanghao Li, Haoqi Fan, Ronghang Hu, Christoph Feichtenhofer, and Kaiming He.
\newblock Scaling language-image pre-training via masking.
\newblock In \emph{Proceedings of the IEEE/CVF Conference on Computer Vision and Pattern Recognition}, pages 23390--23400, 2023{\natexlab{f}}.

\bibitem[Li and Vasconcelos(2024)]{li2023debias}
Yi~Li and Nuno Vasconcelos.
\newblock Debias your {VLM} with counterfactuals: A unified approach, 2024.
\newblock URL \url{https://openreview.net/forum?id=xx05gm7oQw}.

\bibitem[Li et~al.(2023{\natexlab{g}})Li, Du, Zhou, Wang, Zhao, and Wen]{li2023evaluating}
Yifan Li, Yifan Du, Kun Zhou, Jinpeng Wang, Xin Zhao, and Ji-Rong Wen.
\newblock Evaluating object hallucination in large vision-language models.
\newblock In Houda Bouamor, Juan Pino, and Kalika Bali, editors, \emph{Proceedings of the 2023 Conference on Empirical Methods in Natural Language Processing}, pages 292--305, Singapore, December 2023{\natexlab{g}}. Association for Computational Linguistics.
\newblock \doi{10.18653/v1/2023.emnlp-main.20}.
\newblock URL \url{https://aclanthology.org/2023.emnlp-main.20}.

\bibitem[Li et~al.(2016)Li, Song, Cao, Tetreault, Goldberg, Jaimes, and Luo]{tgif-cvpr2016}
Yuncheng Li, Yale Song, Liangliang Cao, Joel Tetreault, Larry Goldberg, Alejandro Jaimes, and Jiebo Luo.
\newblock {TGIF: A New Dataset and Benchmark on Animated GIF Description}.
\newblock In \emph{The IEEE Conference on Computer Vision and Pattern Recognition (CVPR)}, June 2016.

\bibitem[Li et~al.(2023{\natexlab{h}})Li, Yang, Liu, Ma, Zhang, Yang, Sun, Liu, and Bai]{li2024monkey}
Zhang Li, Biao Yang, Qiang Liu, Zhiyin Ma, Shuo Zhang, Jingxu Yang, Yabo Sun, Yuliang Liu, and Xiang Bai.
\newblock Monkey: Image resolution and text label are important things for large multi-modal models.
\newblock \emph{arXiv preprint arXiv:2311.06607}, 2023{\natexlab{h}}.

\bibitem[Liang et~al.(2024)Liang, Zadeh, and Morency]{liang2023foundations}
Paul~Pu Liang, Amir Zadeh, and Louis-Philippe Morency.
\newblock Foundations \& trends in multimodal machine learning: Principles, challenges, and open questions.
\newblock \emph{ACM Comput. Surv.}, apr 2024.
\newblock ISSN 0360-0300.
\newblock \doi{10.1145/3656580}.
\newblock URL \url{https://doi.org/10.1145/3656580}.
\newblock Just Accepted.

\bibitem[Liang et~al.(2022)Liang, Bommasani, Lee, Tsipras, Soylu, Yasunaga, Zhang, Narayanan, Wu, Kumar, et~al.]{liang2022holistic}
Percy Liang, Rishi Bommasani, Tony Lee, Dimitris Tsipras, Dilara Soylu, Michihiro Yasunaga, Yian Zhang, Deepak Narayanan, Yuhuai Wu, Ananya Kumar, et~al.
\newblock Holistic evaluation of language models.
\newblock \emph{arXiv preprint arXiv:2211.09110}, 2022.

\bibitem[Lin et~al.(2023)Lin, Ye, Zhu, Cui, Ning, Jin, and Yuan]{lin2023videollava}
Bin Lin, Yang Ye, Bin Zhu, Jiaxi Cui, Munan Ning, Peng Jin, and Li~Yuan.
\newblock {Video-LLaVA}: Learning united visual representation by alignment before projection.
\newblock \emph{arXiv preprint arXiv:2311.10122}, 2023.

\bibitem[Lin(2004)]{lin-2004-rouge}
Chin-Yew Lin.
\newblock {ROUGE}: A package for automatic evaluation of summaries.
\newblock In \emph{Text Summarization Branches Out}, pages 74--81, Barcelona, Spain, July 2004. Association for Computational Linguistics.
\newblock URL \url{https://aclanthology.org/W04-1013}.

\bibitem[Lin et~al.(2014)Lin, Maire, Belongie, Hays, Perona, Ramanan, Doll{\'a}r, and Zitnick]{lin2014microsoft}
Tsung-Yi Lin, Michael Maire, Serge Belongie, James Hays, Pietro Perona, Deva Ramanan, Piotr Doll{\'a}r, and C~Lawrence Zitnick.
\newblock Microsoft {COCO}: Common objects in context.
\newblock In \emph{Computer Vision--ECCV 2014: 13th European Conference, Zurich, Switzerland, September 6-12, 2014, Proceedings, Part V 13}, pages 740--755. Springer, 2014.

\bibitem[Lin et~al.(2024{\natexlab{a}})Lin, Chen, Pathak, Zhang, and Ramanan]{lin2024revisiting}
Zhiqiu Lin, Xinyue Chen, Deepak Pathak, Pengchuan Zhang, and Deva Ramanan.
\newblock Revisiting the role of language priors in vision-language models.
\newblock \emph{arXiv preprint arXiv:2306.01879}, 2024{\natexlab{a}}.

\bibitem[Lin et~al.(2024{\natexlab{b}})Lin, Pathak, Li, Li, Xia, Neubig, Zhang, and Ramanan]{lin2024evaluating}
Zhiqiu Lin, Deepak Pathak, Baiqi Li, Jiayao Li, Xide Xia, Graham Neubig, Pengchuan Zhang, and Deva Ramanan.
\newblock Evaluating text-to-visual generation with image-to-text generation.
\newblock \emph{arXiv preprint arXiv:2404.01291}, 2024{\natexlab{b}}.

\bibitem[Liu et~al.(2023{\natexlab{a}})Liu, Emerson, and Collier]{liu2023visual}
Fangyu Liu, Guy Emerson, and Nigel Collier.
\newblock Visual spatial reasoning.
\newblock \emph{Transactions of the Association for Computational Linguistics}, 11:\penalty0 635--651, 2023{\natexlab{a}}.
\newblock \doi{10.1162/tacl_a_00566}.
\newblock URL \url{https://aclanthology.org/2023.tacl-1.37}.

\bibitem[Liu et~al.(2023{\natexlab{b}})Liu, Lin, Li, Wang, Yacoob, and Wang]{liu2023aligning}
Fuxiao Liu, Kevin Lin, Linjie Li, Jianfeng Wang, Yaser Yacoob, and Lijuan Wang.
\newblock Aligning large multi-modal model with robust instruction tuning.
\newblock \emph{arXiv preprint arXiv:2306.14565}, 2023{\natexlab{b}}.

\bibitem[Liu et~al.(2023{\natexlab{c}})Liu, Li, Li, and Lee]{liu2023improved}
Haotian Liu, Chunyuan Li, Yuheng Li, and Yong~Jae Lee.
\newblock Improved baselines with visual instruction tuning.
\newblock \emph{arXiv preprint arXiv: 2310.03744}, 2023{\natexlab{c}}.

\bibitem[Liu et~al.(2023{\natexlab{d}})Liu, Li, Wu, and Lee]{liu2023visual_tuning}
Haotian Liu, Chunyuan Li, Qingyang Wu, and Yong~Jae Lee.
\newblock Visual instruction tuning.
\newblock In A.~Oh, T.~Naumann, A.~Globerson, K.~Saenko, M.~Hardt, and S.~Levine, editors, \emph{Advances in Neural Information Processing Systems}, volume~36, pages 34892--34916. Curran Associates, Inc., 2023{\natexlab{d}}.
\newblock URL \url{https://proceedings.neurips.cc/paper_files/paper/2023/file/6dcf277ea32ce3288914faf369fe6de0-Paper-Conference.pdf}.

\bibitem[Liu et~al.(2024{\natexlab{a}})Liu, Li, Li, Li, Zhang, Shen, and Lee]{liu2024llavanext}
Haotian Liu, Chunyuan Li, Yuheng Li, Bo~Li, Yuanhan Zhang, Sheng Shen, and Yong~Jae Lee.
\newblock Llava-next: Improved reasoning, ocr, and world knowledge, January 2024{\natexlab{a}}.
\newblock URL \url{https://llava-vl.github.io/blog/2024-01-30-llava-next/}.

\bibitem[Liu et~al.(2024{\natexlab{b}})Liu, Wang, Yin, Molchanov, Wang, Cheng, and Chen]{liu2024dora}
Shih-Yang Liu, Chien-Yi Wang, Hongxu Yin, Pavlo Molchanov, Yu-Chiang~Frank Wang, Kwang-Ting Cheng, and Min-Hung Chen.
\newblock {DoRA}: Weight-decomposed low-rank adaptation.
\newblock \emph{arXiv preprint arXiv:2402.09353}, 2024{\natexlab{b}}.

\bibitem[Liu et~al.(2023{\natexlab{e}})Liu, Li, Li, Yu, Huang, Peng, Liu, Chen, Li, Jin, et~al.]{liu2024}
Yuliang Liu, Zhang Li, Hongliang Li, Wenwen Yu, Mingxin Huang, Dezhi Peng, Mingyu Liu, Mingrui Chen, Chunyuan Li, Lianwen Jin, et~al.
\newblock On the hidden mystery of {OCR} in large multimodal models.
\newblock \emph{arXiv preprint arXiv:2305.07895}, 2023{\natexlab{e}}.

\bibitem[Lu et~al.(2019)Lu, Batra, Parikh, and Lee]{vilbert}
Jiasen Lu, Dhruv Batra, Devi Parikh, and Stefan Lee.
\newblock Vilbert: Pretraining task-agnostic visiolinguistic representations for vision-and-language tasks.
\newblock In H.~Wallach, H.~Larochelle, A.~Beygelzimer, F.~d\textquotesingle Alch\'{e}-Buc, E.~Fox, and R.~Garnett, editors, \emph{Advances in Neural Information Processing Systems}, volume~32. Curran Associates, Inc., 2019.
\newblock URL \url{https://proceedings.neurips.cc/paper_files/paper/2019/file/c74d97b01eae257e44aa9d5bade97baf-Paper.pdf}.

\bibitem[Lu et~al.(2022)Lu, Mishra, Xia, Qiu, Chang, Zhu, Tafjord, Clark, and Kalyan]{lu2022learn}
Pan Lu, Swaroop Mishra, Tanglin Xia, Liang Qiu, Kai-Wei Chang, Song-Chun Zhu, Oyvind Tafjord, Peter Clark, and Ashwin Kalyan.
\newblock Learn to explain: Multimodal reasoning via thought chains for science question answering.
\newblock \emph{Advances in Neural Information Processing Systems}, 35:\penalty0 2507--2521, 2022.

\bibitem[Ma et~al.(2023)Ma, Hong, Gul, Gandhi, Gao, and Krishna]{crepe}
Zixian Ma, Jerry Hong, Mustafa~Omer Gul, Mona Gandhi, Irena Gao, and Ranjay Krishna.
\newblock {CREPE}: Can vision-language foundation models reason compositionally?
\newblock In \emph{Proceedings of the IEEE/CVF Conference on Computer Vision and Pattern Recognition}, pages 10910--10921, 2023.

\bibitem[Mahmoud et~al.(2024)Mahmoud, Elhoushi, Abbas, Yang, Ardalani, Leather, and Morcos]{mahmoud2023sieve}
Anas Mahmoud, Mostafa Elhoushi, Amro Abbas, Yu~Yang, Newsha Ardalani, Hugh Leather, and Ari Morcos.
\newblock Sieve: Multimodal dataset pruning using image captioning models.
\newblock In \emph{Proceedings of the IEEE/CVF Conference on Computer Vision and Pattern Recognition (CVPR)}, June 2024.

\bibitem[Maini et~al.(2023)Maini, Goyal, Lipton, Kolter, and Raghunathan]{tmars}
Pratyush Maini, Sachin Goyal, Zachary~C Lipton, J~Zico Kolter, and Aditi Raghunathan.
\newblock {T-MARS}: Improving visual representations by circumventing text feature learning.
\newblock \emph{arXiv preprint arXiv:2307.03132}, 2023.

\bibitem[Maji et~al.(2013)Maji, Rahtu, Kannala, Blaschko, and Vedaldi]{aircraft}
Subhransu Maji, Esa Rahtu, Juho Kannala, Matthew Blaschko, and Andrea Vedaldi.
\newblock Fine-grained visual classification of aircraft.
\newblock \emph{arXiv preprint arXiv:1306.5151}, 2013.

\bibitem[Ma{\~n}as et~al.(2023)Ma{\~n}as, Lopez, Ahmadi, Nematzadeh, Goyal, and Agrawal]{manas2023mapl}
Oscar Ma{\~n}as, Pau~Rodriguez Lopez, Saba Ahmadi, Aida Nematzadeh, Yash Goyal, and Aishwarya Agrawal.
\newblock {MAPL}: Parameter-efficient adaptation of unimodal pre-trained models for vision-language few-shot prompting.
\newblock In \emph{Proceedings of the 17th Conference of the European Chapter of the Association for Computational Linguistics}, pages 2523--2548, 2023.

\bibitem[Ma{\~n}as et~al.(2024)Ma{\~n}as, Krojer, and Agrawal]{manas2023improving}
Oscar Ma{\~n}as, Benno Krojer, and Aishwarya Agrawal.
\newblock Improving automatic {VQA} evaluation using large language models.
\newblock In \emph{Proceedings of the AAAI Conference on Artificial Intelligence}, volume~38, pages 4171--4179, 2024.

\bibitem[Mangalam et~al.(2024)Mangalam, Akshulakov, and Malik]{mangalam2024egoschema}
Karttikeya Mangalam, Raiymbek Akshulakov, and Jitendra Malik.
\newblock Egoschema: A diagnostic benchmark for very long-form video language understanding.
\newblock \emph{Advances in Neural Information Processing Systems}, 36, 2024.

\bibitem[Marino et~al.(2019)Marino, Rastegari, Farhadi, and Mottaghi]{marino2019ok}
Kenneth Marino, Mohammad Rastegari, Ali Farhadi, and Roozbeh Mottaghi.
\newblock {OK-VQA}: A visual question answering benchmark requiring external knowledge.
\newblock In \emph{Proceedings of the IEEE/CVF Conference on Computer Vision and Pattern Recognition}, pages 3195--3204, 2019.

\bibitem[Masry et~al.(2022)Masry, Do, Tan, Joty, and Hoque]{masry2022chartqa}
Ahmed Masry, Xuan~Long Do, Jia~Qing Tan, Shafiq Joty, and Enamul Hoque.
\newblock {C}hart{QA}: A benchmark for question answering about charts with visual and logical reasoning.
\newblock In Smaranda Muresan, Preslav Nakov, and Aline Villavicencio, editors, \emph{Findings of the Association for Computational Linguistics: ACL 2022}, pages 2263--2279, Dublin, Ireland, May 2022. Association for Computational Linguistics.
\newblock \doi{10.18653/v1/2022.findings-acl.177}.
\newblock URL \url{https://aclanthology.org/2022.findings-acl.177}.

\bibitem[Mathew et~al.(2021)Mathew, Karatzas, and Jawahar]{mathew2021docvqa}
Minesh Mathew, Dimosthenis Karatzas, and C.~V. Jawahar.
\newblock {DocVQA}: A dataset for vqa on document images.
\newblock In \emph{2021 IEEE Winter Conference on Applications of Computer Vision (WACV)}, pages 2199--2208, 2021.
\newblock \doi{10.1109/WACV48630.2021.00225}.

\bibitem[Mathew et~al.(2022)Mathew, Bagal, Tito, Karatzas, Valveny, and Jawahar]{mathew2021infographicvqa}
Minesh Mathew, Viraj Bagal, Rub{\`e}n Tito, Dimosthenis Karatzas, Ernest Valveny, and CV~Jawahar.
\newblock Infographic{VQA}.
\newblock In \emph{Proceedings of the IEEE/CVF Winter Conference on Applications of Computer Vision}, pages 1697--1706, 2022.

\bibitem[Mathur et~al.(2020)Mathur, Baldwin, and Cohn]{mathur-etal-2020-tangled}
Nitika Mathur, Timothy Baldwin, and Trevor Cohn.
\newblock Tangled up in {BLEU}: Reevaluating the evaluation of automatic machine translation evaluation metrics.
\newblock In Dan Jurafsky, Joyce Chai, Natalie Schluter, and Joel Tetreault, editors, \emph{Proceedings of the 58th Annual Meeting of the Association for Computational Linguistics}, pages 4984--4997, Online, July 2020. Association for Computational Linguistics.
\newblock \doi{10.18653/v1/2020.acl-main.448}.
\newblock URL \url{https://aclanthology.org/2020.acl-main.448}.

\bibitem[McKinzie et~al.(2024)McKinzie, Gan, Fauconnier, Dodge, Zhang, Dufter, Shah, Du, Peng, Weers, et~al.]{mckinzie2024mm1}
Brandon McKinzie, Zhe Gan, Jean-Philippe Fauconnier, Sam Dodge, Bowen Zhang, Philipp Dufter, Dhruti Shah, Xianzhi Du, Futang Peng, Floris Weers, et~al.
\newblock {MM1}: Methods, analysis \& insights from multimodal llm pre-training.
\newblock \emph{arXiv preprint arXiv:2403.09611}, 2024.

\bibitem[Menon and Vondrick(2023)]{menon2022visual}
Sachit Menon and Carl Vondrick.
\newblock Visual classification via description from large language models.
\newblock In \emph{The Eleventh International Conference on Learning Representations}, 2023.
\newblock URL \url{https://openreview.net/forum?id=jlAjNL8z5cs}.

\bibitem[Merullo et~al.(2022)Merullo, Castricato, Eickhoff, and Pavlick]{merullo2022linearly}
Jack Merullo, Louis Castricato, Carsten Eickhoff, and Ellie Pavlick.
\newblock Linearly mapping from image to text space.
\newblock In \emph{The Eleventh International Conference on Learning Representations}, 2022.

\bibitem[Mishra et~al.(2012)Mishra, Alahari, and Jawahar]{mishra2012}
Anand Mishra, Karteek Alahari, and C.~V. Jawahar.
\newblock Top-down and bottom-up cues for scene text recognition.
\newblock In \emph{2012 IEEE Conference on Computer Vision and Pattern Recognition}, pages 2687--2694, 2012.
\newblock \doi{10.1109/CVPR.2012.6247990}.

\bibitem[Mishra et~al.(2019)Mishra, Shekhar, Singh, and Chakraborty]{mishraICDAR19}
Anand Mishra, Shashank Shekhar, Ajeet~Kumar Singh, and Anirban Chakraborty.
\newblock {OCR-VQA}: Visual question answering by reading text in images.
\newblock In \emph{2019 International Conference on Document Analysis and Recognition (ICDAR)}, pages 947--952. IEEE, 2019.

\bibitem[Momeni et~al.(2023)Momeni, Caron, Nagrani, Zisserman, and Schmid]{momeni2023verbs}
Liliane Momeni, Mathilde Caron, Arsha Nagrani, Andrew Zisserman, and Cordelia Schmid.
\newblock Verbs in action: Improving verb understanding in video-language models.
\newblock In \emph{Proceedings of the IEEE/CVF International Conference on Computer Vision}, pages 15579--15591, 2023.

\bibitem[Mu et~al.(2022)Mu, Kirillov, Wagner, and Xie]{mu2022slip}
Norman Mu, Alexander Kirillov, David Wagner, and Saining Xie.
\newblock {SLIP}: Self-supervision meets language-image pre-training.
\newblock In \emph{European Conference on Computer Vision}, pages 529--544. Springer, 2022.

\bibitem[Ng and Jordan(2001)]{ng2001discriminative}
Andrew Ng and Michael Jordan.
\newblock On discriminative vs. generative classifiers: A comparison of logistic regression and naive bayes.
\newblock \emph{Advances in Neural Information Processing Systems}, 14, 2001.

\bibitem[Nguyen et~al.(2023)Nguyen, Gadre, Ilharco, Oh, and Schmidt]{nguyen2023improving}
Thao Nguyen, Samir~Yitzhak Gadre, Gabriel Ilharco, Sewoong Oh, and Ludwig Schmidt.
\newblock Improving multimodal datasets with image captioning.
\newblock In \emph{Thirty-seventh Conference on Neural Information Processing Systems Datasets and Benchmarks Track}, 2023.
\newblock URL \url{https://openreview.net/forum?id=VIRKdeFJIg}.

\bibitem[Nilsback and Zisserman(2008)]{Flower102}
Maria-Elena Nilsback and Andrew Zisserman.
\newblock Automated flower classification over a large number of classes.
\newblock \emph{2008 Sixth Indian Conference on Computer Vision, Graphics \& Image Processing}, pages 722--729, 2008.
\newblock URL \url{https://api.semanticscholar.org/CorpusID:15193013}.

\bibitem[Oord et~al.(2018)Oord, Li, and Vinyals]{oord2018representation}
Aaron van~den Oord, Yazhe Li, and Oriol Vinyals.
\newblock Representation learning with contrastive predictive coding.
\newblock \emph{arXiv preprint arXiv:1807.03748}, 2018.

\bibitem[Ordonez et~al.(2011)Ordonez, Kulkarni, and Berg]{sbu}
Vicente Ordonez, Girish Kulkarni, and Tamara Berg.
\newblock Im2text: Describing images using 1 million captioned photographs.
\newblock In J.~Shawe-Taylor, R.~Zemel, P.~Bartlett, F.~Pereira, and K.Q. Weinberger, editors, \emph{Advances in Neural Information Processing Systems}, volume~24. Curran Associates, Inc., 2011.
\newblock URL \url{https://proceedings.neurips.cc/paper_files/paper/2011/file/5dd9db5e033da9c6fb5ba83c7a7ebea9-Paper.pdf}.

\bibitem[Ouyang et~al.(2022)Ouyang, Wu, Jiang, Almeida, Wainwright, Mishkin, Zhang, Agarwal, Slama, Ray, et~al.]{ouyang2022training}
Long Ouyang, Jeffrey Wu, Xu~Jiang, Diogo Almeida, Carroll Wainwright, Pamela Mishkin, Chong Zhang, Sandhini Agarwal, Katarina Slama, Alex Ray, et~al.
\newblock Training language models to follow instructions with human feedback.
\newblock \emph{Advances in Neural Information Processing Systems}, 35:\penalty0 27730--27744, 2022.

\bibitem[Papineni et~al.(2002)Papineni, Roukos, Ward, and Zhu]{papineni2002bleu}
Kishore Papineni, Salim Roukos, Todd Ward, and Wei-Jing Zhu.
\newblock Bleu: a method for automatic evaluation of machine translation.
\newblock In \emph{Proceedings of the 40th annual meeting on association for computational linguistics}, pages 311--318. Association for Computational Linguistics, 2002.

\bibitem[Parashar et~al.(2023)Parashar, Lin, Li, and Kong]{parashar2023prompting}
Shubham Parashar, Zhiqiu Lin, Yanan Li, and Shu Kong.
\newblock Prompting scientific names for zero-shot species recognition.
\newblock \emph{arXiv preprint arXiv:2310.09929}, 2023.

\bibitem[Parashar et~al.(2024)Parashar, Lin, Liu, Dong, Li, Ramanan, Caverlee, and Kong]{parashar2024neglected}
Shubham Parashar, Zhiqiu Lin, Tian Liu, Xiangjue Dong, Yanan Li, Deva Ramanan, James Caverlee, and Shu Kong.
\newblock The neglected tails of vision-language models.
\newblock \emph{arXiv preprint arXiv:2401.12425}, 2024.

\bibitem[Parkhi et~al.(2012)Parkhi, Vedaldi, Zisserman, and Jawahar]{OxfordPet}
Omkar~M. Parkhi, Andrea Vedaldi, Andrew Zisserman, and C.~V. Jawahar.
\newblock Cats and dogs.
\newblock In \emph{IEEE Conference on Computer Vision and Pattern Recognition}, 2012.

\bibitem[Pathak et~al.(2016)Pathak, Krähenbühl, Donahue, Darrell, and Efros]{inpainting}
Deepak Pathak, Philipp Krähenbühl, Jeff Donahue, Trevor Darrell, and Alexei~A. Efros.
\newblock Context encoders: Feature learning by inpainting.
\newblock In \emph{2016 IEEE Conference on Computer Vision and Pattern Recognition (CVPR)}, pages 2536--2544, 2016.
\newblock \doi{10.1109/CVPR.2016.278}.

\bibitem[Peng et~al.(2024)Peng, Wang, Dong, Hao, Huang, Ma, Ye, and Wei]{peng2024grounding}
Zhiliang Peng, Wenhui Wang, Li~Dong, Yaru Hao, Shaohan Huang, Shuming Ma, Qixiang Ye, and Furu Wei.
\newblock Grounding multimodal large language models to the world.
\newblock In \emph{The Twelfth International Conference on Learning Representations}, 2024.
\newblock URL \url{https://openreview.net/forum?id=lLmqxkfSIw}.

\bibitem[Perez et~al.(2022)Perez, Huang, Song, Cai, Ring, Aslanides, Glaese, McAleese, and Irving]{perez2022red}
Ethan Perez, Saffron Huang, Francis Song, Trevor Cai, Roman Ring, John Aslanides, Amelia Glaese, Nat McAleese, and Geoffrey Irving.
\newblock Red teaming language models with language models.
\newblock In Yoav Goldberg, Zornitsa Kozareva, and Yue Zhang, editors, \emph{Proceedings of the 2022 Conference on Empirical Methods in Natural Language Processing}, pages 3419--3448, Abu Dhabi, United Arab Emirates, December 2022. Association for Computational Linguistics.
\newblock \doi{10.18653/v1/2022.emnlp-main.225}.
\newblock URL \url{https://aclanthology.org/2022.emnlp-main.225}.

\bibitem[Prabhudesai et~al.(2023)Prabhudesai, Ke, Li, Pathak, and Fragkiadaki]{prabhudesai2023test}
Mihir Prabhudesai, Tsung-Wei Ke, Alexander~C Li, Deepak Pathak, and Katerina Fragkiadaki.
\newblock Test-time adaptation of discriminative models via diffusion generative feedback.
\newblock \emph{arXiv preprint arXiv:2311.16102}, 2023.

\bibitem[Pratt et~al.(2023)Pratt, Covert, Liu, and Farhadi]{pratt2023does}
Sarah Pratt, Ian Covert, Rosanne Liu, and Ali Farhadi.
\newblock What does a platypus look like? generating customized prompts for zero-shot image classification.
\newblock In \emph{Proceedings of the IEEE/CVF International Conference on Computer Vision}, pages 15691--15701, 2023.

\bibitem[Radenovic et~al.(2023{\natexlab{a}})Radenovic, Dubey, Kadian, Mihaylov, Vandenhende, Patel, Wen, Ramanathan, and Mahajan]{DiHT}
Filip Radenovic, Abhimanyu Dubey, Abhishek Kadian, Todor Mihaylov, Simon Vandenhende, Yash Patel, Yi~Wen, Vignesh Ramanathan, and Dhruv Mahajan.
\newblock Filtering, distillation, and hard negatives for vision-language pre-training.
\newblock In \emph{Proceedings of the IEEE/CVF Conference on Computer Vision and Pattern Recognition}, pages 6967--6977, 2023{\natexlab{a}}.

\bibitem[Radenovic et~al.(2023{\natexlab{b}})Radenovic, Dubey, Kadian, Mihaylov, Vandenhende, Patel, Wen, Ramanathan, and Mahajan]{radenovic2023filtering}
Filip Radenovic, Abhimanyu Dubey, Abhishek Kadian, Todor Mihaylov, Simon Vandenhende, Yash Patel, Yi~Wen, Vignesh Ramanathan, and Dhruv Mahajan.
\newblock Filtering, distillation, and hard negatives for vision-language pre-training.
\newblock In \emph{Proceedings of the IEEE/CVF Conference on Computer Vision and Pattern Recognition}, pages 6967--6977, 2023{\natexlab{b}}.

\bibitem[Radford et~al.(2021)Radford, Kim, Hallacy, Ramesh, Goh, Agarwal, Sastry, Askell, Mishkin, Clark, Krueger, and Sutskever]{clip_openai_radford21a}
Alec Radford, Jong~Wook Kim, Chris Hallacy, Aditya Ramesh, Gabriel Goh, Sandhini Agarwal, Girish Sastry, Amanda Askell, Pamela Mishkin, Jack Clark, Gretchen Krueger, and Ilya Sutskever.
\newblock Learning transferable visual models from natural language supervision.
\newblock In Marina Meila and Tong Zhang, editors, \emph{Proceedings of the 38th International Conference on Machine Learning}, volume 139 of \emph{Proceedings of Machine Learning Research}, pages 8748--8763. PMLR, 18--24 Jul 2021.
\newblock URL \url{https://proceedings.mlr.press/v139/radford21a.html}.

\bibitem[Raffel et~al.(2020)Raffel, Shazeer, Roberts, Lee, Narang, Matena, Zhou, Li, and Liu]{raffel2020exploring}
Colin Raffel, Noam Shazeer, Adam Roberts, Katherine Lee, Sharan Narang, Michael Matena, Yanqi Zhou, Wei Li, and Peter~J Liu.
\newblock Exploring the limits of transfer learning with a unified text-to-text transformer.
\newblock \emph{Journal of machine learning research}, 21\penalty0 (140):\penalty0 1--67, 2020.

\bibitem[Reid et~al.(2024)Reid, Savinov, Teplyashin, Lepikhin, Lillicrap, Alayrac, Soricut, Lazaridou, Firat, Schrittwieser, et~al.]{geminiteam2024gemini}
Machel Reid, Nikolay Savinov, Denis Teplyashin, Dmitry Lepikhin, Timothy Lillicrap, Jean-baptiste Alayrac, Radu Soricut, Angeliki Lazaridou, Orhan Firat, Julian Schrittwieser, et~al.
\newblock Gemini 1.5: Unlocking multimodal understanding across millions of tokens of context.
\newblock \emph{arXiv preprint arXiv:2403.05530}, 2024.

\bibitem[Rohrbach et~al.(2018)Rohrbach, Hendricks, Burns, Darrell, and Saenko]{objectHallucination}
Anna Rohrbach, Lisa~Anne Hendricks, Kaylee Burns, Trevor Darrell, and Kate Saenko.
\newblock Object hallucination in image captioning.
\newblock In Ellen Riloff, David Chiang, Julia Hockenmaier, and Jun{'}ichi Tsujii, editors, \emph{Proceedings of the 2018 Conference on Empirical Methods in Natural Language Processing}, pages 4035--4045, Brussels, Belgium, October-November 2018. Association for Computational Linguistics.
\newblock \doi{10.18653/v1/D18-1437}.
\newblock URL \url{https://aclanthology.org/D18-1437}.

\bibitem[Rombach et~al.(2021)Rombach, Blattmann, Lorenz, Esser, and Ommer]{rombach2021highresolution}
Robin Rombach, Andreas Blattmann, Dominik Lorenz, Patrick Esser, and Björn Ommer.
\newblock High-resolution image synthesis with latent diffusion models, 2021.

\bibitem[Rombach et~al.(2022)Rombach, Blattmann, Lorenz, Esser, and Ommer]{rombach2022high}
Robin Rombach, Andreas Blattmann, Dominik Lorenz, Patrick Esser, and Bj{\"o}rn Ommer.
\newblock High-resolution image synthesis with latent diffusion models.
\newblock In \emph{Proceedings of the IEEE/CVF Conference on Computer Vision and Pattern Recognition}, pages 10684--10695, 2022.

\bibitem[Ross et~al.(2020)Ross, Katz, and Barbu]{ross2020measuring}
Candace Ross, Boris Katz, and Andrei Barbu.
\newblock Measuring social biases in grounded vision and language embeddings.
\newblock \emph{arXiv preprint arXiv:2002.08911}, 2020.

\bibitem[Rubinstein et~al.(1997)Rubinstein, Hastie, et~al.]{rubinstein1997discriminative}
Y~Dan Rubinstein, Trevor Hastie, et~al.
\newblock Discriminative vs informative learning.
\newblock In \emph{KDD}, volume~5, pages 49--53, 1997.

\bibitem[Röttger et~al.(2024)Röttger, Pernisi, Vidgen, and Hovy]{rottger2024safetyprompts}
Paul Röttger, Fabio Pernisi, Bertie Vidgen, and Dirk Hovy.
\newblock Safetyprompts: a systematic review of open datasets for evaluating and improving large language model safety, 2024.

\bibitem[Sachdeva et~al.(2024)Sachdeva, Coleman, Kang, Ni, Hong, Chi, Caverlee, McAuley, and Cheng]{sachdeva2024train}
Noveen Sachdeva, Benjamin Coleman, Wang-Cheng Kang, Jianmo Ni, Lichan Hong, Ed~H Chi, James Caverlee, Julian McAuley, and Derek~Zhiyuan Cheng.
\newblock How to train data-efficient {LLMs}.
\newblock \emph{arXiv preprint arXiv:2402.09668}, 2024.

\bibitem[Saharia et~al.(2022)Saharia, Chan, Saxena, Li, Whang, Denton, Ghasemipour, Gontijo~Lopes, Karagol~Ayan, Salimans, et~al.]{saharia2022photorealistic}
Chitwan Saharia, William Chan, Saurabh Saxena, Lala Li, Jay Whang, Emily~L Denton, Kamyar Ghasemipour, Raphael Gontijo~Lopes, Burcu Karagol~Ayan, Tim Salimans, et~al.
\newblock Photorealistic text-to-image diffusion models with deep language understanding.
\newblock \emph{Advances in Neural Information Processing Systems}, 35:\penalty0 36479--36494, 2022.

\bibitem[Samadh et~al.(2023)Samadh, Gani, Hussein, Khattak, Naseer, Khan, and Khan]{samadh2023align}
Jameel Hassan~Abdul Samadh, Hanan Gani, Noor~Hazim Hussein, Muhammad~Uzair Khattak, Muzammal Naseer, Fahad Khan, and Salman Khan.
\newblock Align your prompts: Test-time prompting with distribution alignment for zero-shot generalization.
\newblock In \emph{Thirty-seventh Conference on Neural Information Processing Systems}, 2023.
\newblock URL \url{https://openreview.net/forum?id=CusNOTRkQw}.

\bibitem[Santurkar et~al.(2022)Santurkar, Dubois, Taori, Liang, and Hashimoto]{santurkar2022caption}
Shibani Santurkar, Yann Dubois, Rohan Taori, Percy Liang, and Tatsunori Hashimoto.
\newblock Is a caption worth a thousand images? {A} controlled study for representation learning.
\newblock \emph{arXiv preprint arXiv:2207.07635}, 2022.

\bibitem[Scheuerman et~al.(2023)Scheuerman, Weathington, Mugunthan, Denton, and Fiesler]{Scheuerman2023-data-tracing}
Morgan~Klaus Scheuerman, Katy Weathington, Tarun Mugunthan, Emily Denton, and Casey Fiesler.
\newblock From human to data to dataset: Mapping the traceability of human subjects in computer vision datasets.
\newblock \emph{Proceedings of the ACM on Human-Computer Interaction}, 7\penalty0 (CSCW1):\penalty0 1--33, 2023.

\bibitem[Schuhmann(2023)]{laion-aesthetics}
Christoph Schuhmann.
\newblock Laion-aesthetics.
\newblock \url{https://laion.ai/blog/laion-aesthetics/}, 2023.

\bibitem[Schuhmann et~al.(2021)Schuhmann, Vencu, Beaumont, Kaczmarczyk, Mullis, Katta, Coombes, Jitsev, and Komatsuzaki]{schuhmann2021laion}
Christoph Schuhmann, Richard Vencu, Romain Beaumont, Robert Kaczmarczyk, Clayton Mullis, Aarush Katta, Theo Coombes, Jenia Jitsev, and Aran Komatsuzaki.
\newblock Laion-400m: Open dataset of {CLIP}-filtered 400 million image-text pairs.
\newblock \emph{arXiv preprint arXiv:2111.02114}, 2021.

\bibitem[Schuhmann et~al.(2022)Schuhmann, Beaumont, Vencu, Gordon, Wightman, Cherti, Coombes, Katta, Mullis, Wortsman, Schramowski, Kundurthy, Crowson, Schmidt, Kaczmarczyk, and Jitsev]{schuhmann2022laion5b}
Christoph Schuhmann, Romain Beaumont, Richard Vencu, Cade Gordon, Ross Wightman, Mehdi Cherti, Theo Coombes, Aarush Katta, Clayton Mullis, Mitchell Wortsman, Patrick Schramowski, Srivatsa Kundurthy, Katherine Crowson, Ludwig Schmidt, Robert Kaczmarczyk, and Jenia Jitsev.
\newblock Laion-5b: An open large-scale dataset for training next generation image-text models, 2022.

\bibitem[Schwenk et~al.(2022)Schwenk, Khandelwal, Clark, Marino, and Mottaghi]{schwenk2022okvqa}
Dustin Schwenk, Apoorv Khandelwal, Christopher Clark, Kenneth Marino, and Roozbeh Mottaghi.
\newblock A-okvqa: A benchmark for visual question answering using world knowledge.
\newblock In \emph{European Conference on Computer Vision}, pages 146--162. Springer, 2022.

\bibitem[Sharma et~al.(2018{\natexlab{a}})Sharma, Ding, Goodman, and Soricut]{conceptualcaptions}
Piyush Sharma, Nan Ding, Sebastian Goodman, and Radu Soricut.
\newblock Conceptual captions: A cleaned, hypernymed, image alt-text dataset for automatic image captioning.
\newblock In \emph{Proceedings of the 56th Annual Meeting of the Association for Computational Linguistics (Volume 1: Long Papers)}, pages 2556--2565, Melbourne, Australia, July 2018{\natexlab{a}}. Association for Computational Linguistics.
\newblock \doi{10.18653/v1/P18-1238}.
\newblock URL \url{https://aclanthology.org/P18-1238}.

\bibitem[Sharma et~al.(2018{\natexlab{b}})Sharma, Ding, Goodman, and Soricut]{sharma2018conceptual}
Piyush Sharma, Nan Ding, Sebastian Goodman, and Radu Soricut.
\newblock Conceptual captions: A cleaned, hypernymed, image alt-text dataset for automatic image captioning.
\newblock In \emph{Proceedings of the 56th Annual Meeting of the Association for Computational Linguistics (Volume 1: Long Papers)}, pages 2556--2565, 2018{\natexlab{b}}.

\bibitem[Sharma et~al.(2024)Sharma, Padthe, Ardalani, Tirumala, Howes, Xu, Huang, Li, Aghajanyan, and Ghosh]{sharma2024text}
Vasu Sharma, Karthik Padthe, Newsha Ardalani, Kushal Tirumala, Russell Howes, Hu~Xu, Po-Yao Huang, Shang-Wen Li, Armen Aghajanyan, and Gargi Ghosh.
\newblock Text quality-based pruning for efficient training of language models.
\newblock \emph{arXiv preprint arXiv:2405.01582}, 2024.

\bibitem[Shenoy et~al.(2024)Shenoy, Lu, Jayakumar, Chatterjee, Moslehpour, Chuang, Harpale, Bhardwaj, Xu, Zhao, et~al.]{shenoy2024lumos}
Ashish Shenoy, Yichao Lu, Srihari Jayakumar, Debojeet Chatterjee, Mohsen Moslehpour, Pierce Chuang, Abhay Harpale, Vikas Bhardwaj, Di~Xu, Shicong Zhao, et~al.
\newblock Lumos: Empowering multimodal llms with scene text recognition.
\newblock \emph{arXiv preprint arXiv:2402.08017}, 2024.

\bibitem[Shi et~al.(2014)Shi, Wang, Xiao, Gao, and Hu]{svt}
Cunzhao Shi, Chunheng Wang, Baihua Xiao, Song Gao, and Jinlong Hu.
\newblock End-to-end scene text recognition using tree-structured models.
\newblock \emph{Pattern Recognition}, 47:\penalty0 2853--2866, 2014.

\bibitem[Shwartz~Ziv and LeCun(2024)]{shwartz2024compress}
Ravid Shwartz~Ziv and Yann LeCun.
\newblock To compress or not to compress—self-supervised learning and information theory: A review.
\newblock \emph{Entropy}, 26\penalty0 (3):\penalty0 252, 2024.

\bibitem[Si et~al.(2023)Si, Shi, Zhao, Zettlemoyer, and Boyd-Graber]{Si:Shi:Zhao:Zettlemoyer:Boyd-Graber-2023}
Chenglei Si, Weijia Shi, Chen Zhao, Luke Zettlemoyer, and Jordan~Lee Boyd-Graber.
\newblock Getting \underline{MoRE} out of \underline{M}ixture \underline{o}f language model \underline{R}easoning \underline{E}xperts.
\newblock \emph{Findings of Empirical Methods in Natural Language Processing}, 2023.

\bibitem[Singh et~al.(2019)Singh, Natarajan, Shah, Jiang, Chen, Batra, Parikh, and Rohrbach]{singh2019towards}
Amanpreet Singh, Vivek Natarajan, Meet Shah, Yu~Jiang, Xinlei Chen, Dhruv Batra, Devi Parikh, and Marcus Rohrbach.
\newblock Towards {VQA} models that can read.
\newblock In \emph{Proceedings of the IEEE/CVF Conference on Computer Vision and Pattern Recognition}, pages 8317--8326, 2019.

\bibitem[Singh et~al.(2022)Singh, Hu, Goswami, Couairon, Galuba, Rohrbach, and Kiela]{singh2022flava}
Amanpreet Singh, Ronghang Hu, Vedanuj Goswami, Guillaume Couairon, Wojciech Galuba, Marcus Rohrbach, and Douwe Kiela.
\newblock Flava: A foundational language and vision alignment model.
\newblock In \emph{Proceedings of the IEEE/CVF Conference on Computer Vision and Pattern Recognition}, pages 15638--15650, 2022.

\bibitem[Smith et~al.(2023)Smith, Farinha, Hall, Kirk, Shtedritski, and Bain]{smith2023balancing}
Brandon Smith, Miguel Farinha, Siobhan~Mackenzie Hall, Hannah~Rose Kirk, Aleksandar Shtedritski, and Max Bain.
\newblock Balancing the picture: Debiasing vision-language datasets with synthetic contrast sets.
\newblock \emph{arXiv preprint arXiv:2305.15407}, 2023.

\bibitem[Somepalli et~al.(2023)Somepalli, Singla, Goldblum, Geiping, and Goldstein]{somepalli2023diffusion}
Gowthami Somepalli, Vasu Singla, Micah Goldblum, Jonas Geiping, and Tom Goldstein.
\newblock Diffusion art or digital forgery? {I}nvestigating data replication in diffusion models.
\newblock In \emph{Proceedings of the IEEE/CVF Conference on Computer Vision and Pattern Recognition}, pages 6048--6058, 2023.

\bibitem[Sorscher et~al.(2022)Sorscher, Geirhos, Shekhar, Ganguli, and Morcos]{sorscher2022beyond}
Ben Sorscher, Robert Geirhos, Shashank Shekhar, Surya Ganguli, and Ari~S. Morcos.
\newblock Beyond neural scaling laws: beating power law scaling via data pruning.
\newblock In Alice~H. Oh, Alekh Agarwal, Danielle Belgrave, and Kyunghyun Cho, editors, \emph{Advances in Neural Information Processing Systems}, 2022.
\newblock URL \url{https://openreview.net/forum?id=UmvSlP-PyV}.

\bibitem[Srinivasan and Bisk(2021)]{srinivasan2021worst}
Tejas Srinivasan and Yonatan Bisk.
\newblock Worst of both worlds: Biases compound in pre-trained vision-and-language models.
\newblock \emph{arXiv preprint arXiv:2104.08666}, 2021.

\bibitem[Su et~al.(2023)Su, Lan, Li, Xu, Wang, and Cai]{su2023pandagpt}
Yixuan Su, Tian Lan, Huayang Li, Jialu Xu, Yan Wang, and Deng Cai.
\newblock Pandagpt: One model to instruction-follow them all.
\newblock \emph{arXiv preprint arXiv:2305.16355}, 2023.

\bibitem[Sun et~al.(2019)Sun, Myers, Vondrick, Murphy, and Schmid]{sun2019videobert}
Chen Sun, Austin Myers, Carl Vondrick, Kevin Murphy, and Cordelia Schmid.
\newblock Videobert: A joint model for video and language representation learning.
\newblock In \emph{ICCV}, 2019.

\bibitem[Sun et~al.(2024)Sun, Huang, Wang, Wu, Zhang, Gao, Huang, Lyu, Zhang, Li, et~al.]{sun2024trustllm}
Lichao Sun, Yue Huang, Haoran Wang, Siyuan Wu, Qihui Zhang, Chujie Gao, Yixin Huang, Wenhan Lyu, Yixuan Zhang, Xiner Li, et~al.
\newblock Trust{LLM}: Trustworthiness in large language models.
\newblock \emph{arXiv preprint arXiv:2401.05561}, 2024.

\bibitem[Sun et~al.(2023)Sun, Shen, Cao, Liu, Li, Shen, Gan, Gui, Wang, Yang, et~al.]{sun2023aligning}
Zhiqing Sun, Sheng Shen, Shengcao Cao, Haotian Liu, Chunyuan Li, Yikang Shen, Chuang Gan, Liang-Yan Gui, Yu-Xiong Wang, Yiming Yang, et~al.
\newblock Aligning large multimodal models with factually augmented rlhf.
\newblock \emph{arXiv preprint arXiv:2309.14525}, 2023.

\bibitem[Sung et~al.(2022)Sung, Cho, and Bansal]{sung2022vl}
Yi-Lin Sung, Jaemin Cho, and Mohit Bansal.
\newblock Vl-adapter: Parameter-efficient transfer learning for vision-and-language tasks.
\newblock In \emph{Proceedings of the IEEE/CVF Conference on Computer Vision and Pattern Recognition}, pages 5227--5237, 2022.

\bibitem[Tapaswi et~al.(2016)Tapaswi, Zhu, Stiefelhagen, Torralba, Urtasun, and Fidler]{tapaswi2016movieqa}
Makarand Tapaswi, Yukun Zhu, Rainer Stiefelhagen, Antonio Torralba, Raquel Urtasun, and Sanja Fidler.
\newblock Movie{QA}: Understanding stories in movies through question-answering.
\newblock In \emph{Proceedings of the IEEE Conference on Computer Vision and Pattern Recognition}, pages 4631--4640, 2016.

\bibitem[Team(2024)]{chameleonteam2024chameleon}
Chameleon Team.
\newblock Chameleon: Mixed-modal early-fusion foundation models.
\newblock \emph{arXiv preprint arXiv:2405.09818}, 2024.

\bibitem[Thrush et~al.(2022)Thrush, Jiang, Bartolo, Singh, Williams, Kiela, and Ross]{thrush2022winoground}
Tristan Thrush, Ryan Jiang, Max Bartolo, Amanpreet Singh, Adina Williams, Douwe Kiela, and Candace Ross.
\newblock Winoground: Probing vision and language models for visio-linguistic compositionality.
\newblock In \emph{Proceedings of the IEEE/CVF Conference on Computer Vision and Pattern Recognition}, pages 5238--5248, 2022.

\bibitem[Tian et~al.(2023{\natexlab{a}})Tian, Fan, Chen, Katabi, Krishnan, and Isola]{tian2023learning}
Yonglong Tian, Lijie Fan, Kaifeng Chen, Dina Katabi, Dilip Krishnan, and Phillip Isola.
\newblock Learning vision from models rivals learning vision from data.
\newblock \emph{arXiv preprint arXiv:2312.17742}, 2023{\natexlab{a}}.

\bibitem[Tian et~al.(2023{\natexlab{b}})Tian, Fan, Isola, Chang, and Krishnan]{tian2023stablerep}
Yonglong Tian, Lijie Fan, Phillip Isola, Huiwen Chang, and Dilip Krishnan.
\newblock Stablerep: Synthetic images from text-to-image models make strong visual representation learners.
\newblock In \emph{Thirty-seventh Conference on Neural Information Processing Systems}, 2023{\natexlab{b}}.
\newblock URL \url{https://openreview.net/forum?id=xpjsOQtKqx}.

\bibitem[Touvron et~al.(2023)Touvron, Lavril, Izacard, Martinet, Lachaux, Lacroix, Rozière, Goyal, Hambro, Azhar, Rodriguez, Joulin, Grave, and Lample]{touvron2023llama}
Hugo Touvron, Thibaut Lavril, Gautier Izacard, Xavier Martinet, Marie-Anne Lachaux, Timothée Lacroix, Baptiste Rozière, Naman Goyal, Eric Hambro, Faisal Azhar, Aurelien Rodriguez, Armand Joulin, Edouard Grave, and Guillaume Lample.
\newblock Llama: Open and efficient foundation language models, 2023.

\bibitem[Tsimpoukelli et~al.(2021)Tsimpoukelli, Menick, Cabi, Eslami, Vinyals, and Hill]{tsimpoukelli2021multimodal}
Maria Tsimpoukelli, Jacob~L Menick, Serkan Cabi, SM~Eslami, Oriol Vinyals, and Felix Hill.
\newblock Multimodal few-shot learning with frozen language models.
\newblock \emph{Advances in Neural Information Processing Systems}, 34:\penalty0 200--212, 2021.

\bibitem[Udandarao et~al.(2024)Udandarao, Prabhu, Ghosh, Sharma, Torr, Bibi, Albanie, and Bethge]{udandarao2024zeroshot}
Vishaal Udandarao, Ameya Prabhu, Adhiraj Ghosh, Yash Sharma, Philip H.~S. Torr, Adel Bibi, Samuel Albanie, and Matthias Bethge.
\newblock No ``zero-shot'' without exponential data: Pretraining concept frequency determines multimodal model performance.
\newblock \emph{arXiv preprint arXiv:2404.04125}, 2024.

\bibitem[Uppal et~al.(2022)Uppal, Bhagat, Hazarika, Majumder, Poria, Zimmermann, and Zadeh]{uppal2020multimodal}
Shagun Uppal, Sarthak Bhagat, Devamanyu Hazarika, Navonil Majumder, Soujanya Poria, Roger Zimmermann, and Amir Zadeh.
\newblock Multimodal research in vision and language: A review of current and emerging trends.
\newblock \emph{Information Fusion}, 77:\penalty0 149--171, 2022.

\bibitem[Urbanek et~al.(2023)Urbanek, Bordes, Astolfi, Williamson, Sharma, and Romero-Soriano]{urbanek2023picture}
Jack Urbanek, Florian Bordes, Pietro Astolfi, Mary Williamson, Vasu Sharma, and Adriana Romero-Soriano.
\newblock A picture is worth more than 77 text tokens: Evaluating clip-style models on dense captions.
\newblock \emph{arXiv preprint arXiv:2312.08578}, 2023.

\bibitem[Vallaeys et~al.(2024)Vallaeys, Shukor, Cord, and Verbeek]{vallaeys2024improved}
Th{\'e}ophane Vallaeys, Mustafa Shukor, Matthieu Cord, and Jakob Verbeek.
\newblock Improved baselines for data-efficient perceptual augmentation of {LLMs}.
\newblock \emph{arXiv preprint arXiv:2403.13499}, 2024.

\bibitem[Van Den~Oord et~al.(2017)Van Den~Oord, Vinyals, and Kavukcuoglu]{van2017neural}
Aaron Van Den~Oord, Oriol Vinyals, and Koray Kavukcuoglu.
\newblock Neural discrete representation learning.
\newblock \emph{Advances in Neural Information Processing Systems}, 30, 2017.

\bibitem[Vaswani et~al.(2017)Vaswani, Shazeer, Parmar, Uszkoreit, Jones, Gomez, Kaiser, and Polosukhin]{transformers}
Ashish Vaswani, Noam Shazeer, Niki Parmar, Jakob Uszkoreit, Llion Jones, Aidan~N Gomez, \L~ukasz Kaiser, and Illia Polosukhin.
\newblock Attention is all you need.
\newblock In I.~Guyon, U.~Von Luxburg, S.~Bengio, H.~Wallach, R.~Fergus, S.~Vishwanathan, and R.~Garnett, editors, \emph{Advances in Neural Information Processing Systems}, volume~30. Curran Associates, Inc., 2017.
\newblock URL \url{https://proceedings.neurips.cc/paper_files/paper/2017/file/3f5ee243547dee91fbd053c1c4a845aa-Paper.pdf}.

\bibitem[Veit et~al.(2016)Veit, Matera, Neumann, Matas, and Belongie]{veit2016cocotext}
Andreas Veit, Tomas Matera, Lukas Neumann, Jiri Matas, and Serge Belongie.
\newblock {COCO-Text}: Dataset and benchmark for text detection and recognition in natural images.
\newblock \emph{arXiv preprint arXiv:1601.07140}, 2016.

\bibitem[Vidgen et~al.(2023)Vidgen, Kirk, Qian, Scherrer, Kannappan, Hale, and R{\"o}ttger]{vidgen2023simplesafetytests}
Bertie Vidgen, Hannah~Rose Kirk, Rebecca Qian, Nino Scherrer, Anand Kannappan, Scott~A Hale, and Paul R{\"o}ttger.
\newblock Simplesafetytests: a test suite for identifying critical safety risks in large language models.
\newblock \emph{arXiv preprint arXiv:2311.08370}, 2023.

\bibitem[Vincent(2011)]{vincent2011connection}
Pascal Vincent.
\newblock A connection between score matching and denoising autoencoders.
\newblock \emph{Neural Computation}, 23\penalty0 (7):\penalty0 1661--1674, 2011.

\bibitem[Vincent et~al.(2008)Vincent, Larochelle, Bengio, and Manzagol]{denoising_autoencoder}
Pascal Vincent, Hugo Larochelle, Yoshua Bengio, and Pierre-Antoine Manzagol.
\newblock Extracting and composing robust features with denoising autoencoders.
\newblock In \emph{Proceedings of the 25th International Conference on Machine Learning}, ICML '08, page 1096–1103, New York, NY, USA, 2008. Association for Computing Machinery.
\newblock ISBN 9781605582054.
\newblock \doi{10.1145/1390156.1390294}.
\newblock URL \url{https://doi.org/10.1145/1390156.1390294}.

\bibitem[Wah et~al.(2011)Wah, Branson, Welinder, Perona, and Belongie]{WahCUB_200_2011}
C.~Wah, S.~Branson, P.~Welinder, P.~Perona, and S.~Belongie.
\newblock Caltech-ucsd birds-200-2011.
\newblock Technical Report CNS-TR-2011-001, California Institute of Technology, 2011.

\bibitem[Wang et~al.(2022)Wang, Yang, Men, Lin, Bai, Li, Ma, Zhou, Zhou, and Yang]{pmlr-v162-wang22al}
Peng Wang, An~Yang, Rui Men, Junyang Lin, Shuai Bai, Zhikang Li, Jianxin Ma, Chang Zhou, Jingren Zhou, and Hongxia Yang.
\newblock {OFA}: Unifying architectures, tasks, and modalities through a simple sequence-to-sequence learning framework.
\newblock In Kamalika Chaudhuri, Stefanie Jegelka, Le~Song, Csaba Szepesvari, Gang Niu, and Sivan Sabato, editors, \emph{Proceedings of the 39th International Conference on Machine Learning}, volume 162 of \emph{Proceedings of Machine Learning Research}, pages 23318--23340. PMLR, 17--23 Jul 2022.
\newblock URL \url{https://proceedings.mlr.press/v162/wang22al.html}.

\bibitem[Wang et~al.(2023{\natexlab{a}})Wang, Lin, Li, Lin, Yang, Zhang, Liu, and Wang]{eqben}
Tan Wang, Kevin Lin, Linjie Li, Chung-Ching Lin, Zhengyuan Yang, Hanwang Zhang, Zicheng Liu, and Lijuan Wang.
\newblock Equivariant similarity for vision-language foundation models.
\newblock \emph{arXiv preprint arXiv:2303.14465}, 2023{\natexlab{a}}.

\bibitem[Wang et~al.(2023{\natexlab{b}})Wang, Lv, Yu, Hong, Qi, Wang, Ji, Yang, Zhao, Song, Xu, Xu, Li, Dong, Ding, and Tang]{wang2023cogvlm}
Weihan Wang, Qingsong Lv, Wenmeng Yu, Wenyi Hong, Ji~Qi, Yan Wang, Junhui Ji, Zhuoyi Yang, Lei Zhao, Xixuan Song, Jiazheng Xu, Bin Xu, Juanzi Li, Yuxiao Dong, Ming Ding, and Jie Tang.
\newblock Cogvlm: Visual expert for pretrained language models, 2023{\natexlab{b}}.

\bibitem[Wang et~al.(2020)Wang, Liu, Shen, Ng, Luo, Jin, Chan, Hengel, and Wang]{wang2020general}
Xinyu Wang, Yuliang Liu, Chunhua Shen, Chun~Chet Ng, Canjie Luo, Lianwen Jin, Chee~Seng Chan, Anton van~den Hengel, and Liangwei Wang.
\newblock On the general value of evidence, and bilingual scene-text visual question answering.
\newblock In \emph{Proceedings of the IEEE/CVF Conference on Computer Vision and Pattern Recognition}, pages 10126--10135, 2020.

\bibitem[Weidinger et~al.(2022)Weidinger, Uesato, Rauh, Griffin, Huang, Mellor, Glaese, Cheng, Balle, Kasirzadeh, et~al.]{weidinger2022taxonomy}
Laura Weidinger, Jonathan Uesato, Maribeth Rauh, Conor Griffin, Po-Sen Huang, John Mellor, Amelia Glaese, Myra Cheng, Borja Balle, Atoosa Kasirzadeh, et~al.
\newblock Taxonomy of risks posed by language models.
\newblock In \emph{Proceedings of the 2022 ACM Conference on Fairness, Accountability, and Transparency}, pages 214--229, 2022.

\bibitem[Wettig et~al.(2024)Wettig, Gupta, Malik, and Chen]{wettig2024qurating}
Alexander Wettig, Aatmik Gupta, Saumya Malik, and Danqi Chen.
\newblock Qurating: Selecting high-quality data for training language models.
\newblock \emph{arXiv preprint arXiv:2402.09739}, 2024.

\bibitem[Whitehead et~al.(2022)Whitehead, Petryk, Shakib, Gonzalez, Darrell, Rohrbach, and Rohrbach]{whitehead2022reliable}
Spencer Whitehead, Suzanne Petryk, Vedaad Shakib, Joseph Gonzalez, Trevor Darrell, Anna Rohrbach, and Marcus Rohrbach.
\newblock Reliable visual question answering: Abstain rather than answer incorrectly.
\newblock In \emph{European Conference on Computer Vision}, pages 148--166. Springer, 2022.

\bibitem[Wiles et~al.(2022)Wiles, Albuquerque, and Gowal]{wiles2022discovering}
Olivia Wiles, Isabela Albuquerque, and Sven Gowal.
\newblock Discovering bugs in vision models using off-the-shelf image generation and captioning.
\newblock \emph{arXiv preprint arXiv:2208.08831}, 2022.

\bibitem[Wolfe and Caliskan(2022)]{wolfe2022american}
Robert Wolfe and Aylin Caliskan.
\newblock American == white in multimodal language-and-image ai.
\newblock In \emph{Proceedings of the 2022 AAAI/ACM Conference on AI, Ethics, and Society}, pages 800--812, 2022.

\bibitem[Wolfe et~al.(2023)Wolfe, Yang, Howe, and Caliskan]{wolfe2023contrastive}
Robert Wolfe, Yiwei Yang, Bill Howe, and Aylin Caliskan.
\newblock Contrastive language-vision ai models pretrained on web-scraped multimodal data exhibit sexual objectification bias.
\newblock In \emph{Proceedings of the 2023 ACM Conference on Fairness, Accountability, and Transparency}, pages 1174--1185, 2023.

\bibitem[Wu et~al.(2018)Wu, Xiong, Yu, and Lin]{wu2018unsupervised}
Zhirong Wu, Yuanjun Xiong, Stella~X Yu, and Dahua Lin.
\newblock Unsupervised feature learning via non-parametric instance discrimination.
\newblock In \emph{Proceedings of the IEEE Conference on Computer Vision and Pattern Recognition}, pages 3733--3742, 2018.

\bibitem[Xu et~al.(2017)Xu, Zhao, Xiao, Wu, Zhang, He, and Zhuang]{xu2017msvdqa}
Dejing Xu, Zhou Zhao, Jun Xiao, Fei Wu, Hanwang Zhang, Xiangnan He, and Yueting Zhuang.
\newblock Video question answering via gradually refined attention over appearance and motion.
\newblock In \emph{Proceedings of the 25th ACM International Conference on Multimedia}, pages 1645--1653, 2017.

\bibitem[Xu et~al.(2024)Xu, Xie, Tan, Huang, Howes, Sharma, Li, Ghosh, Zettlemoyer, and Feichtenhofer]{xu2023metaclip}
Hu~Xu, Saining Xie, Xiaoqing~Ellen Tan, Po-Yao Huang, Russell Howes, Vasu Sharma, Shang-Wen Li, Gargi Ghosh, Luke Zettlemoyer, and Christoph Feichtenhofer.
\newblock Demystifying clip data.
\newblock In \emph{International Conference on Learning Representations}, 2024.
\newblock URL \url{https://openreview.net/pdf?id=5BCFlnfE1g}.

\bibitem[Xu et~al.(2016)Xu, Mei, Yao, and Rui]{xu2016msr-vtt}
Jun Xu, Tao Mei, Ting Yao, and Yong Rui.
\newblock {MSR-VTT}: A large video description dataset for bridging video and language.
\newblock In \emph{Proceedings of the IEEE conference on computer vision and pattern recognition}, pages 5288--5296, 2016.

\bibitem[Xu et~al.(2015)Xu, Xiong, Chen, and Corso]{xu2015jointly}
Ran Xu, Caiming Xiong, Wei Chen, and Jason Corso.
\newblock Jointly modeling deep video and compositional text to bridge vision and language in a unified framework.
\newblock In \emph{Proceedings of the AAAI conference on artificial intelligence}, volume~29, 2015.

\bibitem[Yang et~al.(2023)Yang, Peng, Li, Guo, Chen, Li, Ma, Zhou, Zhang, Loy, et~al.]{yang2023panoptic}
Jingkang Yang, Wenxuan Peng, Xiangtai Li, Zujin Guo, Liangyu Chen, Bo~Li, Zheng Ma, Kaiyang Zhou, Wayne Zhang, Chen~Change Loy, et~al.
\newblock Panoptic video scene graph generation.
\newblock In \emph{Proceedings of the IEEE/CVF Conference on Computer Vision and Pattern Recognition}, pages 18675--18685, 2023.

\bibitem[Yu et~al.(2022{\natexlab{a}})Yu, Li, Koh, Zhang, Pang, Qin, Ku, Xu, Baldridge, and Wu]{yu2021vector}
Jiahui Yu, Xin Li, Jing~Yu Koh, Han Zhang, Ruoming Pang, James Qin, Alexander Ku, Yuanzhong Xu, Jason Baldridge, and Yonghui Wu.
\newblock Vector-quantized image modeling with improved {VQGAN}.
\newblock In \emph{International Conference on Learning Representations}, 2022{\natexlab{a}}.
\newblock URL \url{https://openreview.net/forum?id=pfNyExj7z2}.

\bibitem[Yu et~al.(2022{\natexlab{b}})Yu, Wang, Vasudevan, Yeung, Seyedhosseini, and Wu]{yu2022coca}
Jiahui Yu, Zirui Wang, Vijay Vasudevan, Legg Yeung, Mojtaba Seyedhosseini, and Yonghui Wu.
\newblock Coca: Contrastive captioners are image-text foundation models.
\newblock \emph{Transactions on Machine Learning Research}, 2022{\natexlab{b}}.
\newblock ISSN 2835-8856.
\newblock URL \url{https://openreview.net/forum?id=Ee277P3AYC}.

\bibitem[Yu et~al.(2022{\natexlab{c}})Yu, Xu, Koh, Luong, Baid, Wang, Vasudevan, Ku, Yang, Ayan, et~al.]{parti}
Jiahui Yu, Yuanzhong Xu, Jing~Yu Koh, Thang Luong, Gunjan Baid, Zirui Wang, Vijay Vasudevan, Alexander Ku, Yinfei Yang, Burcu~Karagol Ayan, et~al.
\newblock Scaling autoregressive models for content-rich text-to-image generation.
\newblock \emph{arXiv preprint arXiv:2206.10789}, 2\penalty0 (3):\penalty0 5, 2022{\natexlab{c}}.

\bibitem[Yu et~al.(2023)Yu, Shi, Pasunuru, Muller, Golovneva, Wang, Babu, Tang, Karrer, Sheynin, et~al.]{yu2023scaling}
Lili Yu, Bowen Shi, Ramakanth Pasunuru, Benjamin Muller, Olga Golovneva, Tianlu Wang, Arun Babu, Binh Tang, Brian Karrer, Shelly Sheynin, et~al.
\newblock Scaling autoregressive multi-modal models: Pretraining and instruction tuning.
\newblock \emph{arXiv preprint arXiv:2309.02591}, 2023.

\bibitem[Yu et~al.(2019)Yu, Xu, Yu, Yu, Zhao, Zhuang, and Tao]{yu2019activitynet}
Zhou Yu, Dejing Xu, Jun Yu, Ting Yu, Zhou Zhao, Yueting Zhuang, and Dacheng Tao.
\newblock Activitynet-qa: A dataset for understanding complex web videos via question answering.
\newblock In \emph{Proceedings of the AAAI Conference on Artificial Intelligence}, volume~33, pages 9127--9134, 2019.

\bibitem[Yuan et~al.(2022)Yuan, Liu, Dikubab, Liu, Ji, Wu, and Bai]{yuan2022syntaxaware}
Ye~Yuan, Xiao Liu, Wondimu Dikubab, Hui Liu, Zhilong Ji, Zhongqin Wu, and Xiang Bai.
\newblock Syntax-aware network for handwritten mathematical expression recognition.
\newblock In \emph{Proceedings of the IEEE/CVF Conference on Computer Vision and Pattern Recognition (CVPR)}, pages 4553--4562, June 2022.

\bibitem[Yue et~al.(2023)Yue, Ni, Zhang, Zheng, Liu, Zhang, Stevens, Jiang, Ren, Sun, et~al.]{yue2023mmmu}
Xiang Yue, Yuansheng Ni, Kai Zhang, Tianyu Zheng, Ruoqi Liu, Ge~Zhang, Samuel Stevens, Dongfu Jiang, Weiming Ren, Yuxuan Sun, et~al.
\newblock {MMMU}: A massive multi-discipline multimodal understanding and reasoning benchmark for expert {AGI}.
\newblock \emph{arXiv preprint arXiv:2311.16502}, 2023.

\bibitem[Yuille and Kersten(2006)]{yuille2006vision}
Alan Yuille and Daniel Kersten.
\newblock Vision as bayesian inference: analysis by synthesis?
\newblock \emph{Trends in Cognitive Sciences}, 10\penalty0 (7):\penalty0 301--308, 2006.

\bibitem[Yuksekgonul et~al.(2023)Yuksekgonul, Bianchi, Kalluri, Jurafsky, and Zou]{aro_yuksekgonul2023when}
Mert Yuksekgonul, Federico Bianchi, Pratyusha Kalluri, Dan Jurafsky, and James Zou.
\newblock When and why vision-language models behave like bags-of-words, and what to do about it?
\newblock In \emph{International Conference on Learning Representations}, 2023.
\newblock URL \url{https://openreview.net/forum?id=KRLUvxh8uaX}.

\bibitem[Zbontar et~al.(2021)Zbontar, Jing, Misra, LeCun, and Deny]{zbontar2021barlow}
Jure Zbontar, Li~Jing, Ishan Misra, Yann LeCun, and St{\'e}phane Deny.
\newblock Barlow twins: Self-supervised learning via redundancy reduction.
\newblock In \emph{International Conference on Machine Learning}, pages 12310--12320. PMLR, 2021.

\bibitem[Zellers et~al.(2021)Zellers, Lu, Hessel, Yu, Park, Cao, Farhadi, and Choi]{rowan2021merlot}
Rowan Zellers, Ximing Lu, Jack Hessel, Youngjae Yu, Jae~Sung Park, Jize Cao, Ali Farhadi, and Yejin Choi.
\newblock Merlot: Multimodal neural script knowledge models.
\newblock In M.~Ranzato, A.~Beygelzimer, Y.~Dauphin, P.S. Liang, and J.~Wortman Vaughan, editors, \emph{Advances in Neural Information Processing Systems}, volume~34, pages 23634--23651. Curran Associates, Inc., 2021.
\newblock URL \url{https://proceedings.neurips.cc/paper_files/paper/2021/file/c6d4eb15f1e84a36eff58eca3627c82e-Paper.pdf}.

\bibitem[Zeng et~al.(2022)Zeng, Zhang, and Li]{zeng2021multi}
Yan Zeng, Xinsong Zhang, and Hang Li.
\newblock Multi-grained vision language pre-training: Aligning texts with visual concepts.
\newblock In Kamalika Chaudhuri, Stefanie Jegelka, Le~Song, Csaba Szepesv{\'{a}}ri, Gang Niu, and Sivan Sabato, editors, \emph{International Conference on Machine Learning, {ICML} 2022, 17-23 July 2022, Baltimore, Maryland, {USA}}, volume 162 of \emph{Proceedings of Machine Learning Research}, pages 25994--26009. {PMLR}, 2022.
\newblock URL \url{https://proceedings.mlr.press/v162/zeng22c.html}.

\bibitem[Zhai et~al.(2023{\natexlab{a}})Zhai, Yang, Zhao, Xu, Shen, Zhao, Keutzer, Li, Yan, and Fan]{zhai2023halle}
Bohan Zhai, Shijia Yang, Xiangchen Zhao, Chenfeng Xu, Sheng Shen, Dongdi Zhao, Kurt Keutzer, Manling Li, Tan Yan, and Xiangjun Fan.
\newblock Halle-switch: Rethinking and controlling object existence hallucinations in large vision language models for detailed caption.
\newblock \emph{arXiv preprint arXiv:2310.01779}, 2023{\natexlab{a}}.

\bibitem[Zhai et~al.(2023{\natexlab{b}})Zhai, Mustafa, Kolesnikov, and Beyer]{zhai2023sigmoid}
Xiaohua Zhai, Basil Mustafa, Alexander Kolesnikov, and Lucas Beyer.
\newblock Sigmoid loss for language image pre-training.
\newblock In \emph{2023 IEEE/CVF International Conference on Computer Vision (ICCV)}, pages 11941--11952, Los Alamitos, CA, USA, oct 2023{\natexlab{b}}. IEEE Computer Society.
\newblock \doi{10.1109/ICCV51070.2023.01100}.
\newblock URL \url{https://doi.ieeecomputersociety.org/10.1109/ICCV51070.2023.01100}.

\bibitem[Zhang et~al.(2023{\natexlab{a}})Zhang, Zhang, Zhang, and Kweon]{zhang2023texttoimage}
Chenshuang Zhang, Chaoning Zhang, Mengchun Zhang, and In~So Kweon.
\newblock Text-to-image diffusion model in generative {AI}: A survey.
\newblock \emph{arXiv preprint arXiv:2303.07909}, 2023{\natexlab{a}}.

\bibitem[Zhang et~al.(2023{\natexlab{b}})Zhang, Li, and Bing]{zhang2023videollama}
Hang Zhang, Xin Li, and Lidong Bing.
\newblock Video-{LL}a{MA}: An instruction-tuned audio-visual language model for video understanding.
\newblock In Yansong Feng and Els Lefever, editors, \emph{Proceedings of the 2023 Conference on Empirical Methods in Natural Language Processing: System Demonstrations}, pages 543--553, Singapore, December 2023{\natexlab{b}}. Association for Computational Linguistics.
\newblock \doi{10.18653/v1/2023.emnlp-demo.49}.
\newblock URL \url{https://aclanthology.org/2023.emnlp-demo.49}.

\bibitem[Zhang et~al.(2024{\natexlab{a}})Zhang, Huang, Jin, and Lu]{zhang2024visionlanguage}
Jingyi Zhang, Jiaxing Huang, Sheng Jin, and Shijian Lu.
\newblock Vision-language models for vision tasks: A survey.
\newblock \emph{IEEE Transactions on Pattern Analysis and Machine Intelligence}, 2024{\natexlab{a}}.

\bibitem[Zhang et~al.(2024{\natexlab{b}})Zhang, Mo, Chen, Sun, and Su]{zhang2024magicbrush}
Kai Zhang, Lingbo Mo, Wenhu Chen, Huan Sun, and Yu~Su.
\newblock Magic{B}rush: A manually annotated dataset for instruction-guided image editing.
\newblock \emph{Advances in Neural Information Processing Systems}, 36, 2024{\natexlab{b}}.

\bibitem[Zhang et~al.(2019)Zhang, Jiang, Cui, Garnett, and Chen]{zhang2019d}
Muhan Zhang, Shali Jiang, Zhicheng Cui, Roman Garnett, and Yixin Chen.
\newblock {D-VAE}: A variational autoencoder for directed acyclic graphs.
\newblock In \emph{Advances in Neural Information Processing Systems}, pages 1586--1598, 2019.

\bibitem[Zhang et~al.(2022)Zhang, Roller, Goyal, Artetxe, Chen, Chen, Dewan, Diab, Li, Lin, et~al.]{zhang2022opt}
Susan Zhang, Stephen Roller, Naman Goyal, Mikel Artetxe, Moya Chen, Shuohui Chen, Christopher Dewan, Mona Diab, Xian Li, Xi~Victoria Lin, et~al.
\newblock {OPT}: Open pre-trained transformer language models.
\newblock \emph{arXiv preprint arXiv:2205.01068}, 2022.

\bibitem[Zhang et~al.(2023{\natexlab{c}})Zhang, Zhang, Gu, Zhou, Lipka, Yang, and Sun]{zhang2024llavar}
Yanzhe Zhang, Ruiyi Zhang, Jiuxiang Gu, Yufan Zhou, Nedim Lipka, Diyi Yang, and Tong Sun.
\newblock {LLaVAR}: Enhanced visual instruction tuning for text-rich image understanding.
\newblock \emph{arXiv preprint arXiv:2306.17107}, 2023{\natexlab{c}}.

\bibitem[Zhang et~al.(2023{\natexlab{d}})Zhang, Huang, Ma, Li, Luo, Xie, Qin, Luo, Li, Liu, et~al.]{zhang2023recognize}
Youcai Zhang, Xinyu Huang, Jinyu Ma, Zhaoyang Li, Zhaochuan Luo, Yanchun Xie, Yuzhuo Qin, Tong Luo, Yaqian Li, Shilong Liu, et~al.
\newblock Recognize anything: A strong image tagging model.
\newblock \emph{arXiv preprint arXiv:2306.03514}, 2023{\natexlab{d}}.

\bibitem[Zhao et~al.(2024)Zhao, Gundavarapu, Yuan, Zhou, Yan, Sun, Friedman, Qian, Weyand, Zhao, et~al.]{zhao2024videoprism}
Long Zhao, Nitesh~B Gundavarapu, Liangzhe Yuan, Hao Zhou, Shen Yan, Jennifer~J Sun, Luke Friedman, Rui Qian, Tobias Weyand, Yue Zhao, et~al.
\newblock Videoprism: A foundational visual encoder for video understanding.
\newblock \emph{arXiv preprint arXiv:2402.13217}, 2024.

\bibitem[Zhao et~al.(2022)Zhao, Zhang, Zhu, Shen, Lee, Lu, and Yin]{vlchecklist}
Tiancheng Zhao, Tianqi Zhang, Mingwei Zhu, Haozhan Shen, Kyusong Lee, Xiaopeng Lu, and Jianwei Yin.
\newblock Vl-checklist: Evaluating pre-trained vision-language models with objects, attributes and relations.
\newblock \emph{arXiv preprint arXiv:2207.00221}, 2022.

\bibitem[Zheng et~al.(2023)Zheng, He, and Wang]{zheng2023minigpt5}
Kaizhi Zheng, Xuehai He, and Xin~Eric Wang.
\newblock {MiniGPT-5}: Interleaved vision-and-language generation via generative vokens.
\newblock \emph{arXiv preprint arXiv:2310.02239}, 2023.

\bibitem[Zhou et~al.(2022)Zhou, Yang, Loy, and Liu]{zhou2022learning}
Kaiyang Zhou, Jingkang Yang, Chen~Change Loy, and Ziwei Liu.
\newblock Learning to prompt for vision-language models.
\newblock \emph{International Journal of Computer Vision}, 130\penalty0 (9):\penalty0 2337--2348, 2022.

\bibitem[Zhou and Shimada(2023)]{zhou2023vision}
Yutong Zhou and Nobutaka Shimada.
\newblock Vision + language applications: A survey.
\newblock In \emph{Proceedings of the IEEE/CVF Conference on Computer Vision and Pattern Recognition (CVPR) Workshops}, pages 826--842, June 2023.

\bibitem[Zhu et~al.(2023{\natexlab{a}})Zhu, Chen, Shen, Li, and Elhoseiny]{zhu2023minigpt}
Deyao Zhu, Jun Chen, Xiaoqian Shen, Xiang Li, and Mohamed Elhoseiny.
\newblock {MiniGPT-4}: Enhancing vision-language understanding with advanced large language models.
\newblock \emph{arXiv preprint arXiv:2304.10592}, 2023{\natexlab{a}}.

\bibitem[Zhu et~al.(2023{\natexlab{b}})Zhu, Hessel, Awadalla, Gadre, Dodge, Fang, Yu, Schmidt, Wang, and Choi]{zhu2023multimodal}
Wanrong Zhu, Jack Hessel, Anas Awadalla, Samir~Yitzhak Gadre, Jesse Dodge, Alex Fang, Youngjae Yu, Ludwig Schmidt, William~Yang Wang, and Yejin Choi.
\newblock Multimodal c4: An open, billion-scale corpus of images interleaved with text.
\newblock In A.~Oh, T.~Naumann, A.~Globerson, K.~Saenko, M.~Hardt, and S.~Levine, editors, \emph{Advances in Neural Information Processing Systems}, volume~36, pages 8958--8974. Curran Associates, Inc., 2023{\natexlab{b}}.
\newblock URL \url{https://proceedings.neurips.cc/paper_files/paper/2023/file/1c6bed78d3813886d3d72595dbecb80b-Paper-Datasets_and_Benchmarks.pdf}.

\end{thebibliography}

\end{document}